\documentclass[10pt,twocolumn,letterpaper]{article}

\usepackage[pagenumbers]{cvpr}

\usepackage{multirow}
\usepackage{booktabs}
\usepackage{graphicx}
\usepackage{xcolor}
\usepackage{longtable}
\usepackage{array}
\usepackage{amsmath}
\usepackage{amssymb}

\usepackage{algorithm}
\usepackage{algorithmic}
\newcommand{\appref}[1]{\hyperref[#1]{Appendix~\ref*{#1}}}
\newcommand{\holdoutmaintableref}{\hyperref[tab:scene_holdout]{Table~\ref*{tab:scene_holdout}}}
\usepackage{microtype}

\patchcmd{\abstract}{\vspace*{12pt}}{\vspace*{6pt}}{}{
  \PackageError{arxiv-layout}{Could not adjust abstract spacing}{Check the cvpr abstract definition.}
}

\definecolor{cvprblue}{rgb}{0.21,0.49,0.74}
\usepackage[breaklinks,colorlinks,allcolors=cvprblue,hypertexnames=false,pdfusetitle]{hyperref}

\def\confName{CVPR}
\def\confYear{2027}

\title{Beyond Linear Decodability: Understanding and Improving Instruction Following in World Action Models}

\author{
Bokai Ji$^{1}$\thanks{Equal contribution.} \quad
Jiapeng Chen$^{2*}$ \quad
Yi Zhang$^{3*}$ \quad
Zeyuan Ding$^{3}$ \quad
Yong Dai$^{3}$ \\[3pt]
Xiaozhu Ju$^{3}$ \quad
Jian Tang$^{3}$ \quad
Lemao Liu$^{2}$ \quad
Dejing Dou$^{2}$ \quad
Yong Li$^{1}$\thanks{Corresponding author.} \\[6pt]
{\normalsize $^1$Renmin University of China} \\
{\normalsize $^2$Fudan University} \\
{\normalsize $^3$Beijing Humanoid Robot Innovation Center}
}

\hypersetup{pdfauthor={Bokai Ji, Jiapeng Chen, Yi Zhang, Zeyuan Ding, Yong Dai, Xiaozhu Ju, Jian Tang, Lemao Liu, Dejing Dou, Yong Li}}

\makeatletter
\patchcmd{\@maketitle}{\vskip .5em}{%
  \vskip .5em
  {\normalsize Project Homepage: \url{https://ruflect-intelligence.github.io/LDS-Homepage/}\par}%
  \vskip .5em
}{}{\PackageError{arxiv-layout}{Could not add the project homepage}{Check the cvpr title layout.}}
\makeatother
\usepackage{placeins}
\begin{document}
\maketitle
\begin{abstract}
World action models (WAMs) exhibit strong manipulation capabilities yet often fail to follow test-time language instructions.
Prior work shows that non-following (NF) WAMs encode information about the instructed target, but this information does not control their behavior.
By comparing NF WAM with its instruction-following (IF) counterpart, we find that target information is substantially more linearly decodable in IF WAMs, despite being encoded in both.
Motivated by this observation, we introduce linear decodability supervision (LDS), a simple yet effective auxiliary training objective that promotes the linear decodability of the instructed target information.
On LIBERO-CF, InternW0-$\Delta$ with LDS achieves a success rate of 74.6\% and a grounding rate of 89.2\%, surpassing the strongest baseline by 26.8 and 42.9 percentage points, respectively.
The effectiveness of LDS challenges the prevailing interpretation that encoded target information does not contribute to WAMs' behavior.
To reconcile this apparent contradiction, we propose a prior–instruction competition mechanism, which explains NF and IF behavior as arising from the dominance of prior and instruction signals, respectively.
We provide causal evidence for this mechanism by evaluating two training-free interventions derived from it.
We reinterpret LDS through this mechanism, showing that it acts as a form of instruction signal enhancement.
Finally, more comprehensive evaluations across diverse manipulation tasks and visual domains demonstrate the consistent effectiveness of LDS.

\end{abstract}
\section{Introduction}
\label{sec:intro}

Instruction following is a fundamental capability of vision-language-action (VLA) models and world action models (WAM), requiring them to manipulate objects according to natural language instructions.
Despite their ability to perform complex manipulation tasks~\cite{DBLP:journals/corr/abs-2601-21998, DBLP:journals/corr/abs-2603-16666, DBLP:journals/corr/abs-2609-07398}, recent works find that these models have a strong bias to execute training-time instruction, ignoring instructions given at test time~\cite{DBLP:journals/corr/abs-2602-17659, DBLP:journals/corr/abs-2603-00592, DBLP:journals/ral/GlossopCBSL26}.
Prior studies on VLAs~\cite{DBLP:journals/corr/abs-2602-17659, DBLP:journals/corr/abs-2606-23641} address this issue by recalibrating attention toward instruction-relevant tokens.
However, the mechanisms underlying similar failures in WAMs remain unclear.
In the WAM architecture we study, the ones employing Wan~\cite{DBLP:journals/corr/abs-2503-20314} as their backbone, language is injected into visual representations through cross-attention, rather than competing with visual tokens for attention mass within a shared self-attention operation.

\begin{figure*}
    \centering
    \includegraphics[width=1\linewidth]{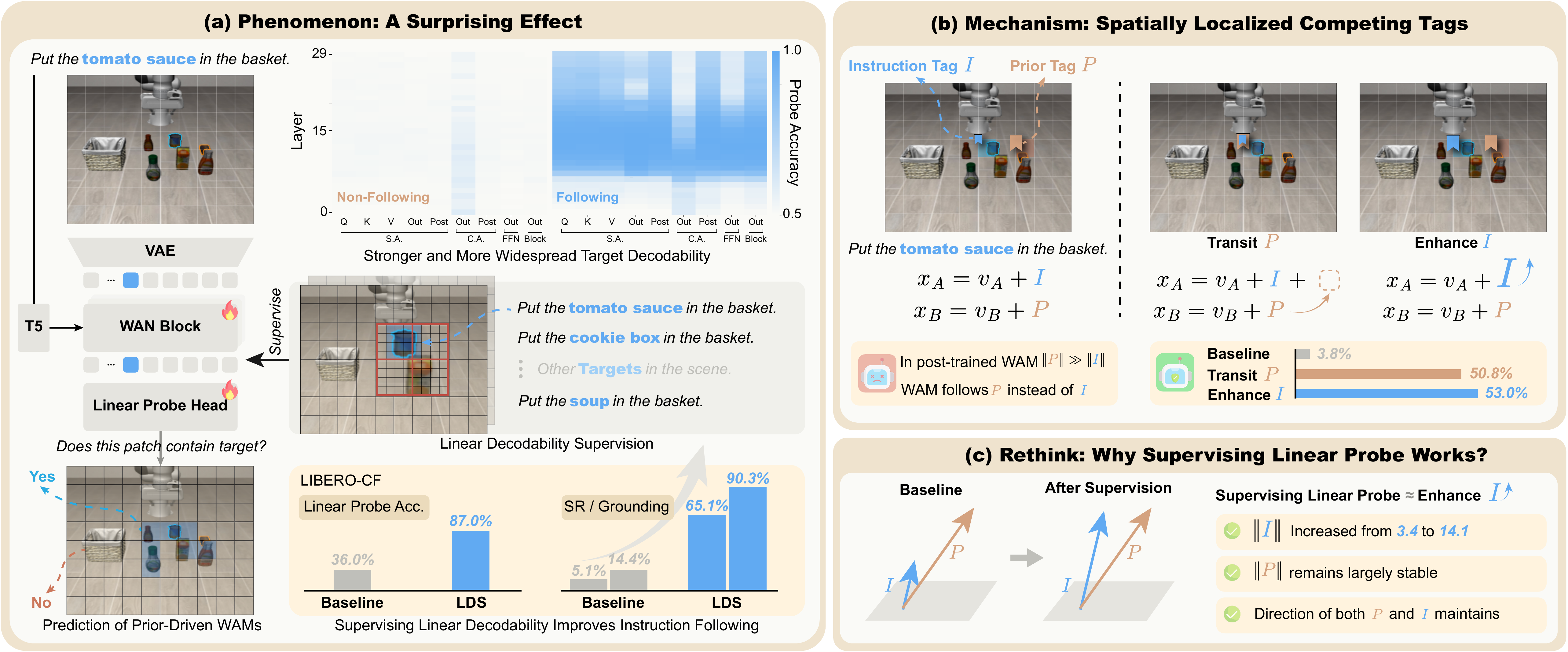}
    \caption{\textbf{Overview.}
    (a) \textbf{A surprising phenomenon.} Auxiliary LDS improves instruction-following success from 5.1\% to 65.1\%, despite removing the linear head at inference.
    (b) \textbf{Prior--instruction competition.} A weak instruction signal $I$ competes with a prior-associated signal, and manipulating either signal substantially improves instruction following without training.
    (c) \textbf{Why LDS works.} Counterfactual grounding supervision amplifies the existing instruction signal $I$ while largely preserving its direction.
    }
    \label{fig:main}
    \vspace{-0.3cm}
\end{figure*}

To study IF in WAMs, we start from training an IF WAM with collected counterfactual data and comparing its internal states with its non-following (NF) counterparts.
Similar to~\cite{jiang2026encodedcontrolrevealinggrounding}, we probe instruction-related information across different layers and internal components in WAMs and find both IF and NF WAMs encode such information.
However, as shown in Figure~\ref{fig:main}, the IF WAM exhibit significantly stronger linear decodability of the instructed-target information than its NF counterpart.
Motivated by such observation, we propose Linear-Decodability Supervision (LDS) that directly improves the linear decodability of the intructed-target information as an auxiliary loss along side the flow-matching loss.
InternW0-$\Delta$ with LDS achieves a success rate of 74.6\% and a grounding rate of 89.2\%, surpassing the strongest baseline by 26.8 and 42.9 percentage points, respectively.
Despite the effectiveness of LDS, these results challenges the prevailing interpretation that encoded target information does not contribute to WAMs' behavior~\cite{jiang2026encodedcontrolrevealinggrounding}.

We propose a prior–instruction competition mechanism to reconcile such contradiction.
Specifically, there are two latent tags, a \emph{prior tag} $P$ and an \emph{instruction tag} $I$, are attached to the prior-preferred and the instruction-specified visual target, respectively.
Here, tags refer to additive components of intermediate visual-token representations.
Both of these two tags contribute to WAM's behavior.
They compete with each other, and the winning one finally determines the action.
Therefore, NF and IF WAMs can be explained as the winning of prior tag $P$ and instruction tag $I$, respectively.
We validate our proposed hypothesis through two training-free interventions derived from it: (1) enhancing the instruction tag $I$; (2) transiting the prior tag $P$ toward the instructed target.
Remarkably, enhancing $I$ improves OpenWAM's success/grounding rate from 3.8/7.0\% to 53.0/80.0\%, while transiting $P$ achieves 50.8/84.0\%, on a picking-target subset of LIBERO-CF without updating the model's parameters.
Together, these results provide causal evidence supporting our proposed mechanism.
We further revisit the effectiveness of LDS through this mechanistic lens.
Since LDS aims to correctly identify the visual target under counterfactual instructions and the easiest way to do so is to exploit the existing $I$, LDS can be seen as a specific form of enhancing $I$.
We evaluate the explanation by empirically measuring the magnitude and direction of $I$ and $P$ and find the result supports the above explanation.
Specifically, the magnitude of $I$ increases whereas the magnitude of $P$ changes only modestly.
Meanwhile, the direction of $I$ and $P$ before and after LDS remains approximately the same.

Finally, we conduct more extensive experiments to evaluate the effectiveness of LDS.
On LIBERO-CF, InternW0-$\Delta$ equipped with LDS achieves a success/grounding rate of 74.6/89.2\%, outperforming the strongest baseline by 26.8 and 42.9 percentage points, respectively.
On LangGap and RoboTwin-IF, LDS further demonstrates its effectiveness in both object selection and placement-destination grounding, achieving success rates (SR) of 71.4\% and 47.4\% with InternW0-$\Delta$, respectively.
This implies a more general prior-instruction competition mechanism of visual targets in WAMs.
Experiments on RoboTwin-IF and real-world tasks further establish the effectiveness of LDS across diverse visual domains beyond LIBERO-based benchmarks.
Evaluations on held-out splits of LIBERO-CF and LangGap demonstrate strong generalization to unseen initial configurations, with LDS maintaining robust performance even in completely out-of-distribution scenes.
Our main contributions are summarized as follows:
\begin{enumerate}
    \item \textbf{A simple yet effective way to improve IF in WAMs.} We propose LDS, an auxiliary training objective that improves IF of WAMs without the need of counterfactual demonstrations collection. Experimental results show LDS outperforming existing methods by a large margin.
    \item \textbf{A novel prior-instruction competing mechanism.} We provide a novel way of understanding ``encode but not control'' phenomenon by introducing competition with prior signals. We evaluate the correctness of the proposed mechanism through causal interventions and explain the effectiveness of LDS under such mechanism.
\end{enumerate}

% Compact the transition after the contribution list so the heading and
% opening lines can use the remaining space in the second-page column.
\vspace{-8pt}
\section{Related Works}
\label{sec:related_work}

\textbf{World Action Models.}
WAMs learn robot policies by jointly modeling visual dynamics and actions, often leveraging pretrained video generation models.
DreamZero~\cite{DBLP:journals/corr/abs-2602-15922} builds on Wan2.1~\cite{DBLP:journals/corr/abs-2503-20314}, while LingBot-VA~\cite{DBLP:journals/corr/abs-2601-21998}, Fast-WAM~\cite{DBLP:journals/corr/abs-2603-16666}, and OpenWAM~\cite{DBLP:journals/corr/abs-2609-07398} adopt Wan2.2.
Their shared architectural foundations motivate our investigation of how instruction information interacts with visual representations to influence action generation.
Other WAMs, such as Cosmos Policy~\cite{DBLP:conf/iclr/KimGLLGLLSLFG26} and ImageWAM~\cite{DBLP:journals/corr/abs-2606-19531}, employ different generative backbones.
Our mechanistic findings therefore primarily concern Wan-based WAMs.

\textbf{Instruction Following in Embodied Policies.}
Recent studies reveal that embodied policies may rely on visual shortcuts rather than faithfully following language instructions.
LIBERO-CF~\cite{DBLP:journals/corr/abs-2602-17659}, LangGap~\cite{DBLP:journals/corr/abs-2603-00592}, and InstructMove~\cite{DBLP:journals/corr/abs-2608-22990} expose such failures through instruction-sensitive evaluations, while Counterfactual Action Guidance~\cite{DBLP:journals/corr/abs-2602-17659} and IGAR~\cite{DBLP:journals/corr/abs-2603-06001} seek to mitigate them.
However, the underlying mechanisms in WAMs remain underexplored.

Importantly, decodable information need not causally influence model behavior~\cite{DBLP:journals/tacl/ElazarRJG21}.
Recent work~\cite{jiang2026encodedcontrolrevealinggrounding} demonstrates this decodability-control gap in both VLAs and WAMs, where instructed targets remain encoded despite visually biased actions.
Building on this finding, we investigate competing instruction-dependent and visual-prior-associated signals in WAMs, establish their behavioral roles through representation interventions, and show how auxiliary linear-probe supervision strengthens IF by amplifying a behaviorally relevant instruction signal.

\section{Improving Instruction Following of WAMs via Linear Decodability Supervision}
\label{sec:dec2behavior}

In this section, we investigate whether the linear decodability of instructed target information in WAMs is relevant to IF behavior.
We first construct IF WAMs through collecting and training on counterfactual demonstrations, enabling a controlled comparison with their NF counterparts.
We find that IF WAMs exhibit substantially stronger linear decodability of instructed target information than NF WAMs, and that higher decodability is associated with higher IF success rates.
Motivated by these observations, we introduce linear-decodability supervision (LDS) that explicitly promotes such linear decodability during training.
We reveal that LDS leads to substantial IF improvements.

% \subsection{Formalization of Wan-based WAMs}

% We study WAMs that use a Wan-based video diffusion transformer (DiT)~\cite{DBLP:journals/corr/abs-2503-20314}, such as OpenWAM~\cite{DBLP:journals/corr/abs-2609-07398}.
% A video variational autoencoder encodes the visual input into latents, which the DiT patchifies into a grid of visual tokens.
% Each token is indexed by its position on this grid, so its activations throughout the DiT can be aligned with a patch of the preprocessed observation.
% Let $X^{\ell,\mathrm{in}}$ be the visual residual stream entering block $\ell$.
% Self-attention (SA), text cross-attention (CA), and the feed-forward network (FFN) update it in sequence:
% \begin{equation}
%     \nonumber
%     \begin{aligned}
%         X^{\ell,\mathrm{PostSA}} &= X^{\ell,\mathrm{in}}+\Delta_{\mathrm{SA}}^\ell,\\
%         X^{\ell,\mathrm{PostCA}} &= X^{\ell,\mathrm{PostSA}}+\Delta_{\mathrm{CA}}^\ell,\\
%         X^{\ell,\mathrm{out}} &= X^{\ell,\mathrm{PostCA}}+\Delta_{\mathrm{FFN}}^\ell.
%     \end{aligned}
% \end{equation}
% Each $\Delta^\ell$ is the update computed from the preceding residual state.
% In CA, queries are the post-SA visual states and the instruction $L$ is encoded by umT5~\cite{DBLP:journals/corr/abs-2503-20314} to supply keys and values.
% The attention-weighted text values are projected and added to the corresponding visual tokens as $\Delta_{\mathrm{CA}}^\ell$.

\begin{figure}
    \centering
    \includegraphics[width=1\linewidth]{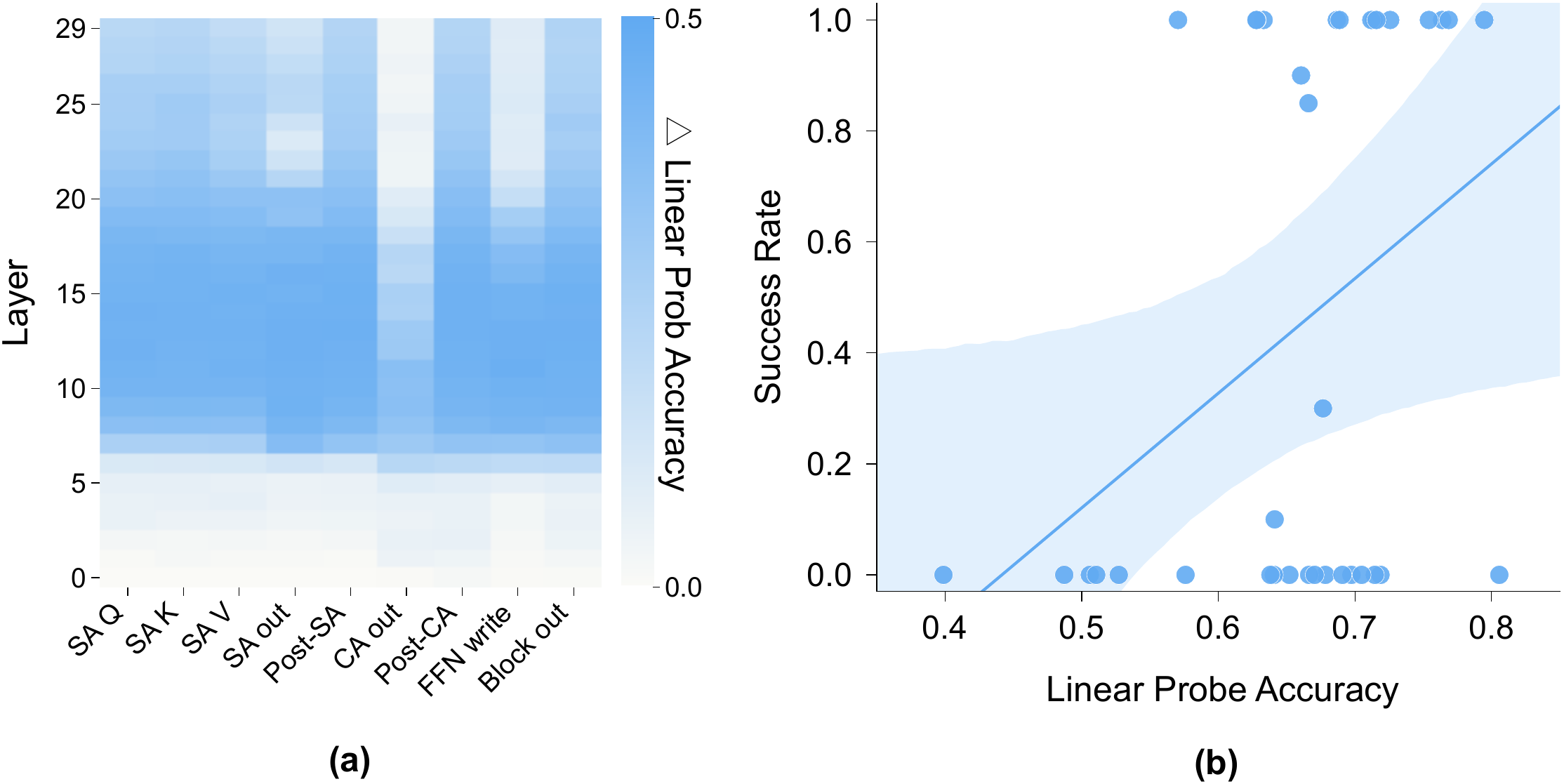}
    \caption{\textbf{Instruction-Following WAM show stronger and more wide spread linear decodability.} (a) Layer- and site-wise gain in instruction-target decodability from RoboTwin-IF training. (b) Probe accuracy versus instruction success on the scenes in $\mathcal{D}_{\mathrm{probe}}$. Each point is an instruction. The line shows the result of least-squares fit and the shaded band represents 95\% confidence interval of the fitted mean. The trend is positive with a slope  of 2.07.}
    \label{fig:correlation}
    \vspace{-0.3cm}
\end{figure}

\subsection{Constructing Instruction-Following WAMs}
\label{sec:construct}

We start by constructing IF WAMs to enable direct comparison on internal states with its NF counterpart.
Previous studies suggest that IF failures can arise from insufficient counterfactual supervision~\cite{DBLP:journals/ral/GlossopCBSL26,DBLP:journals/corr/abs-2602-17659}.
When actions and future are strongly predictable from the current visual observation, WAMs may exploit an observation-conditioned shortcut and ignore language instruction.
To mitigate this shortcut, training should expose models to situations where the same visual observation admits different target behaviors depending on the instruction, making language information necessary to distinguish the correct behavior.

Therefore, we develop RoboTwin-IF, a pipeline built upon RoboTwin2.0~\cite{DBLP:journals/corr/abs-2506-18088} to synthesize counterfactual tasks and demonstrations.
Specifically, counterfactual demonstrations share the same scene and identical initial visual observations but differ in their language instructions and corresponding behaviors.
For example, in a scene containing a red and a green cup, we include demonstrations of grasping each cup under its respective instruction.
This makes observation-only predictions insufficient to minimize training loss, encouraging the model to condition its behavior on both visual observation and language instructions.

We partition the scenes generated by RoboTwin-IF into disjoint training and held-out evaluation sets.
We collect expert demonstrations for the training scenes, forming the counterfactual dataset $\mathcal{D}_{\mathrm{CF}}$.
Given a pre-trained NF WAM $M_{\mathrm{NF}}$, we construct its IF counterpart by fine-tuning on $\mathcal{D}_{\mathrm{CF}}$
\begin{equation}
    \nonumber
    M_{\mathrm{IF}} =
    \operatorname{FineTune}_{\mathcal L_{\mathrm{FM}}}
    (M_{\mathrm{NF}}, \mathcal D_{\mathrm{CF}})
\end{equation}
where $\mathcal L_{\mathrm{FM}}$ denotes the flow-matching (FM) training loss.
For OpenWAM, SR on held-out RoboTwin-IF scenes increases from 11.05\% to 47.76\% after training, demonstrating a substantial improvement in IF capability.
RoboTwin-IF and model details are in \appref{app:robotwin_if} and \appref{app:probe}.

\subsection{Stronger Linear Decodability}
\label{sec:stronger_linear_decodability}

Having WAMs $m \in \{\text{NF}, \text{IF}\}$, we investigate whether the difference in IF capability is reflected in the linear decodability of information about instructed targets.
Let $x_{s,j,i}^{m,\ell,c}$ be the activation of the token aligned with head-camera patch $i$ when processing initial observation of scene $s$ paired with instruction $L_j$ that requires to manipulate candidate $j$.
Here $\ell$ indexes the block and $c$ identifies the computation site within it.
We read nine sites per block in following analysis: the SA query, key, and value projections before query/key normalization and rotary position encoding, the SA and CA attention outputs before their respective output projections, the Post-SA and Post-CA residual states, the gated FFN update, and the block output.
% We read nine sites per block in following analysis: the SA query, key, and value projections before query/key normalization and rotary position encoding, the SA and CA attention outputs before their respective output projections, the Post-SA and Post-CA residual states, the gated FFN update $\Delta_{\mathrm{FFN}}^\ell$, and the block output $X^{\ell,\mathrm{out}}$.

We construct a probe panel of 515 instructions across 256 RoboTwin-IF scenes.
Each scene $s$ pairs the same observation $O_s$ with multiple instructions $L_{j}$ selecting different manipulation targets $j$.
We assign each patch to the candidate that covers the largest fraction of it.
For instruction $L_j$, the label $y_{s,j,i}$ is 1 if patch $i$ belongs to the selected candidate $j$ and 0 if it belongs to another candidate.
Switching the instruction changes which object patches are annotated as $y_{s,j,i}=1$ while holding $O_s$ and the patch grid fixed.

For each probing point $(m,\ell,c)$, we fit a linear probe shared across all patches and instructions.
Its logit for the label $y_{s,j,i}$ is
\begin{equation}
    z_{s,j,i}^{m,\ell,c}
    = (w^{m,\ell,c})^\top x_{s,j,i}^{m,\ell,c} + b^{m,\ell,c},
    \qquad z_{s,j,i}^{m,\ell,c} \in \mathbb{R}^1
    \label{eq:patch_probe}
\end{equation}
A patch is classified as target if its logit is positive, and as non-target otherwise.
We fit only the probe parameters $(w_{\ell,c}^m,b_{\ell,c}^m)$ with frozen $x_{s,j,i}^{m,\ell,c}$.
For each $(O_s, L_{j})$, bi-directional accuracy averages target recall and non-target rejection.
See \appref{app:probe} for more details.

Figure~\ref{fig:correlation}(a) shows the IF WAM's advantage to its NF counterpart in scene-averaged bi-directional accuracy at each layer and computation site.
Averaged across layers and sites, bi-directional accuracy rises from 51.5\% for NF WAM to 77.5\% for IF WAM which is a gain of 26.0 percentage points.
Figure~\ref{fig:correlation}(b) further links linear decodability to SR.
Instructions with higher linear probe accuracy tend to achieve higher SR.
This association suggests that linear decodability of instructed target information is behaviorally relevant, although it does not establish causality.
It motivates testing whether explicitly strengthening linear decodability of instructed target information can improve IF of WAMs.

\subsection{Linear-Decodability Supervision}
\label{sec:surprising_effect}

\begin{figure}
    \centering
    \includegraphics[width=1\linewidth]{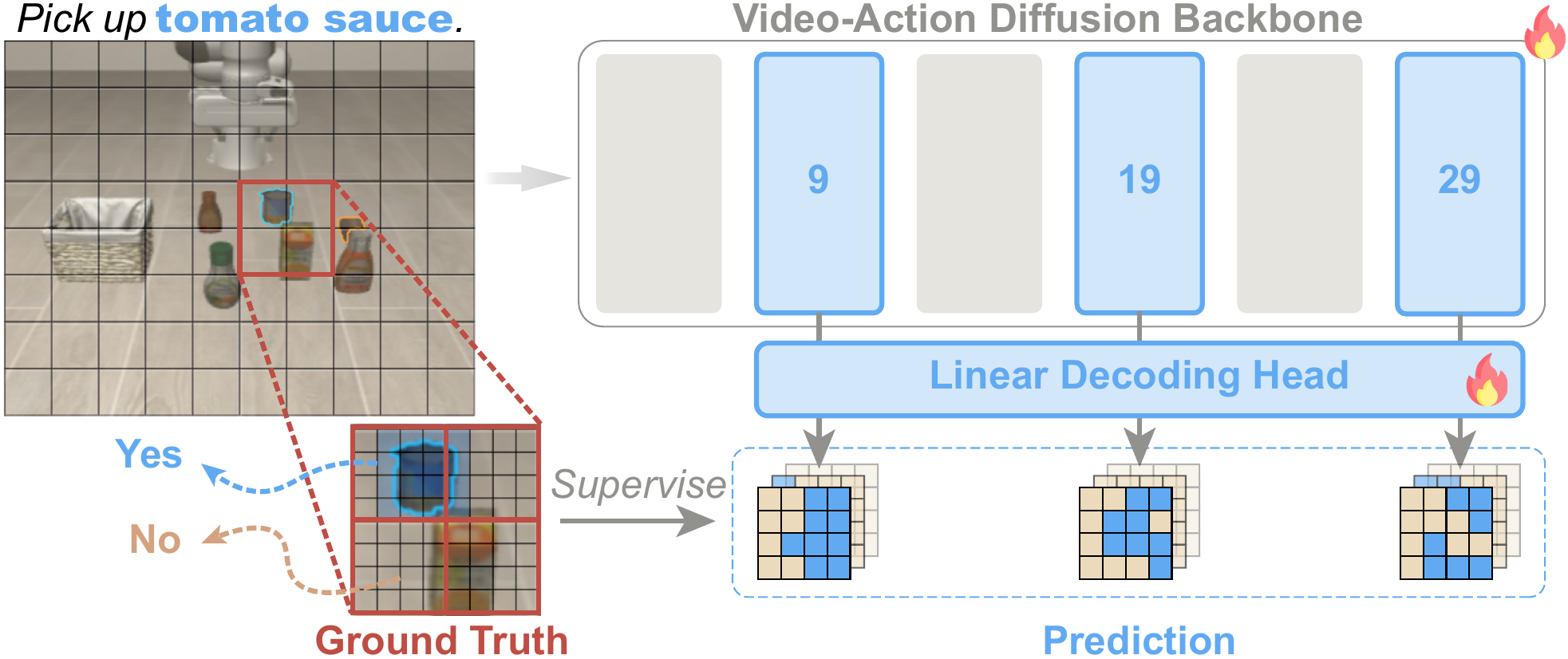}
    \caption{\textbf{Illustration of LDS.} Auxiliary linear heads at selected layers predict fine-grained grounding maps, supervised to distinguish the instructed target from other objects and background regions. The heads are discarded at inference time.}
    \label{fig:lds}
    \vspace{-0.3cm}
\end{figure}

To test this possibility, we introduce linear-decodability supervision (LDS), which uses the token-wise linear readout from Section~\ref{sec:stronger_linear_decodability} to provide auxiliary grounding supervision during training.
As in the diagnostic linear probe in Section~\ref{sec:stronger_linear_decodability}, a shared linear head reads each visual token $x^{m,\ell,c}$ independently to predict target membership in the corresponding visual region.
However, the unit of visual regions is a patch in the diagnostic probe.
A single label summarizes the whole patch even when the target covers only a little part of it, making the supervision at coarse granularity.
As illustrated in Figure~\ref{fig:lds}, LDS further splits each patch to a $4\times4$ grid and assigns labels to each cell.
Therefore, the linear heads in LDS predicts a $4\times4$ grid of logits instead of a single one for each patch.
At the first denoising step, LDS takes visual tokens from three layers chosen from each of the shallow, middel, and deep fraction of the DiT
\begin{equation}
    z_{s,j,i}^\ell=W_\ell^\top x_{s,j,i}^{m, \ell,\mathrm{out}}+b^\ell,
    \qquad z_i^\ell\in\mathbb R^{16}
\end{equation}
Here $x_{s,j,i}^{m, \ell,\mathrm{out}}$ is the block-output activation defined in Equation~\ref{eq:patch_probe}.
The head weights $W$ and bias $b$ are shared across token positions $i$, physical scenes $s$, and instructions $L_j$.

To supervise the resulting predictions, we distinguish three spatial regions: the instructed target, alternative candidate objects, and background.
We adopt a region-balanced binary classification loss
\begin{equation}
    \mathcal L_{\mathrm{aux}}
    =\frac1{|\mathcal L_s|}
    \sum_{\ell\in\mathcal L_s}
    \left(
        \lambda_1 \mathcal L_{\mathrm{target}}^{(\ell)}
        +\lambda_2 \mathcal L_{\mathrm{other}}^{(\ell)}
        +\lambda_3 \mathcal L_{\mathrm{bg}}^{(\ell)}
    \right)
\end{equation}
where $\lambda_1, \lambda_2, \text{and} \ \lambda_3$ are balancing coefficients and each regional loss is normalized by its corresponding spatial coverage.
Similar to the diagnostic linear probe in Section~\ref{sec:stronger_linear_decodability}, the objective emphasizes distinguishing the instructed target from alternative objects, while assigning a small weight to background regions to discourage spurious predictions.
The WAM is trained by jointly optimizing the original FM objective and the auxiliary loss
\begin{equation}
    \mathcal L_{\mathrm{total}}
    =\mathcal L_{\mathrm{FM}}(\mathcal D_{\mathrm{original}})
    +\lambda\mathcal L_{\mathrm{aux}}(\mathcal D_{\mathrm{ground}})
\end{equation}
Importantly, the counterfactual data provide only target-grounding supervision, without corresponding action or future-video supervision.
Note that unlike the diagnostic linear probes, we allow parameter update of both the auxiliary heads and the WAM's parameters.
Further implementation details are provided in \appref{app:lds}.
Despite supervising only intermediate instructed target representations, the auxiliary objective substantially improves downstream IF of WAMs.
On LIBERO-CF, OpenWAM's success rate increases from 5.1\% to 65.1\%, accompanied by an improvement in grounding rate from 14.4\% to 90.3\% (Table~\ref{tab:libero_cf}).

\textbf{LDS v.s. Counterfactual Data Collection.}
Although we obtained the IF WAMs for comparison through collecting and training on counterfactual demonstrations at the beginning of this section, it is worth noting that LDS serves as a more efficient and practical way to obtain IF WAMs compare to collecting counterfactual demonstrations.
Collecting counterfactual demonstrations, as in describe in Section~\ref{sec:construct}, presupposes the ability to revisit the states from which the original demonstrations were recorded.
For publicly released robot datasets, access to the recordings does not guarantee access to the original environments or reconstruction of their initial states.
When these states cannot be recreated, collecting counterfactual demonstrations is impossible.
Even when the environments remain accessible, obtaining successful trajectories for additional instructions incurs substantial collection and curation costs~\cite{DBLP:journals/ral/GlossopCBSL26}.
LDS can augment these existing datasets offline by merely pairing recorded observations with alternative instructions and target masks without collecting new demonstrations.

\section{Understanding Instruction Following of WAMs Through Competing Signals}

% \begin{figure}
%     \centering
%     \includegraphics[width=1\linewidth]{sec//figures/ratio_by_layer_compare.pdf}
%     \caption{Enter Caption}
%     \label{fig:placeholder}
% \end{figure}

% These results are surprising because linear decodable information is not guaranteed to be used by the downstream policy~\cite{DBLP:journals/tacl/ElazarRJG21,jiang2026encodedcontrolrevealinggrounding}.
% This motivates us to figure out the reason of such improvement.

The effectiveness of LDS is surprising.
Existing work~\cite{DBLP:journals/tacl/ElazarRJG21,jiang2026encodedcontrolrevealinggrounding}suggests that linearly decodable information may lie outside the subspaces to which downstream readers are sensitive.
However, the success of LDS raises the possibility that instruction-dependent information is already represented in directions that can influence downstream video-action generation.
In this section, we propose a prior-instruction competition mechanism to reconcile these observations: instruction-dependent signals can influence behavior but are outweighed by competing learned priors.
We evaluate the correctness of this mechanism by applying causal interventions derived from it.
Experiments show that both interventions substantially improve IF success rate on LIBERO-CF, providing strong evidence on the correctness of our proposed mechanism.
Finally, we revisit LDS through this mechanistic lens to explain why it can improve IF of WAMs.

\subsection{Prior-Instruction Competing Mechanism}

% We hypothesize that IF failures arise from competition between two tags in the visual representations: (1) a \emph{visual-prior tag} $P$, associated with the object favored by the model's learned behavioral experience, and (2) an \emph{instruction tag} $I$, associated with the object specified by the language instruction.

% $P$ dominants in NF WAMs and drives NF WAMs' behavior.

% Under counterfactual instructions, they favor different objects, and the weaker instruction signal may fail to redirect downstream action selection.

Consider the scene shown in Figure~\ref{fig:main}(b), where objects $A$ and $B$ are outlined in blue and orange, respectively.
We assume that $B$ is the object favored by the WAM's learned prior.
That is, an NF WAM consistently manipulates $B$ regardless of the given instruction.
We denote the representations of the patches corresponding to object $i$ in model $m$, at layer $\ell$ and site $c$, given scene $s$ and instruction $L_j$, as $x^{m,\ell,c}_{s,j,i}$.
For simplicity, we omit the model, layer, site, and scene indices and write $x_{j,i}$.
Thus, the representations of patches corresponding to objects $A$ and $B$ are denoted by $x_{j,A}$ and $x_{j,B}$, respectively.

We hypothesize two types of tags in $x_{\cdot, \cdot}$: (1) a \emph{prior tag} $P$, associated with the target favored by the model's learned behavioral experience (object $B$ in this case), and (2) an \emph{instruction tag} $I$, associated with the target $j$ specified by the language instruction $L_j$.
Then we can decompose $x_{j,i}$ under counterfactual instructions as
\begin{equation}
    \left\{
    \begin{alignedat}{2}
        x_{A,A} &= v_A + I, &\qquad x_{A,B} &= v_B + P\\
        x_{B,A} &= v_A,     &\qquad x_{B,B} &= v_B + P + I
    \end{alignedat}
    \right.
\end{equation}
where $v_A$ and $v_B$ denote the instruction-invariant visual representations (e.g., appearance).
Here $P$ and $I$ competes with eachother, and the winner controls the WAMs' policy.
Thus, the NF behavior of WAMs can be explained as the dominance of $P$.

\begin{figure}[t]
    \centering
    \includegraphics[width=\linewidth]{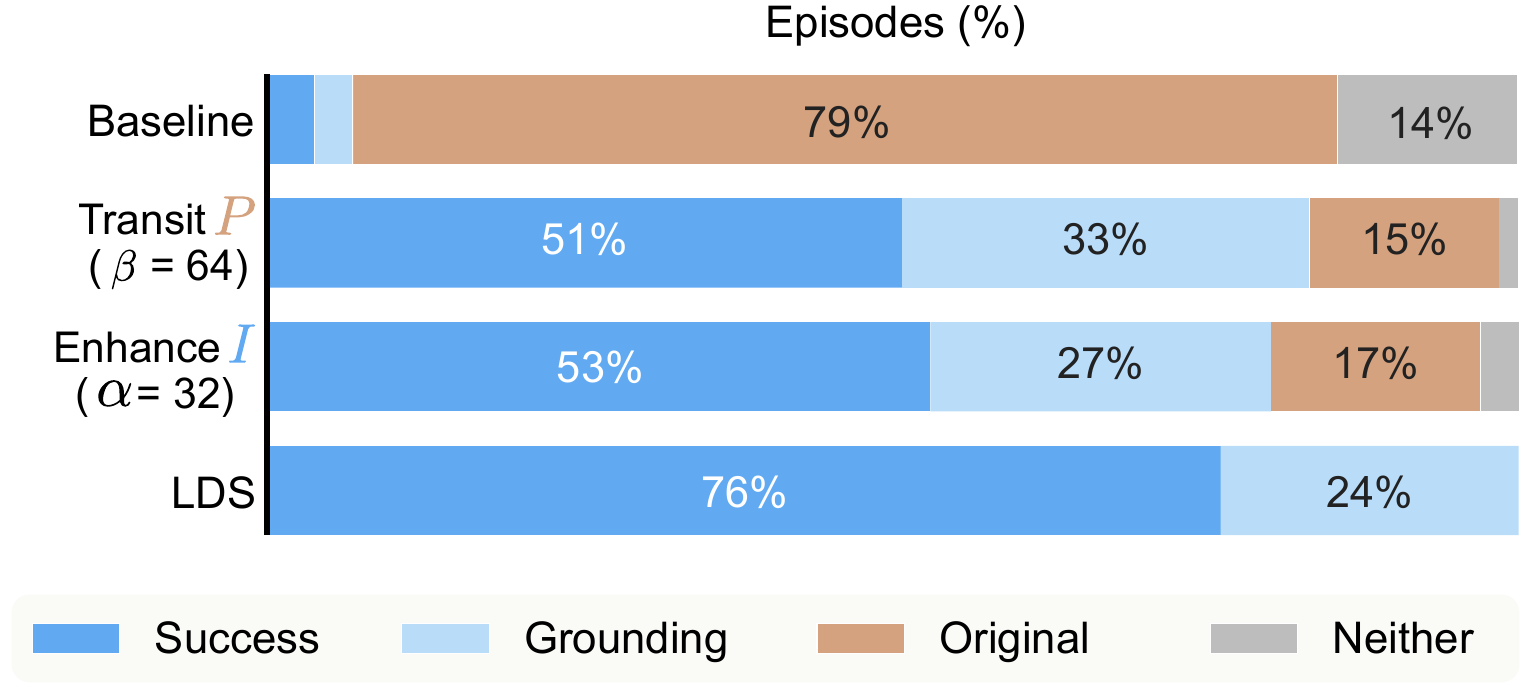}
    \caption{\textbf{Compare SR of training-free interventions to baseline and LDS on LIBERO-CF picking-subset.}}
    \label{fig:trainfree_result}
    \vspace{-0.3cm}
\end{figure}

\subsection{Evaluation via Causal Interventions}

We evaluate our hypothesized prior-instruction competing mechanism via its induced training-free interventions.
According to our hypothesis, IF failures arise from the dominance of $P$ in the competition with $I$, then two intervention strategies should redirect the NF WAM's policy under $L_A$
\begin{enumerate}
    \item \textbf{Transit $P$:} Redirect the prior tag $P$ from $B$ toward $A$, such that both $P$ and $I$ attach to the instructed target.
    \item \textbf{Enhance $I$:} Amplify the existing instruction tag $I$ in $A$, allowing it to overcome the competing prior.
\end{enumerate}
These strategies probe complementary aspects of our hypothesis.
If both strategies independently redirect behavior toward the instructed target, it provides evidence on not only both $P$ and $I$ causally influence downstream behavior but also the proposed competition mechanism.

Given the same visual observation under $L_A$ and $L_B$, we compute
\begin{equation}
    \begin{aligned}
\Delta_A
&=x_{A,A}-x_{A,B}
 =(v_A-v_B)-P+I\\
\Delta_B
&=x_{B,A}-x_{B,B}
 =(v_A-v_B)-P-I
\end{aligned}
\end{equation}
Taking their difference and average gives $I$ and $U$
\begin{equation}
I=\frac{\Delta_A-\Delta_B}{2},
\qquad
U=\frac{\Delta_A+\Delta_B}{2}
\end{equation}
where $U=(v_A-v_B)-P$.
Notably, while the $I$ can be isolated through paired counterfactual instructions, the prior tag $P$ cannot be independently separated since it is bonded on the prior object.
We therefore refer to $U$ as the prior-associated component in the following analysis.

Under $L_A$, we implement Enhance $I$ as
\begin{equation}
    \widetilde x_{A,A} = x_{A,A} +\alpha I
\end{equation}
Transit $P$ uses the unit scene-averaged Base direction $\hat U=\bar U/\|\bar U\|$:
\begin{equation}
    \widetilde x_{A,B} =x_{A,B}+\beta\hat U, \qquad
    \widetilde x_{A,A} =x_{A,A}-\beta\hat U
\end{equation}
Here, $\alpha$ and $\beta$ control intervention strength (\appref{app:openwam_tags_causal}).

As shown in Figure~\ref{fig:trainfree_result}, both interventions substantially improve IF SR on LIBERO-CF.
Enhancing $I$ increases OpenWAM's success rate from 3.8\% to 53.0\% at $\alpha=32$, while transit $P$ achieves 50.8\% at $\beta=64$.
Success of both interventions provide strong evidence on the correctness of our proposed prior-instruction competing mechansim.
The NF WAMs correctly represent instructed target, but such signal is too weak to overcome the learned prior.

\providecommand{\best}[1]{%
    \textcolor[HTML]{61AAF2}{\textbf{#1}}%
}

% Table 1: LIBERO-CF
\begin{table*}[!t]
\tiny
\centering
\setlength{\tabcolsep}{3pt}
\renewcommand{\arraystretch}{0.98}
\setlength{\aboverulesep}{0.30ex}
\setlength{\belowrulesep}{0.45ex}

\caption{
\textbf{LIBERO-CF evaluation results (\%).}
\best{Blue bold} values indicate the best result per column.
}
\label{tab:libero_cf}

\resizebox{\textwidth}{!}{%
\begin{tabular}{llll*{11}{c}}
\toprule

\multirow{2}{*}{Metric}
& \multirow{2}{*}{Type}
& \multirow{2}{*}{Model}
& \multirow{2}{*}{Setting}
& \multicolumn{2}{c}{CF-Spatial}
& \multicolumn{2}{c}{CF-Object}
& \multicolumn{2}{c}{CF-Long}
& \multicolumn{2}{c}{CF-OOD}
& \multicolumn{2}{c}{Average}
& LIBERO \\

\cmidrule(lr){5-6}
\cmidrule(lr){7-8}
\cmidrule(lr){9-10}
\cmidrule(lr){11-12}
\cmidrule(lr){13-14}

& & & 
& F.$\uparrow$ & B.$\downarrow$
& F.$\uparrow$ & B.$\downarrow$
& F.$\uparrow$ & B.$\downarrow$
& F.$\uparrow$ & B.$\downarrow$
& F.$\uparrow$ & B.$\downarrow$
& SR$\uparrow$ \\

\midrule

% ========================================================
%                    SUCCESS
% ========================================================

\multirow{16}{*}{Success}
& \multirow{10}{*}{VLA}
& \multirow{3}{*}{OpenVLA-OFT}
& Base
& 1.1 & 80.7
& 0.0 & 97.6
& 0.2 & 54.4
& 0.1 & 81.7
& 0.4 & 78.6
& \multirow{3}{*}{--} \\

& & & TF~\cite{DBLP:journals/corr/abs-2602-17659}
& 3.3 & 73.3
& 0.0 & 94.4
& 0.2 & 54.0
& 0.1 & 83.5
& 0.9 & 76.3
& \\

& & & VA~\cite{DBLP:journals/corr/abs-2602-17659}
& 7.9 & 35.2
& 0.0 & 57.0
& 0.4 & 48.9
& 0.1 & 22.7
& 2.1 & 41.0
& \\

\cmidrule(lr){3-15}

& & \multirow{3}{*}{$\pi_0$}
& Base
& 34.4 & 19.1
& 0.0 & 45.0
& 1.1 & 63.1
& 2.7 & 52.9
& 9.6 & 45.0
& \multirow{3}{*}{--} \\

& & & TF~\cite{DBLP:journals/corr/abs-2602-17659}
& 33.5 & 14.4
& 0.6 & 15.0
& 2.0 & 38.7
& 2.1 & 42.3
& 9.6 & 27.6
& \\

& & & VA~\cite{DBLP:journals/corr/abs-2602-17659}
& 36.7 & 6.1
& 0.0 & 29.2
& 2.9 & 48.7
& 2.1 & 19.2
& 10.4 & 25.8
& \\

\cmidrule(lr){3-15}

& & \multirow{4}{*}{$\pi_{0.5}$}
& Base
& 24.4 & 56.9
& 5.8 & 66.8
& 15.8 & 50.4
& 6.9 & 69.6
& 13.2 & 60.9
& \multirow{4}{*}{--} \\

& & & TF~\cite{DBLP:journals/corr/abs-2602-17659}
& 27.3 & 33.3
& 9.2 & 41.4
& 20.9 & 36.7
& 9.9 & 62.7
& 16.8 & 43.5
& \\

& & & VA~\cite{DBLP:journals/corr/abs-2602-17659}
& 31.6 & 36.0
& 18.0 & 29.8
& 26.7 & 41.8
& 10.3 & 37.1
& 21.7 & 36.2
& \\

& & & SAM~\cite{DBLP:journals/corr/abs-2606-23641}
& 55.3 & --
& 26.2 & --
& 55.4 & --
& 54.1 & --
& 47.8 & --
& \\

\cmidrule(lr){2-15}

% ---------------- WAM: SUCCESS ----------------

& \multirow{6}{*}{WAM}
& \multirow{2}{*}{InternW0-$\Delta$}
& Base
& 24.0 & 58.5
& 3.0 & 80.4
& 39.8 & 57.6
& 27.2 & 45.3
& 23.5 & 60.5
& 96.0 \\

& & & LDS (Ours)
& \best{92.8} & 3.3
& \best{86.0} & 8.8
& \best{56.4} & \best{36.4}
& 63.1 & 0.5
& \best{74.6} & 12.3
& \best{96.8} \\

\cmidrule(lr){3-15}

& & \multirow{2}{*}{LingBot-VA}
& Base
& 39.7 & 30.8
& 0.0 & 43.6
& 7.2 & 77.2
& 2.4 & 40.9
& 12.3 & 48.1
& 89.0 \\

& & & LDS (Ours)
& 54.9 & 6.8
& 0.4 & 38.0
& 17.8 & 58.2
& 46.1 & 12.7
& 29.8 & 28.9
& 88.2 \\

\cmidrule(lr){3-15}

& & \multirow{2}{*}{OpenWAM}
& Base
& 7.9 & 68.9
& 0.0 & 87.2
& 12.0 & 79.2
& 0.4 & 88.1
& 5.1 & 80.9
& 94.0 \\

& & & LDS (Ours)
& 88.3 & \best{0.4}
& 72.6 & \best{1.4}
& 31.4 & 43.8
& \best{68.3} & \best{0.1}
& 65.1 & \best{11.4}
& 89.2 \\

% ========================================================
%                  GROUNDING
% ========================================================

\midrule

\multirow{15}{*}{Grounding}
& \multirow{9}{*}{VLA}
& \multirow{3}{*}{OpenVLA-OFT}
& Base
& 6.8 & 84.7
& 2.0 & 99.6
& 4.0 & 62.9
& 6.0 & 87.3
& 4.7 & 83.6
& \\

& & & TF~\cite{DBLP:journals/corr/abs-2602-17659}
& 11.1 & 78.5
& 1.6 & 100.0
& 4.9 & 61.1
& 5.6 & 88.1
& 5.8 & 81.9
& \\

& & & VA~\cite{DBLP:journals/corr/abs-2602-17659}
& 26.5 & 59.5
& 3.8 & 91.8
& 6.9 & 60.0
& 7.9 & 50.1
& 11.3 & 65.4
& \\

\cmidrule(lr){3-15}

& & \multirow{3}{*}{$\pi_0$}
& Base
& 70.9 & 37.1
& 6.0 & 59.4
& 20.4 & 83.1
& 18.0 & 72.4
& 28.8 & 63.0
& \\

& & & TF~\cite{DBLP:journals/corr/abs-2602-17659}
& 74.5 & 32.5
& 12.2 & 50.6
& 19.6 & 81.8
& 21.1 & 72.4
& 31.9 & 59.3
& \\

& & & VA~\cite{DBLP:journals/corr/abs-2602-17659}
& 85.2 & 24.4
& 13.8 & 53.8
& 22.7 & 82.2
& 27.6 & 47.5
& 37.3 & 52.0
& \\

\cmidrule(lr){3-15}

& & \multirow{3}{*}{$\pi_{0.5}$}
& Base
& 39.3 & 61.3
& 11.4 & 68.8
& 51.6 & 58.4
& 20.7 & 74.0
& 30.8 & 65.6
& \\

& & & TF~\cite{DBLP:journals/corr/abs-2602-17659}
& 52.0 & 45.3
& 24.6 & 65.2
& 60.7 & \best{56.7}
& 24.8 & 71.9
& 40.5 & 59.8
& \\

& & & VA~\cite{DBLP:journals/corr/abs-2602-17659}
& 50.7 & 45.2
& 34.0 & 49.8
& 64.2 & 61.1
& 36.4 & 52.8
& 46.3 & 52.2
& \\

\cmidrule(lr){2-15}

% ---------------- WAM: GROUNDING ----------------

& \multirow{6}{*}{WAM}
& \multirow{2}{*}{InternW0-$\Delta$}
& Base
& 34.7 & 61.3
& 9.2 & 80.4
& 52.8 & 65.2
& 52.4 & 48.7
& 37.3 & 63.9
&  \\

& & & LDS (Ours)
& 96.4 & 8.4
& 91.4 & 9.0
& \best{69.6} & 57.8
& 99.5 & \best{2.7}
& 89.2 & \best{19.5}
&  \\

\cmidrule(lr){3-15}

& & \multirow{2}{*}{LingBot-VA}
& Base
& 59.9 & 43.5
& 7.0 & 51.8
& 40.0 & 86.8
& 21.1 & 57.3
& 32.0 & 59.9
& \\

& & & LDS (Ours)
& 86.8 & 16.0
& 9.4 & 51.2
& 67.0 & 67.8
& 74.1 & 22.7
& 59.3 & 39.4
& \\

\cmidrule(lr){3-15}

& & \multirow{2}{*}{OpenWAM}
& Base
& 20.1 & 72.9
& 1.6 & 89.2
& 30.6 & 84.8
& 5.2 & 90.4
& 14.4 & 84.3
& \\

& & & LDS (Ours)
& \best{99.9} & \best{7.6}
& \best{97.6} & \best{5.2}
& 64.2 & 61.6
& \best{99.7} & 4.0
& \best{90.3} & 19.6
& \\

\bottomrule
\end{tabular}%
}
\end{table*}

\subsection{Rethinking Linear Decodability Supervision}

We now revisit the unexpected success in Section~\ref{sec:surprising_effect}.
Consider a shared linear readout $w$ that must distinguish the all candidate objects under counterfactual instructions.
Since
\begin{equation}
    \Delta_A=U+I,\qquad
    \Delta_B=U-I
\end{equation}
correctly predicting the instructed target under both $L_A$ and $L_B$ requires
\begin{equation}
    w^\top(U+I)>0,\qquad
    w^\top(U-I)<0
\end{equation}
These conditions imply
\begin{equation}
    w^\top I>|w^\top U|
\end{equation}
Remind that $U$ remains invariant when the instruction changes, whereas $I$ transits along with the instruction.
Thus, amplifying the existing instruction-dependent component $I$ provides a direct way to increase the classification margin under both counterfactual instructions.

We compare $U$ and $I$ before and after LDS, reporting norms of scene-averaged vectors and cosines between these averages.
At OpenWAM layer 8 on held-out scenes, $\|\bar I\|$ rises from 3.4 to 14.1, while $\|\bar U\|$ decreases from 26.1 to 22.9.
Base-to-LDS cosines are 0.95 ($\bar U$) and 0.90 ($\bar I$).
This suggests that training predominantly enhances $I$ rather than creating new representations.
Complete results can be found in \appref{app:openwam_tags}.

Beyond the effect of enhancing $I$, there is still an unexplained performance gap to LDS.
This can be probably explained by adaption of out-of-distribution activations.
Specifically, Training-free interventions modify hidden representations using externally specified coefficients (e.g., $\alpha=32$), potentially moving activations away from the distribution encountered during training.
In contrast, LDS updates WAMs' parameters, allowing the downstream readers to adapt to the shifted activations.

\section{Simulation Experiments}

% Tables 2--3: one paired top float, queued on page 6 for page 7.
% Separate captions preserve table numbering and cross-references.
\begin{table*}[!t]
\centering
% Caption row: the taller caption sets a common starting height for both tables.
\begin{minipage}[t]{\columnwidth}
\caption{\textbf{LangGap evaluation results (\%).}
\best{Blue bold} values indicate the best result per column.}
\label{tab:langgap}
\end{minipage}\hfill
\begin{minipage}[t]{\columnwidth}
\vspace{0pt}
\caption{RoboTwin-IF evaluation results (\%).
\best{Blue bold} values indicate the best result per column.}
\label{tab:robotwin_if}
\end{minipage}
\par\vspace{2pt}
% Table row: common type size and rule style, with balanced overall heights.
\begin{minipage}[t]{\columnwidth}
\vspace{0pt}
\centering
\fontsize{8}{9.6}\selectfont
\setlength{\tabcolsep}{2pt}
\renewcommand{\arraystretch}{1.20}
\begin{tabular*}{\linewidth}{@{\extracolsep{\fill}}lllccccc@{}}
\toprule

\multirow{2}{*}{Type}
& \multirow{2}{*}{Model}
& \multirow{2}{*}{Setting}
& \multicolumn{4}{c}{LangGap}
& LIBERO \\

\cmidrule(lr){4-7}
\cmidrule(lr){8-8}

& &
& Spatial
& Object
& Goal
& Avg.
& SR$\uparrow$ \\

\midrule

% ========================================================
% VLA
% ========================================================

\multirow{5}{*}{VLA}

& \multirow{5}{*}{$\pi_{0.5}$}
& Base
& 5.9 & 37.7 & 30.0 & 24.5 & -- \\

& & CFG~\cite{DBLP:journals/corr/abs-2207-12598}
& 8.5 & 35.8 & 43.8 & 29.4 & -- \\

& & LangGap~\cite{DBLP:journals/corr/abs-2603-00592}
& 10.2 & 37.0 & 27.2 & 24.8 & -- \\

& & LoRA~\cite{DBLP:journals/corr/abs-2606-23641}
& 0.2 & 0.3 & 0.2 & 0.2 & -- \\

& & SAM~\cite{DBLP:journals/corr/abs-2606-23641}
& 19.7 & 48.7 & 56.7 & 41.7 & -- \\

\midrule

% ========================================================
% WAM
% ========================================================

\multirow{6}{*}{WAM}

& \multirow{2}{*}{InternW0-$\Delta$}
& Base
& 6.4 & 42.7 & \best{63.3} & 37.5 & \best{96.0} \\

& & LDS (Ours)
& 57.5 & \best{94.5} & 62.2 & \best{71.4} & \best{96.0} \\

\cmidrule(lr){2-8}

& \multirow{2}{*}{LingBot-VA}
& Base
& 8.6 & 35.9 & 4.4 & 16.3 & 89.0 \\

& & LDS (Ours)
& 18.2 & 36.4 & 10.0 & 21.5 & 89.8 \\

\cmidrule(lr){2-8}

& \multirow{2}{*}{OpenWAM}
& Base
& 1.1 & 34.1 & 24.4 & 19.9 & 94.0 \\

& & LDS (Ours)
& \best{62.9} & 73.6 & 45.6
& 60.7 & 91.2 \\

\bottomrule
\end{tabular*}
\end{minipage}\hfill
\begin{minipage}[t]{\columnwidth}
\vspace{0pt}
\centering
\fontsize{8}{9.6}\selectfont
\setlength{\tabcolsep}{2pt}
\renewcommand{\arraystretch}{1.08}
\setlength{\tabcolsep}{1.8pt}
\begin{tabular*}{\linewidth}{@{\extracolsep{\fill}}lllcccc@{}}
\toprule
\multirow{2}{*}{Metric} & \multirow{2}{*}{Model} & \multirow{2}{*}{Setting}
& Pick & Placement & Avg. & Traj.-sup. \\
\cmidrule(lr){4-4}\cmidrule(lr){5-5}\cmidrule(lr){6-6}\cmidrule(lr){7-7}
& & & Rate$\uparrow$ & Rate$\uparrow$ & Rate$\uparrow$ & SR$\uparrow$ \\
\midrule
\multirow{6}{*}{Success} & \multirow{2}{*}{InternW0-$\Delta$} & Base & 0.0 & 18.0 & 9.0 & 90.8 \\
 &  & LDS (Ours) & 61.6 & 33.2 & 47.4 & 92.0 \\
\cmidrule(lr){2-7}
 & \multirow{2}{*}{LingBot-VA} & Base & 4.8 & 14.4 & 9.6 & 78.8 \\
 &  & LDS (Ours) & 34.8 & 13.2 & 24.0 & 78.6 \\
\cmidrule(lr){2-7}
 & \multirow{2}{*}{OpenWAM} & Base & 6.0 & 9.2 & 7.6 & \best{99.0} \\
 &  & LDS (Ours) & \best{71.2} & \best{37.6} & \best{54.4} & 97.6 \\
\midrule
\multirow{6}{*}{Grounding} & \multirow{2}{*}{InternW0-$\Delta$} & Base & 5.6 & 18.0 & 11.8 &  \\
 &  & LDS (Ours) & 90.4 & 34.0 & 62.2 &  \\
\cmidrule(lr){2-7}
 & \multirow{2}{*}{LingBot-VA} & Base & 10.0 & 16.0 & 13.0 &  \\
 &  & LDS (Ours) & 88.4 & 20.0 & 54.2 &  \\
\cmidrule(lr){2-7}
 & \multirow{2}{*}{OpenWAM} & Base & 6.4 & 15.6 & 11.0 &  \\
 &  & LDS (Ours) & \best{99.6} & \best{52.8} & \best{76.2} &  \\
\bottomrule
\end{tabular*}
\end{minipage}
\end{table*}

% RQ1: Does linear-probe supervision improve instruction following?
% RQ2: Does it support grounding and execution of placement destinations?
% RQ3: How does performance extend beyond localization-supervised initial states and scenes?

We evaluate whether LDS improves instruction following in varies of tasks, embodiment, and visual domains (RQ1), supports grounding and execution of new placement goals (RQ2), and generalizes to initial states and scenes excluded from grounding supervision (RQ3).

\subsection{Settings}

\textbf{Benchmarks.}
We evaluate instruction following on three complementary simulation benchmarks: LIBERO-CF~\cite{DBLP:journals/corr/abs-2602-17659}, LangGap~\cite{DBLP:journals/corr/abs-2603-00592}, and RoboTwin-IF (\appref{app:robotwin_if}).
LIBERO-CF and LangGap share LIBERO's visual domain and third-person scene-camera view.
LIBERO-CF assigns alternative feasible goals to LIBERO scenes, testing whether the policy follows the new instruction or repeats the original training task.
LangGap introduces 59 tasks that vary the manipulated object, placement destination, spatial reference, or action within fixed scene layouts.
Our RoboTwin-IF (\appref{app:robotwin_if}) uses RoboTwin's distinct visual assets and first-person head-camera view in a dual-arm setting.
We evaluate 25 Pick and 25 Placement scenes with paired instructions that share identical initial observations but specify different target objects or destinations, respectively.

\textbf{Training.}
% Avoid a single trailing Training line below the page-7 table pair.
% \looseness=-1
For LIBERO-CF and LangGap, we use the released LIBERO checkpoints of OpenWAM, LingBot-VA, and InternW0-$\Delta$ as Base models.
For RoboTwin-IF, we randomly select one of the two instructions in each evaluated scene and fine-tune the released RoboTwin checkpoints on the corresponding trajectories to obtain Base models.
We then fine-tune each Base model using the linear-probe supervision objective introduced in Section~\ref{sec:surprising_effect} for 1024 steps.
We jointly optimize the native flow-matching loss and an auxiliary grounding loss weighted by $\lambda=0.25$.
We take layer $\ell \in \{9, 19, 29\}$ and $(\lambda_1, \lambda_2, \lambda_3) = (0.5, 0.45, 0.05)$.
% We use separate optimizers for the WAM parameters and the auxiliary linear heads.
% The WAM parameters are updated using the combined loss with a learning rate of $10^{-5}$.
% The auxiliary linear heads are updated only through the grounding loss with a learning rate of $3\times10^{-4}$.
% Trajectory supervision is restricted to the original trajectory training set, while grounding supervision also covers instructions without trajectory demonstrations.
% Grounding supervision is applied to the grasp target or placement destination in the first input frame.

\textbf{Evaluation.}
Success measures completion of the evaluated instruction.
Grounding measures whether the gripper reaches the target object on LIBERO-CF and Pick-subset in RoboTwin-IF.
For Placement, Grounding measures whether the grasped object reaches and remains in the instructed destination region.
LIBERO-CF evaluates 50 tasks with 50 initial states each and scores the instructed goal (Faithful, F.) and the original training goal (Biased, B.).
LangGap uses 10 initial states per extended task, and RoboTwin-IF uses 10 episodes per instruction in each scene.
We also evaluate the 40 original LIBERO tasks with 10 episodes each to assess retention of the original task.

\subsection{RQ1: Unlocking Instruction Following}

% Table 4: scene hold-out coverage
\begin{table*}[!t]
\centering
\small
\caption{\textbf{Scene hold-out evaluation.} Rates pool episodes within each group.
Blue bold values indicate the best result per column.}
\label{tab:scene_holdout}
\begin{tabular}{lccccccccc}
\toprule
\multirow{3}{*}{Model} & \multicolumn{6}{c}{LIBERO-CF} & \multicolumn{3}{c}{LangGap} \\
\cmidrule(lr){2-7}\cmidrule(lr){8-10}
 & \multicolumn{3}{c}{F.\ success$\uparrow$} & \multicolumn{3}{c}{B.\ success$\downarrow$} & \multicolumn{3}{c}{Success$\uparrow$} \\
\cmidrule(lr){2-4}\cmidrule(lr){5-7}\cmidrule(lr){8-10}
 & Sup. & Unsup.\ init & Unsup.\ scene & Sup. & Unsup.\ init & Unsup.\ scene & Sup. & Unsup.\ init & Unsup.\ scene \\
\midrule
InternW0-$\Delta$ & \best{71.8} & \best{79.1} & \best{67.7} & 14.5 & 8.1 & \best{9.4} & \best{69.1} & \best{76.1} & \best{45.8} \\
LingBot-VA & 26.5 & 32.0 & 27.4 & 31.6 & 23.1 & 24.1 & 14.8 & 24.8 & 25.0 \\
OpenWAM & 61.5 & 66.9 & 49.6 & \best{13.7} & \best{4.8} & 32.0 & 65.4 & 64.7 & 39.2 \\
\bottomrule
\end{tabular}
\end{table*}

On LIBERO-CF, LDS increases average faithful success and Grounding and reduces biased success across all WAMs (Table~\ref{tab:libero_cf}).
This shift supports better instructed target selection.
InternW0-$\Delta$ w/ LDS leads the leaderboard, achieving 26.8 percentage points of faithful success and 42.9 percentage points above the strongest baseline.
These gains extend to LangGap's semantic changes, as shown in Table~\ref{tab:langgap}.
InternW0-$\Delta$ w/ LDS achieves the highest average success, 29.7 percentage points above SAM.
Across both benchmarks, all WAMs w/ LDS largely retain performance of their Base counterparts on the original LIBERO tasks.

RoboTwin-IF results (Table~\ref{tab:robotwin_if}) extend these gains to a different visual domain, camera viewpoint, and embodiment.
All WAMs improve average success and Grounding on instructions without trajectory demonstrations.
OpenWAM w/ LDS gains 46.8 and 65.2 percentage points in average success and Grounding, respectively.
Together, these results support the effectiveness of LDS across models and environments, consistent with our prior-instruction competition mechanism that strengthening $I$ can counteract learned behavioral priors.

% On LIBERO-CF, LDS improves instruction following across all evaluated WAMs (Table~\ref{tab:libero_cf}).
% Average faithful success and grounding increase, while biased success decreases.
% These opposing changes within in-distribution scenes support better instructed target selection.
% InternW0-$\Delta$ w/ LDS achieves the highest average faithful success, 26.8 percentage points above the best baseline; while OpenWAM w/ LDS reaches the best Grounding rate of 90.3\%, outperforming the highest baseline by 44 percentage points.

% These gains extend to LangGap's semantic changes, as shown in Table~\ref{tab:langgap}.
% InternW0-$\Delta$ w/ LDS achieves the highest average success, 29.7 percentage points above SAM.
% Across both benchmarks, all WAMs w/ LDS largely retain performance of their Base counterparts on the original LIBERO tasks.

% Results on RoboTwin-IF (Table~\ref{tab:robotwin_if}) demonstrate the effectiveness of LDS across different visual domain, camera view, and embodiment.
% All evaluated WAMs improve dramatically in both target contact and task completion for instructions without trajectory demonstrations.
% OpenWAM reaches an improvement on Success and Grounding rate of 46.8 and 65.2 percentage points, respectively.
% These results indicates LDS and Prior-Instruction Competition Mechanism generally holds in a wide range of settings.

\subsection{RQ2: Supervision on Placement Destinations}

LDS extends target grounding from grasp targets to placement destinations.
On both LangGap and RoboTwin-IF, InternW0-$\Delta$ w/ LDS and OpenWAM w/ LDS consistently outperform their Base counterparts on these tasks (Tables~\ref{tab:langgap} and~\ref{tab:robotwin_if}).
For example, OpenWAM w/ LDS improves success over Base by 57.7 percentage points on LangGap's 13 destination-changing tasks and by 28.4 points on RoboTwin-IF Placement.
Together with the Pick results, these gains support an instruction-defined target concept encompassing both objects and destinations, and suggest a shared prior-instruction competition mechanism for selecting either type of target.

LingBot-VA w/ LDS shows limited transfer to placement tasks.
On RoboTwin-IF Placement, destination Grounding improves from 16.0\% to 20.0\%, but task success decreases from 14.4\% to 13.2\%.
Examining the grasping stage helps explain this discrepancy.
The fraction of episodes with valid grasp acquisition drops from 95.2\% for Base to 64.4\% with LDS.
% This loss in grasp acquisition limits the end-to-end benefit of better destination selection.
We hypothesize that LingBot-VA's shared FFN parameters tightly couple video and action representations, making grasp execution sensitive to updates intended to improve target grounding.

\subsection{RQ3: Generalization Capability}

We evaluate whether WAMs w/ LDS follow instructions beyond the states and scenes used for grounding supervision on LIBERO-CF and LangGap.
Held-out initial states (Unsup.\ init) use object poses absent from grounding supervision for otherwise supervised tasks and scenes.
For held-out scenes (Unsup.\ scene), all grounding labels from a physical scene are removed.
We train LDS models on the same split across model families.
\appref{app:scene_holdout} gives split details and standardized rates.

As shown in Table~\ref{tab:scene_holdout}, on held-out initial states, InternW0-$\Delta$ w/ LDS scores 79.1\% on LIBERO-CF and 76.1\% on LangGap, versus 71.8\% and 69.1\% on supervised states.
Instruction following thus extends beyond supervised configurations, although group task compositions differ (\appref{app:scene_holdout}).
Held-out scenes are more challenging: InternW0-$\Delta$ w/ LDS leads with 67.7\% and 45.8\% success on LIBERO-CF and LangGap, respectively; OpenWAM w/ LDS reaches 49.6\% and 39.2\%.
One possible difficulty is grounding object identities absent from LDS supervision.

% Together with our representation analysis and training-free interventions, these results support the view that LDS strengthens the influence of the instruction tag $I$ on downstream target selection, making behavior more responsive to the instruction.

\begin{figure}[t]
    \centering
    \includegraphics[width=1\linewidth]{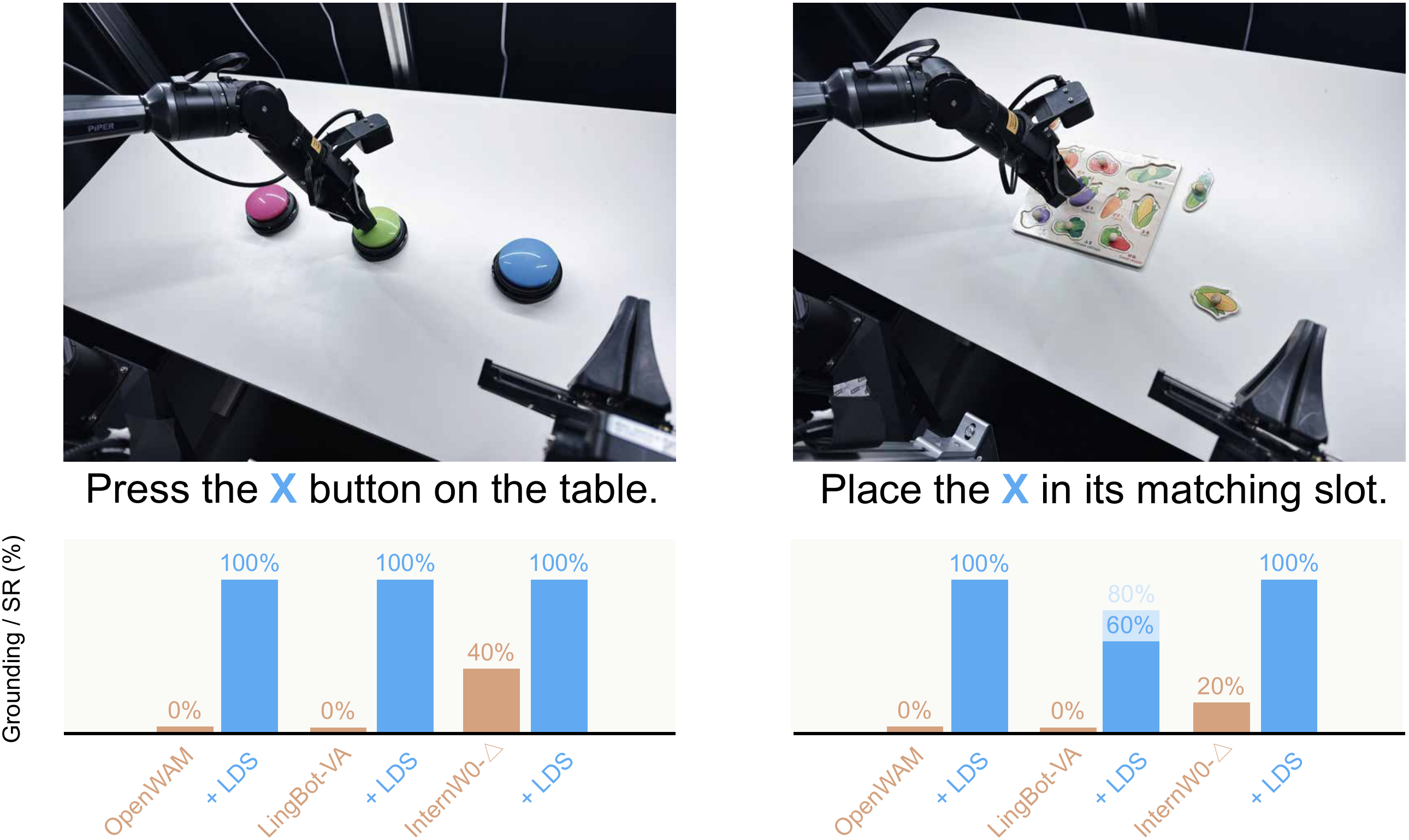}
    \caption{Real-world instruction following evaluation.}
    \label{fig:real_world}
    \vspace{-0.3cm}
\end{figure}

\section{Real-World Experiments}
\label{sec:real_world}

% A short section lead avoids keeping two consecutive headings together
% across the column break after Figure 5.
We test whether LDS improves instruction following when the trajectory-training data contain explicit visual shortcuts.
We compare Base and LDS variants of OpenWAM, LingBot-VA, and InternW0-$\Delta$ on an AgileX Cobot Magic dual-arm robot using Press-Button and Place-in-Slot (Figure~\ref{fig:real_world}).

\subsection{Settings}

Both variants use the same biased trajectory data; LDS adds first-frame target-localization supervision.
Each Base or LDS variant is evaluated for 10 trials per task.
For \textbf{Press-Button}, there are red, green, and blue buttons on the table.
Training data only contains trajectory of pressing the red button.
At test-time, $X\in\{\text{green},\text{blue}\}$ and we test five trials for each color.
For \textbf{Place-in-Slot}, training data come from recordings with fixed piece combinations and placement orders.
Consequently, the visible set of remaining pieces strongly predict which piece is placed next.
At test time, $X$ is randomly selected on each trial from \emph{carrot}, \emph{eggplant}, \emph{broccoli}, \emph{corn}, and \emph{cucumber}.
Data preparation and annotation details are provided in \appref{app:real_robot}.

\subsection{Performance}

Figure~\ref{fig:real_world} shows that LDS improves instruction following across all three WAMs despite the shortcuts in their trajectory data.
On Press-Button, OpenWAM and LingBot-VA Base both achieve 0\% success, while InternW0-$\Delta$ Base achieves 40\%.
With LDS, all three models reach 100\% success on the green and blue targets, neither of which has a trajectory demonstration in the training set.

On Place-in-Slot, Base success is 0\% for OpenWAM and LingBot-VA and 20\% for InternW0-$\Delta$.
OpenWAM w/ LDS and InternW0-$\Delta$ w/ LDS both reach 100\% success; LingBot-VA w/ LDS reaches 80\% Grounding and 60\% success.
% The gap for LingBot-VA distinguishes successful target acquisition from completion of the insertion, indicating that correct object selection does not guarantee successful execution.
Together, these results show that LDS can make real-robot behavior more responsive to the instruction while retaining the same biased training data.

\section{Conclusion}
\label{sec:conclusion}

We study why WAMs encode instructed targets yet act on learned visual priors.
Our prior-instruction competition mechanism explains how an existing instruction signal can be outweighed by a stronger prior signal in downstream target selection.
Training-free interventions that amplify the instruction signal or redirect the prior-associated component provide causal support for this account.
We also propose LDS and explain why it work through the lens of our proposed mechansim.
Simulation and real-world experiments demonstrate improved target grounding with generalization.

% Together, these findings connect a mechanistic understanding of instruction-following failures with a practical means of improving instruction-driven behavior.
% Extending this approach to long-horizon, compositional instructions is a promising next step.

% The arXiv version is a single linked PDF with continuous page numbering.
\clearpage
\appendix
\numberwithin{equation}{section}
% main.tex starts the appendix on a new page with continuous page numbering.
% Begin directly with Appendix A; the combined paper needs no repeated title.
\hypertarget{supplement.start}{}

% =====================================================================
%  Appendix: RoboTwin-IF
%  Required packages (preamble):
%    \usepackage{graphicx,booktabs,longtable,array,multirow,xcolor,amsmath,amssymb}
%  Figures are expected under figures/ (generated by make_figures.py).
%  Task rows are generated verbatim from the evaluation records by
%  make_task_tables.py.
% =====================================================================
\definecolor{rtifblue}{HTML}{61AAF2}
\definecolor{rtifsand}{HTML}{D4A27F}
\newlength{\rtiftaskwidth}

\section{The RoboTwin-IF Benchmark}
\label{app:robotwin_if}

RoboTwin-IF is an instruction-following (IF) benchmark and training corpus for
bimanual manipulation. It is built on top of the RoboTwin~2.0
simulation platform~\citep{DBLP:journals/corr/abs-2506-18088}. Because RoboTwin-IF has not yet
been released, this appendix describes how it is constructed, reports dataset
statistics, and lists the task configuration we use to evaluate LDS
in the main text.

\subsection{Motivation and Design Principle}
\label{app:robotwin_if_motivation}

\begin{figure*}[t]
\centering
% Respect the original Illustrator artwork bounds and retain vector content.
\includegraphics[width=\textwidth,pagebox=artbox]{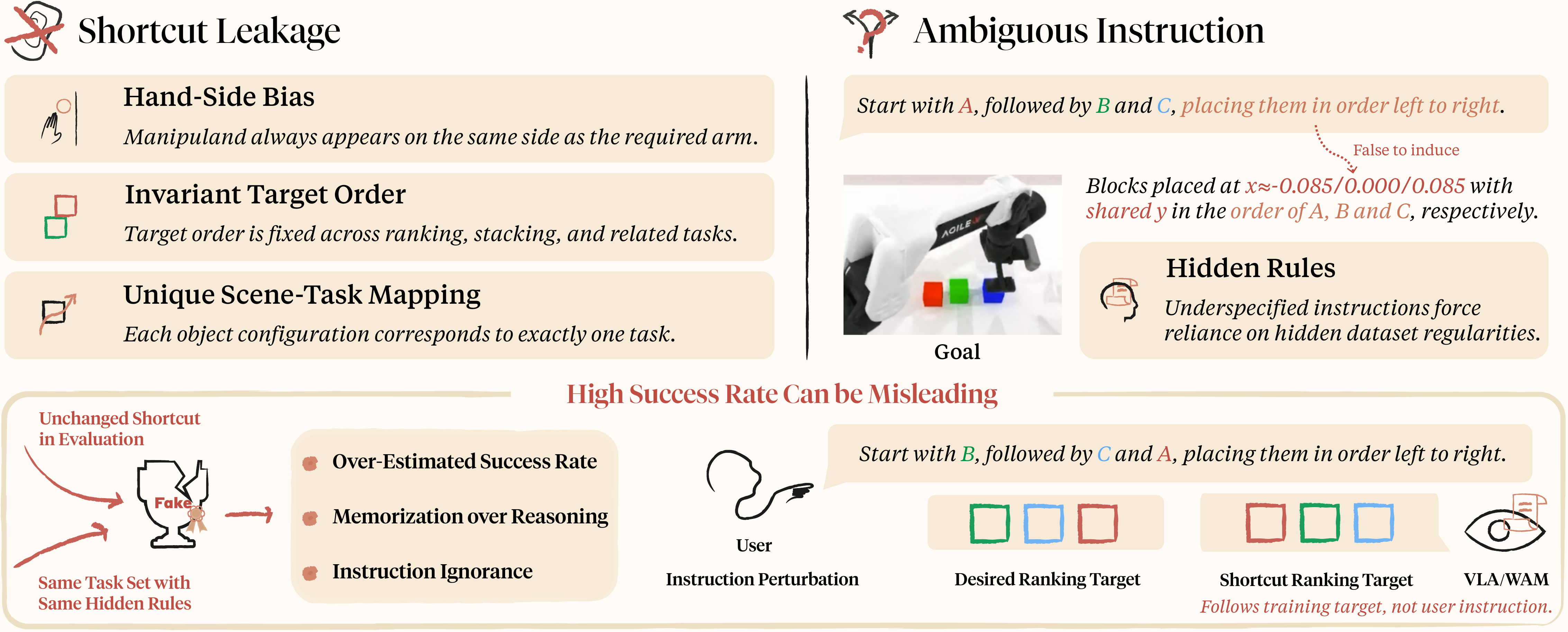}
\caption{\textbf{Motivation for RoboTwin-IF.}
Hand-side bias, fixed target orders, and one-to-one scene--task mappings can
provide shortcuts, while underspecified instructions encourage reliance on
hidden dataset rules. When evaluation preserves these regularities, high
success rates can conceal a failure to respond to instruction changes, as
illustrated by the ordering example. RoboTwin-IF tests instruction following
by holding the initial observation fixed while varying the instructed goal.}
\label{fig:rtif_motivation}
\end{figure*}

In most manipulation benchmarks, the initial observation largely determines the
task. In a standard pick-and-place episode there is usually only one plausible
object to grasp and one plausible receptacle. A policy can therefore achieve a
high success rate while ignoring most of the instruction, and its success rate
does not tell us whether it actually grounds language.
Figure~\ref{fig:rtif_motivation} illustrates how such shortcuts and
underspecified instructions can hide failures of instruction following.
RoboTwin-IF instead
follows the principle
\begin{center}
\emph{different goals, same initial observation.}
\end{center}
Every RoboTwin-IF scene is built so that several distinct goals are
simultaneously legal and physically achievable from the same initial state,
including the same robot configuration. Only the instruction distinguishes
between these goals. Two instructions that apply to the same scene and differ
in exactly one controlled semantic factor form a \emph{counterfactual minimal
pair}. Because the two episodes share pixels, proprioception, and object
poses, any systematic behavioral difference must come from the language input.

\paragraph{Platform.}
All scenes are simulated in SAPIEN with the RoboTwin~2.0 \texttt{aloha-agilex}
embodiment. This is a dual-arm robot with two 6-DoF arms and parallel-jaw
grippers. Its state and action space is a 14-dimensional joint-position
vector, with six joints and one normalized gripper opening per arm. The
observations are synchronized RGB streams from one head camera and two wrist
cameras, each at $320\times240$ resolution. Physics is integrated at
250\,Hz, and observations and actions are recorded at 25\,Hz. We use the
clean RoboTwin~2.0 rendering profile, without domain randomization, so that
counterfactual pairs differ only in their instructions.

\begin{figure}[t]
\centering
\includegraphics[width=\linewidth]{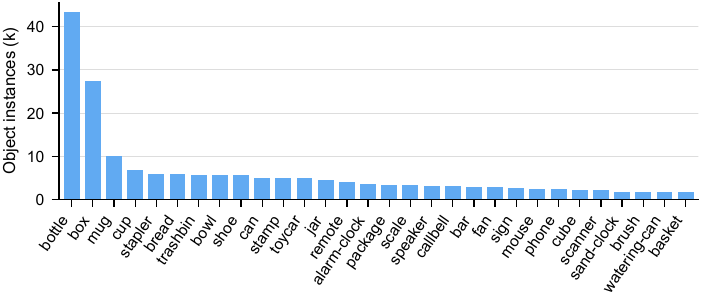}
\caption{The 30 most frequent object categories, counted as object instances
over all solved SceneSpecs; the scenes contain 75 categories in total.}
\label{fig:rtif_categories}
\end{figure}

\paragraph{Object library.}
RoboTwin-IF draws from the RoboTwin~2.0 object library: 125 asset classes
with 575 object models. We review every model by hand and annotate its
attributes, giving 2{,}386 reviewed descriptors over 552 models. The
descriptors cover 101 object types, 13 colors, 3 size classes, 8 shapes, and
6 materials. Every model is checked in simulation before it is used. A
canonical-skill validation stage checks 796 asset--model--skill
configurations with 2{,}544 physical attempts. A pair-validation stage
certifies 11{,}285 subject--receiver terminal pairs and 4{,}114 executable
skill pairs for the \textit{in}, \textit{on}, and \textit{stack} relations. In
total, the released scenes contain 95 asset classes, 434 distinct object
models, and 75 semantic categories (Fig.~\ref{fig:rtif_categories}).

\paragraph{Instruction-following axes.}
RoboTwin-IF organizes language variation into six \emph{axes}. Each axis names
the single factor that differs between the two members of a minimal pair
(Table~\ref{tab:rtif_axes}). Axes D1a--D1c vary the \emph{referring
expression}, that is, which object is manipulated. D2a and D2b vary the
\emph{goal}, that is, which state or spatial relation must hold at the end of
the episode. D4 varies the \emph{embodiment}, that is, which arm must perform
the action. Each axis takes values from a closed vocabulary of 174
\emph{cells} in total. An abstract scene family can combine at most one D1 axis
and at most one D2 axis, with or without D4, all within a single manipulation
step. This rule gives 23 legal axis signatures (6 single-axis, 11 two-axis,
and 6 three-axis).

\begin{table*}[t]
\centering
\caption{Instruction-following axes in RoboTwin-IF. Each row lists the factor
that changes within a minimal pair; all other aspects of the goal and the
initial scene are held fixed. ``Cells'' counts the declared values of each
axis.}
\label{tab:rtif_axes}
\small
\setlength{\tabcolsep}{4pt}
\begin{tabular*}{\textwidth}{@{\extracolsep{\fill}}l l p{0.64\textwidth} r@{}}
\toprule
Code & Name & Values (examples) & Cells \\
\midrule
D1a & Attribute reference & type (101), color (13), size (3), shape (8), material (6): \textit{``the \underline{white} shoe''} & 131 \\
D1b & Spatial reference & left/right/front/behind of a landmark, quadrants, \textit{in}, \textit{on}: \textit{``the remote \underline{left of} the microphone''} & 10 \\
D1c & Ordinal reference & leftmost/rightmost, smallest/largest, nearest/farthest to the robot or to a landmark & 8 \\
D2a & Object state & open/close, on/off, signed in-plane rotation $\in\{\pm30^\circ,\pm45^\circ,\pm60^\circ,\pm90^\circ\}$ & 12 \\
D2b & Terminal relation & \textit{in}, \textit{on}, stack, left/right/front/behind, between, left/right of a pair & 11 \\
D4 & Arm choice & left arm vs.\ right arm & 2 \\
\bottomrule
\end{tabular*}
\end{table*}

\subsection{Counterfactual Data Generation Pipeline}
\label{app:robotwin_if_pipeline}

\begin{figure}[t]
\centering
\includegraphics[width=\linewidth]{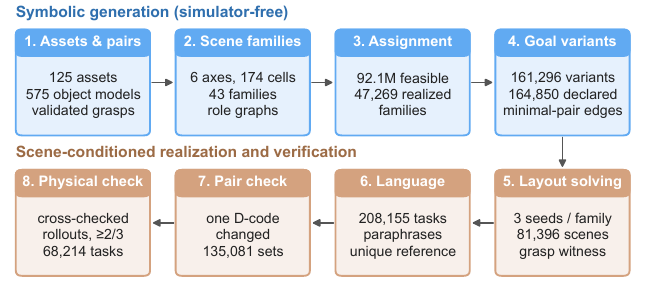}
\caption{The RoboTwin-IF counterfactual generation pipeline. Stages~1--4 are
symbolic and simulator-free; stages~5--8 realize, verify, and physically
validate concrete scenes. The counts are from the released generation run.}
\label{fig:rtif_pipeline}
\end{figure}

Many earlier generators are \emph{recipe-first}: they sample a task template
and then fill its slots one at a time. Slot-by-slot sampling breaks the
counterfactual guarantees we need. For example, replacing a ``red mug'' with
another red object can make the descriptor non-unique, change grasp
feasibility, or invalidate the success checker. RoboTwin-IF is instead
\emph{constraint-first}. All semantic, geometric, and physical requirements of
a contrast are written as an explicit constraint set. Scenes are then solved
from these constraints, and every stage either proves that its output
satisfies them or records a structured exclusion. Figure~\ref{fig:rtif_pipeline}
summarizes the pipeline, which has eight stages.

\paragraph{1. Asset and pair evidence.}
Every object model has reviewed attributes, a canonical resting pose, and
grasp contact annotations. In simulation we verify which skills each model
supports, such as left-arm or right-arm grasping, articulation, pressing,
clicking, and reorientation. We also verify which subject--receiver pairs can
settle stably in each terminal relation. Downstream stages may only use
capabilities that have this physical evidence.

\paragraph{2. Abstract scene families.}
An \emph{abstract scene family} pairs an axis signature with an archetype and
expands into a \emph{role graph}. The nodes of the role graph are abstract
roles, such as a subject, its contrast peers, a landmark, a receiver, or
anchors. Its edges are flat, typed constraints, for example: the descriptor
must be unique among the peers; the subject must be graspable by the specified
arm; the receiver must accept the subject in the requested relation. Each
family also declares its \emph{goal-variant schema}, which is the set of goals
the scene must support, and the minimal-pair edges between those goals. The
current release defines 43 families.

\paragraph{3. Combination-level assignment.}
A finite-domain solver enumerates complete role-to-model assignments that
satisfy every semantic, asset, and pair constraint \emph{jointly}. It never
samples slot by slot. Across the 37 families with a non-empty domain there are
$9.2\times10^{7}$ feasible assignments. We keep at most 2{,}048 assignments
per family, chosen deterministically so that every marginal is covered:
role--model, role--category, attribute value, and pair relation. This yields
47{,}269 \emph{realized families}.

\paragraph{4. Goal realization.}
Instantiating the variant schema of each realized family gives typed concrete
goals. The release contains 161{,}296 goal variants, linked by 164{,}850
declared minimal-pair edges. Each goal is stored in a structured intermediate
representation, from which the instruction, the success checker, and a
symbolic skill plan are compiled.

\paragraph{5. Unit building and layout solving.}
Objects that must stay together, such as a subject and the receiver it rests
in, are grouped into rigid \emph{units}. A solver then places these units on
the support plane under reachability, non-overlap, camera-visibility, and
reservation constraints. For every goal variant, an exact-mesh motion planner
must certify a collision-free grasp-and-lift for the arm that variant requires.
\textbf{The layout is sampled once per scene and shared by all goal
variants.} Choosing a variant never moves any object, so no variant-specific
visual cue can leak the goal. Each realized family gets three independent
layout seeds, giving 81{,}396 solved \emph{SceneSpecs}. We also audit the
dataset for layout shortcuts. For example, no single extreme position of the manipulated
object (leftmost or rightmost) accounts for more than 43\% of goal variants,
and no single reachability zone accounts for more than 44\%.

\paragraph{6. Instructions and checkers.}
For each goal we write three or four natural-language paraphrase templates and
sample one with a seed. We check that the resulting reference picks out
exactly one object, using the reviewed nouns and attributes of every object in
the scene. Goals that fail this check (1{,}056 tasks, 0.5\%) are excluded
rather than repaired. Each task also receives an executable checker, expressed
in the same goal representation, relative to the shared initial state. This
stage produces 208{,}155 task records.

\paragraph{7. Minimal-pair verification.}
For every declared edge we decompose both goals into typed D1a/D1b/D1c/D2a/D2b/D4
projections and an invariant skeleton. The edge is accepted only if the two
goals share the same SceneSpec and initial observation, differ in
\emph{exactly one} D-code, and have no semantic difference outside that code.
Valid pairs are grouped into \emph{contrast sets}, which are complete cliques
of $K$ goals that differ only along one axis. We find $K=2$ for most families,
$K=3$ for the two-anchor D2b family (between, left of the pair, or right of
the pair), and $K=4$ for D1b quadrants. Of 158{,}628 declared edges,
157{,}668 are valid, and they form 135{,}081 complete contrast sets.

\paragraph{8. Physical validation and cross-checking.}
Each task is executed in simulation with a scene-conditioned scripted expert.
Before execution, a \emph{descriptor gate} renders the shared initial head
camera view and requires each referenced object to be perceptually resolvable.
The gate requires a mask area of at least 60 pixels and a bounding-box fill of
at least 0.12. For spatial and ordinal references it also requires an image
separation of at least 8 pixels, and for depth relations a depth difference
of at least 10\,mm. Size references require a projected area ratio of at
least 1.12. After each rollout we evaluate the \emph{checkers of every goal in
the scene}, not just the instructed one, and classify the outcome as
\emph{intended} (only the instructed checker holds), \emph{other} (a different
goal's checker holds), \emph{ambiguous} (several checkers hold), or
\emph{failure} (no checker holds). Every checker also includes a
scene-integrity guard: objects that the goal does not mention may move by at
most 4\,mm and $3^\circ$. A task is accepted only if the skill completes, the
scene settles stably, and at least two of three execution seeds are classified
as intended. The successful rollouts are exported as expert demonstrations.

\begin{figure}[t]
\centering
\includegraphics[width=\linewidth]{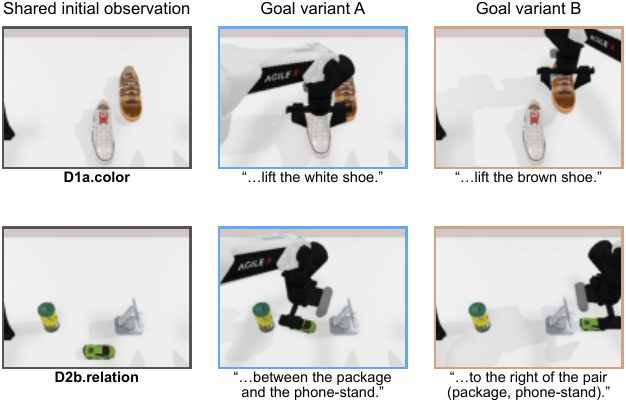}
\caption{Two counterfactual minimal pairs from RoboTwin-IF (head camera). Within
each row, both goal variants start from the identical scene; only the
instruction differs. \emph{Top} (D1a, color): ``Grasp and lift the white
shoe.'' vs.\ ``Grasp and lift the brown shoe.'' \emph{Bottom} (D2b, relation):
``Pick up the toycar and place it between the package and the phone-stand.''
vs.\ ``\ldots to the right of the pair formed by the package and the
phone-stand.''}
\label{fig:rtif_examples}
\end{figure}

\subsection{Dataset Statistics}
\label{app:robotwin_if_stats}

\begin{figure}[t]
\centering
\begin{minipage}[b]{0.49\linewidth}
\centering
\includegraphics[width=\linewidth]{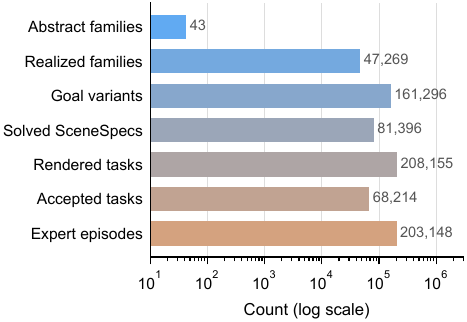}
\end{minipage}\hfill
\begin{minipage}[b]{0.48\linewidth}
\caption{Corpus size after each pipeline stage (log scale). Counts increase
from families to goal variants because each realized family supports
several goals, and from tasks to demonstrations because each accepted task
contributes up to three seeds. The main reductions occur at layout solving
and at physical validation.}
\label{fig:rtif_funnel}
\end{minipage}
\end{figure}

\begin{figure}[t]
\centering
\includegraphics[width=\linewidth]{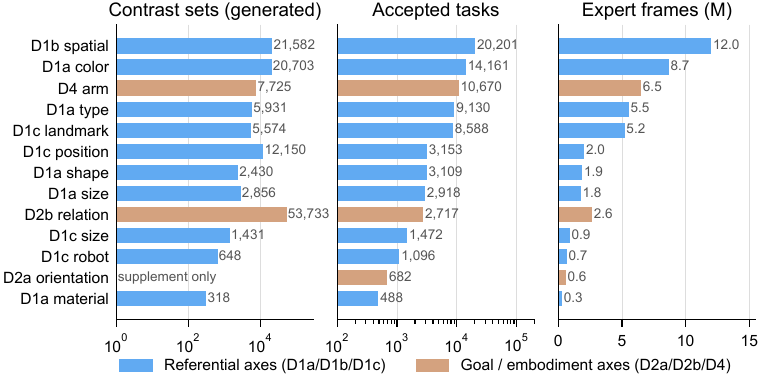}
\caption{Breakdown by axis subtype. \emph{Left}: complete contrast sets
produced by symbolic generation (stage~7). \emph{Middle}: tasks accepted after
physical validation. \emph{Right}: expert frames. D2a reorientation tasks come
from a separately validated supplement. Blue bars are referential axes; tan
bars are goal and embodiment axes.}
\label{fig:rtif_axis}
\end{figure}

Table~\ref{tab:rtif_funnel} and Fig.~\ref{fig:rtif_funnel} report the size of
the corpus after each stage. The base release has 68{,}214 accepted tasks and
203{,}148 expert demonstrations, containing $4.22\times10^{7}$ frames, which
is about 469 hours at 25\,Hz. This total includes a supplementary set of
2{,}172 demonstrations for D2a reorientation and for press/click actions.
Figure~\ref{fig:rtif_axis} breaks the corpus down by axis subtype. A task can
belong to several axes (for example, D1a+D4), so the per-axis counts overlap
and do not add up to the total. Of the accepted tasks, 58{,}043 are
single-axis, 9{,}598 combine a D1 axis with D4, and 573 combine a D1 axis
with D2b.

\begin{table*}[t]
\centering
\caption{Size of RoboTwin-IF after each pipeline stage.}
\label{tab:rtif_funnel}
\small
\begin{tabular*}{\textwidth}{@{\extracolsep{\fill}}l r l@{}}
\toprule
Stage & Count & Remark \\
\midrule
Abstract scene families & 43 & 23 axis signatures, 174 cells \\
Feasible complete assignments & $9.2\times10^{7}$ & exact count, 37 non-empty families \\
Realized scene families & 47{,}269 & $\le$2{,}048 per family, coverage-complete \\
Goal variants & 161{,}296 & 164{,}850 declared minimal-pair edges \\
Solved SceneSpecs & 81{,}396 & 27{,}132 families $\times$ 3 layout seeds \\
Rendered task records & 208{,}155 & 1{,}056 ambiguous references removed \\
Valid minimal pairs & 157{,}668 & 99.4\% of declared edges \\
Matrix-ready contrast sets & 135{,}081 & $K{=}2$/3/4 \\
Physically accepted tasks & 68{,}214 & $\ge$2/3 seeds classified intended \\
Expert demonstrations & 203{,}148 & 42.2\,M frames at 25\,Hz \\
\bottomrule
\end{tabular*}
\end{table*}

\begin{figure}[t]
\centering
\includegraphics[width=\linewidth]{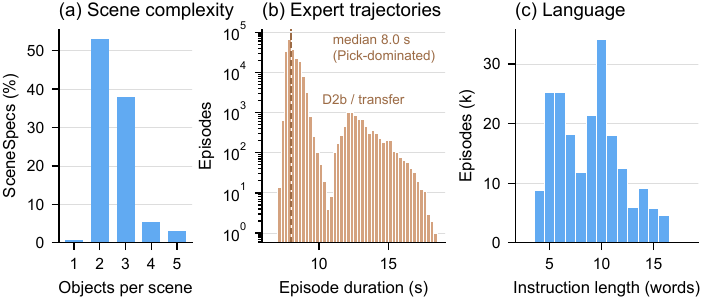}
\caption{(a) Number of objects per solved SceneSpec. (b) Duration of the expert
demonstrations (log scale). Two modes are visible: single-step acquisition
tasks (median 8.0\,s) and object-transfer (D2b) tasks (about 12--18\,s).
(c) Instruction length in words; the corpus contains 18{,}270 distinct
instruction strings.}
\label{fig:rtif_scene}
\end{figure}

\subsection{Evaluation Protocol for LDS}
\label{app:robotwin_if_eval}

The main text reports LDS on a fixed panel of 50 RoboTwin-IF scenes.
We chose the panel to make the comparison controlled: every scene comes with
two instructions that form a verified minimal pair, and \emph{only one} of the
two instructions is ever paired with action supervision.

\paragraph{Scene panel.}
The panel contains 25 \emph{Pick} scenes and 25 \emph{D2b} (pick-and-place)
scenes, all taken from the physically accepted training split. Each scene has
two instructions that form a minimal pair (Fig.~\ref{fig:rtif_examples}),
giving 100 instructions in total. Every scene and its expert videos were inspected
manually. We removed scenes whose descriptors were
visually ambiguous, along with scenes whose physical outcomes were
questionable, adjacency (\textit{next to}) goals, and near-duplicate
object/reference combinations. The Pick scenes cover ten referential subtypes:
color (5), type (3), size (2), material (2), shape (1), spatial relation (4),
ordinal position (2), ordinal size (2), distance to the robot (1), and distance
to a landmark (3). Each Pick scene keeps the descriptor contrast intact by
using a different object model or geometry from every other Pick scene. The
D2b scenes cover six destination contrasts: between vs.\ left of the pair
(7), between vs.\ right of the pair (6), left vs.\ right of the pair (3),
front vs.\ behind (3), left vs.\ right (3), and right vs.\ behind (3). We
ranked expert difficulty within each pool and drew the scenes to be balanced
across easy, medium, and hard terciles (Pick 7/10/8; D2b 8/9/8). On
average, the expert demonstrations last 8.3\,s for Pick and 16.3\,s for D2b.

\paragraph{Supervision split.}
For each scene, one of the two instructions is drawn uniformly at random
(Python \texttt{random}, seed 42) and fixed for all models and runs. Only this
instruction is paired with trajectories: its three accepted expert
demonstrations, giving 50 instructions and 150 demonstrations in total. The
\emph{counterpart} instruction is never paired with an action trajectory.
LDS adds linear-decodability supervision on top of this
trajectory training, and it applies this supervision to \emph{both}
instructions in each scene, 100 queries in total. For Pick scenes the target
is the instructed object's visible mask in the initial frame. For D2b scenes
the target is the instructed destination region on the table, also labeled
in the first input frame. Reference-object masks are never used as
destination targets. The grounding objective and training procedure are
described in Appendix~\ref{app:lds}. The readout heads are used only during training and are
discarded at inference. Both models start from the officially released RoboTwin checkpoint of their
backbone. The trajectory-only baseline is trained on the 150 demonstrations
with the model's native loss. LDS continues from this checkpoint
and keeps the same demonstrations and native loss on its action branch. Because LDS sees both instructions during
localization training, we call the counterpart instructions
\emph{localization-supervised}, not language-out-of-distribution.

\paragraph{Rollouts and metrics.}
Every instruction is evaluated with ten seeds (0--9), which control both
simulator initialization and policy sampling noise. Each checkpoint is
therefore evaluated on $50\times2\times10=1{,}000$ closed-loop rollouts
(500 Pick and 500 D2b). For a given scene and seed, both instructions start
from the bit-identical initial state. Policies receive the three rendered RGB
streams, the measured proprioceptive state, and the instruction. They receive
no expert action prefix and no auxiliary-head output. Each model runs with its
native action chunking and denoising configuration at a 25\,Hz control rate.
An episode lasts at most 1{,}000 control steps (40\,s). We use the
grasp-attachment physics and the canonical checkers from data collection
without modification. We report three metrics:
\begin{itemize}
\item \textbf{Pick contact rate}: the fraction of Pick rollouts in which a
gripper physically touches the \emph{instructed} object (summed contact
impulse $>10^{-6}$) at any time. Contacts between objects, or between an
object and the table, do not count.
\item \textbf{Pick success rate}: the fraction of Pick rollouts in which the
canonical Pick checker succeeds. The checker requires the instructed object to
be grasped, lifted off the table, and held stably, and it applies the
scene-integrity guard. Either arm may be used.
\item \textbf{D2b success rate}: the fraction of D2b rollouts in which the
canonical relation checker succeeds. The checker requires the subject to be
grasped, transferred, released in the instructed destination relation, and
settled stably, again under the scene-integrity guard.
\end{itemize}
Rollouts are assigned to workers in complete (scene, seed) groups, so the two
instructions of a minimal pair are always evaluated together.
Infrastructure errors are retried and are never counted as task failures.
A metric is reported only after all of its rollouts have completed.

\paragraph{Task list.}
Tables~\ref{tab:rtif_pick_tasks} and~\ref{tab:rtif_d2b_tasks} list all 100
evaluation instructions verbatim. Some instructions are grammatically
awkward (e.g., ``the farthest to the robot phone''); these come from the
generation templates, and we did not edit them.

\begin{table*}[t]
\centering
\footnotesize
\setlength{\tabcolsep}{3pt}
\renewcommand{\arraystretch}{1.12}
\setlength{\rtiftaskwidth}{\dimexpr\textwidth-6\tabcolsep\relax}
\caption{The 25 \emph{Pick} evaluation scenes. Each row is one shared initial
scene with its two minimal-pair instructions. $^{\dagger}$ marks the
instruction whose expert trajectories are used in training; the other
instruction receives only LDS's localization supervision.}
\label{tab:rtif_pick_tasks}
\begin{tabular}{@{}p{0.055\rtiftaskwidth} p{0.145\rtiftaskwidth} p{0.4\rtiftaskwidth} p{0.4\rtiftaskwidth}@{}}
\toprule
ID & Contrast & Instruction A & Instruction B \\
\midrule
P01 & D1a color & Grasp and lift the white shoe.$^{\dagger}$ & Grasp and lift the brown shoe. \\
P02 & D1a color & Lift the black scanner from the table.$^{\dagger}$ & Lift the multicolor scanner from the table. \\
P03 & D1a color & Pick up the brown cup.$^{\dagger}$ & Pick up the multicolor cup. \\
P04 & D1a color & Lift the red stapler from the table.$^{\dagger}$ & Lift the white stapler from the table. \\
P05 & D1a type & Lift the scanner from the table.$^{\dagger}$ & Lift the remote from the table. \\
P06 & D1a type & Grasp and lift the cards.$^{\dagger}$ & Grasp and lift the phone. \\
P07 & D1a type & Pick up the microphone.$^{\dagger}$ & Pick up the toycar. \\
P08 & D1a size & Lift the small bottle from the table.$^{\dagger}$ & Lift the medium bottle from the table. \\
P09 & D1a size & Grasp and lift the small stamp.$^{\dagger}$ & Grasp and lift the medium stamp. \\
P10 & D1a material & Lift the ceramic mug from the table.$^{\dagger}$ & Lift the metal mug from the table. \\
P11 & D1a shape & Lift the cuboid box from the table.$^{\dagger}$ & Lift the cube box from the table. \\
P12 & D1b spatial & Pick up the remote to the left of the microphone.$^{\dagger}$ & Pick up the remote to the right of the microphone. \\
P13 & D1c size & Lift the largest mouse from the table.$^{\dagger}$ & Lift the smallest mouse from the table. \\
P14 & D1a material & Pick up the metal bowl.$^{\dagger}$ & Pick up the ceramic bowl. \\
P15 & D1b spatial & Grasp and lift the cube to the right of the toycar.$^{\dagger}$ & Grasp and lift the cube to the left of the toycar. \\
P16 & D1b spatial & Grasp and lift the fan to the right of the mug.$^{\dagger}$ & Grasp and lift the fan to the left of the mug. \\
P17 & D1c position & Pick up the leftmost scanner.$^{\dagger}$ & Pick up the rightmost scanner. \\
P18 & D1c position & Pick up the rightmost jar.$^{\dagger}$ & Pick up the leftmost jar. \\
P19 & D1b spatial & Pick up the toycar to the left of the phone-stand.$^{\dagger}$ & Pick up the toycar to the right of the phone-stand. \\
P20 & D1c size & Lift the smallest trashbin from the table.$^{\dagger}$ & Lift the largest trashbin from the table. \\
P21 & D1a color & Pick up the blue bar.$^{\dagger}$ & Pick up the brown bar. \\
P22 & D1c robot & Lift the farthest to the robot phone from the table.$^{\dagger}$ & Lift the nearest to the robot phone from the table. \\
P23 & D1c landmark & Lift the nearest to the basket phone from the table.$^{\dagger}$ & Lift the farthest to the basket phone from the table. \\
P24 & D1c landmark & Grasp and lift the farthest to the bowl toycar.$^{\dagger}$ & Grasp and lift the nearest to the bowl toycar. \\
P25 & D1c landmark & Pick up the farthest to the stamp fan.$^{\dagger}$ & Pick up the nearest to the stamp fan. \\
\bottomrule
\end{tabular}
\end{table*}

\begin{table*}[t]
\centering
\footnotesize
\setlength{\tabcolsep}{3pt}
\renewcommand{\arraystretch}{1.12}
\setlength{\rtiftaskwidth}{\dimexpr\textwidth-6\tabcolsep\relax}
\caption{The 25 \emph{D2b} (pick-and-place) evaluation scenes. ``Contrast''
gives the destination relation of instructions A and B. All D2b contrasts
vary only the terminal relation (axis D2b); the subject and reference objects
are the same in both instructions. $^{\dagger}$ as in
Table~\ref{tab:rtif_pick_tasks}.}
\label{tab:rtif_d2b_tasks}
\begin{tabular}{@{}p{0.055\rtiftaskwidth} p{0.145\rtiftaskwidth} p{0.4\rtiftaskwidth} p{0.4\rtiftaskwidth}@{}}
\toprule
ID & Contrast & Instruction A & Instruction B \\
\midrule
T01 & between / right of pair & Pick up the toycar and place it between the package and the phone-stand.$^{\dagger}$ & Pick up the toycar and place it to the right of the pair formed by the package and the phone-stand. \\
T02 & right of pair / between & Pick up the bell, then set it to the right of the pair formed by the stamp and the mug.$^{\dagger}$ & Pick up the bell, then set it between the stamp and the mug. \\
T03 & between / left of pair & Transfer the cube so that it ends between the phone-stand and the cup.$^{\dagger}$ & Transfer the cube so that it ends to the left of the pair formed by the phone-stand and the cup. \\
T04 & between / right of pair & Pick up the bread, then set it between the pen holder and the phone.$^{\dagger}$ & Pick up the bread, then set it to the right of the pair formed by the pen holder and the phone. \\
T05 & left of pair / between & Pick up the phone, then set it to the left of the pair formed by the gong and the cup.$^{\dagger}$ & Pick up the phone, then set it between the gong and the cup. \\
T06 & left of pair / between & Transfer the remote so that it ends to the left of the pair formed by the toycar and the cube.$^{\dagger}$ & Transfer the remote so that it ends between the toycar and the cube. \\
T07 & between / left of pair & Transfer the rolling-pin so that it ends between the kettle and the potted-plant.$^{\dagger}$ & Transfer the rolling-pin so that it ends to the left of the pair formed by the kettle and the potted-plant. \\
T08 & between / left of pair & Transfer the bar so that it ends between the box and the cup.$^{\dagger}$ & Transfer the bar so that it ends to the left of the pair formed by the box and the cup. \\
T09 & between / left of pair & Transfer the small box so that it ends between the large box and the pen holder.$^{\dagger}$ & Transfer the small box so that it ends to the left of the pair formed by the large box and the pen holder. \\
T10 & front / behind & Pick up the cup and place it in front of the block.$^{\dagger}$ & Pick up the cup and place it behind the block. \\
T11 & right of pair / between & Transfer the stamp so that it ends to the right of the pair formed by the trashbin and the speaker.$^{\dagger}$ & Transfer the stamp so that it ends between the trashbin and the speaker. \\
T12 & behind / front & Pick up the mouse and place it behind the can.$^{\dagger}$ & Pick up the mouse and place it in front of the can. \\
T13 & between / right of pair & Pick up the box and place it between the potted-plant and the bell.$^{\dagger}$ & Pick up the box and place it to the right of the pair formed by the potted-plant and the bell. \\
T14 & right / behind & Move the sign to the right of the cube.$^{\dagger}$ & Move the sign behind the cube. \\
T15 & front / behind & Move the can in front of the bread.$^{\dagger}$ & Move the can behind the bread. \\
T16 & right of pair / left of pair & Pick up the toycar, then set it to the right of the pair formed by the bar and the bowl.$^{\dagger}$ & Pick up the toycar, then set it to the left of the pair formed by the bar and the bowl. \\
T17 & between / right of pair & Transfer the stamp so that it ends between the book and the box.$^{\dagger}$ & Transfer the stamp so that it ends to the right of the pair formed by the book and the box. \\
T18 & left of pair / right of pair & Transfer the stapler so that it ends to the left of the pair formed by the block and the sand-clock.$^{\dagger}$ & Transfer the stapler so that it ends to the right of the pair formed by the block and the sand-clock. \\
T19 & between / left of pair & Pick up the stapler and place it between the mug and the trophy.$^{\dagger}$ & Pick up the stapler and place it to the left of the pair formed by the mug and the trophy. \\
T20 & left of pair / right of pair & Pick up the stamp and place it to the left of the pair formed by the bottle and the remote.$^{\dagger}$ & Pick up the stamp and place it to the right of the pair formed by the bottle and the remote. \\
T21 & left / right & Pick up the toycar, then set it to the left of the package.$^{\dagger}$ & Pick up the toycar, then set it to the right of the package. \\
T22 & right / behind & Pick up the small speaker, then set it to the right of the medium speaker.$^{\dagger}$ & Pick up the small speaker, then set it behind the medium speaker. \\
T23 & right / left & Move the stapler to the right of the bread.$^{\dagger}$ & Move the stapler to the left of the bread. \\
T24 & left / right & Transfer the sign so that it ends to the left of the rack.$^{\dagger}$ & Transfer the sign so that it ends to the right of the rack. \\
T25 & right / behind & Move the speaker to the right of the can.$^{\dagger}$ & Move the speaker behind the can. \\
\bottomrule
\end{tabular}
\end{table*}

% Continue directly into the probing analysis without changing column mode.
% Probing protocol supplied by the author; paired with the closed-loop analysis below.
% Requires: \usepackage{booktabs,amsmath,amssymb}
\newcommand{\basemodel}{\textsc{Base}}
\newcommand{\ifmodel}{\textsc{IF}}
\newcommand{\ifsmall}{\textsc{IF-S}}

\section{Details of the Instruction-Target Probing Analysis}
\label{app:probe}
\setcounter{equation}{0}
\renewcommand{\theequation}{\thesection.\arabic{equation}}
\renewcommand{\theHequation}{\thesection.\arabic{equation}}

This appendix describes the analysis behind the probing heatmaps in the main text. We explain how the stimuli and patch-level labels are built (\S\ref{app:probe:labels}), how the
representations are extracted (\S\ref{app:probe:features}), how the linear probes are fit (\S\ref{app:probe:optim}), and how they are evaluated and compared statistically
(\S\ref{app:probe:eval}). We then relate probe accuracy to closed-loop success on a separate evaluation panel (\S\ref{app:probe:sr}) and compare cross-attention mass and spatial selectivity (\S\ref{app:attention_mass}).

We use \basemodel{} for the original non-following (NF) checkpoint and \ifmodel{} for its instruction-following counterpart.

\subsection{Stimuli and Label Construction}
\label{app:probe:labels}

\paragraph{Controlled multi-instruction scenes.}
We build a panel of $256$ scenes. Each scene contains a set of candidate objects $\mathcal{C}_s=\{o_1,\dots,o_{K_s}\}$ and comes with $K_s$ instructions $\{\ell_{s,1},\dots,\ell_{s,K_s}\}$,
where instruction $\ell_{s,k}$ refers unambiguously to candidate $o_k$. Of these scenes, $254$ have two candidates, one has three, and one has four, for $515$ instructions in total. The scenes
cover all ten referring-expression sub-categories of the grounding dimension in roughly equal numbers ($25$--$26$ scenes each). These are color, material, shape, size, object type, spatial
relation, landmark, absolute position, robot-relative position, and relative size. All instructions in a scene share one initial observation: the first frame of an expert demonstration, with
three RGB views and the proprioceptive state. We checked pixel by pixel that this observation is identical across the instructions of a scene. As a result, any change in the representation of
a scene can only come from the language input. Each scene belongs to a different layout group, meaning a set of scenes that share the same object layout.

\paragraph{Patch-level object ownership.}
We take a binary mask $M_k$ for each candidate from the simulator's instance segmentation. These masks are used only to construct labels and are never given to the model. Each mask is resized
to the head-camera region of the model input ($256\times320$ pixels) with the model's own preprocessing. We then area-pool it onto the $8\times10$ grid of visual tokens, where each token
covers $32\times32$ pixels. This gives the coverage $c_{k,p}\in[0,1]$ of patch $p\in\{1,\dots,80\}$ by candidate $k$. Each patch is assigned an owner and a weight:
\begin{equation}
\begin{aligned}
a_p&=\begin{cases}
\arg\max_k c_{k,p}, & \max_k c_{k,p}>0,\\
\varnothing, & \text{otherwise},
\end{cases}\\
\omega_p&=\max_k c_{k,p}.
\end{aligned}
\end{equation}
Here $\varnothing$ denotes background. The weight $\omega_p$ lowers the influence of patches that an object only partly covers. The mean weight over object-owned patches is about $0.39$.

\paragraph{Instruction-conditioned labels.}
For an instruction whose referent is $o_{k^\star}$, the label of a non-background patch is
\begin{equation}
y_p=\mathbb{1}\left[a_p=k^\star\right],\qquad a_p\neq\varnothing .
\end{equation}
Patches of the target object are positives, and patches of the other candidates are negatives. Background patches get no label and are not used to fit the probe or to compute the main metric.
On average, an instruction has about $7$ target patches, $7$ non-target-object patches, and $66$ background patches. We leave out background patches on purpose. Their label does not depend on
the instruction, and they make up most of the image. A trivial reader that always predicts ``not target'' would already reach $91.1\%$ per-patch accuracy over all $80$ tokens. Such a score
would mainly measure object-versus-background discrimination, not grounding of the instruction. Within a scene, changing the instruction only swaps labels among the object patches. The
instruction-dependent signal is therefore found only in the contrast between the target and the non-target objects.

\subsection{Representation Extraction}
\label{app:probe:features}

We run the model with its standard inference configuration: the original multi-view input layout, prompt template, and $10$-step denoising schedule, with a fixed random seed. Weights are not
changed. We record activations with forward hooks. To confirm that the hooks do not alter computation, we checked that the predicted action chunk with hooks matches the original one exactly
(maximum absolute difference $0$).

We read the $80$ head-camera tokens of the clean first observation frame during the first denoising call. Under the model's first-frame-causal attention mask, these tokens do not attend to
future-frame or action tokens. Later denoising calls recompute them in the same way. The first call therefore fully determines the first-frame representation. Each token has dimension
$d=3072$. We record features at all $L=30$ transformer blocks and at nine sites per block (Table~\ref{tab:probe-sites}), giving $30\times9=270$ (layer, site) cells. Some sites lie on the
residual stream and others are branch outputs before their output projection.

\begin{table*}[t]
\centering
\small
\caption{Probing sites within each transformer block.}
\label{tab:probe-sites}
\begin{tabular*}{\textwidth}{@{\extracolsep{\fill}}p{0.20\textwidth} p{0.75\textwidth}@{}}
\toprule
Site & Definition \\
\midrule
SA Q / SA K / SA V & Self-attention query/key/value projections (before normalization and RoPE) \\
SA out & Attention-weighted values of self-attention, before $W_o$ \\
Post-SA & Residual stream after the gated self-attention update \\
CA out & Attention-weighted values of text cross-attention, before $W_o$ \\
Post-CA & Residual stream after the cross-attention update \\
FFN write & Gated FFN output added to the residual stream \\
Block out & Residual stream at the block output \\
\bottomrule
\end{tabular*}
\end{table*}

\subsection{Probe Optimization}
\label{app:probe:optim}

For every model and every (layer, site) cell, we fit an independent linear probe with logit $f(x)=w^\top\widetilde{x}+b$, suppressing the model, layer, and site indices. The probe is shared by all patch positions and all instructions.
Here $\widetilde{x}=(x-\boldsymbol{\mu})\oslash\boldsymbol{\sigma}$ is the token activation standardized per dimension. The mean and standard deviation are computed on the training split
only. The training objective is a class-balanced, coverage-weighted, $\ell_2$-regularized logistic loss:
\begin{equation}
\begin{gathered}
\min_{w,b}\ \sum_{i\in\mathcal{D}_{\text{train}}}\alpha_i\,\mathrm{BCE}\!\left(f(x_i),y_i\right)
+\tfrac12\|w\|_2^2,\\
\alpha_i=\frac{N}{2}\cdot\frac{\omega_i}{\sum_{j:\,y_j=y_i}\omega_j}.
\end{gathered}
\end{equation}
where $N$ is the number of labelled training patches. The weights $\alpha_i$ give the two classes equal total weight, and within each class they are proportional to patch coverage. The bias is
not regularized. With unit weights, this objective is the same as standard $\ell_2$-regularized logistic regression with inverse regularization strength $C=1$. We optimize it with full-batch
L-BFGS: strong-Wolfe line search, history size $50$, gradient tolerance $10^{-7}$, and at most $2000$ iterations. The largest number of iterations used by any fit was $911$. As a check, we
compared one cell against a reference CPU solver, and $99.7\%$ of test-split decisions were the same.

\paragraph{Cross-validation.}
Probes are trained and scored with $5$-fold grouped cross-validation, where the grouping variable is the layout group. All instructions of a scene, and all scenes sharing a layout, fall in the
same fold. Every patch is scored by a probe fit on other scenes. The decision threshold is fixed at $f(x)>0$.

\subsection{Evaluation}
\label{app:probe:eval}

\paragraph{Bi-directional accuracy.}
Let $\hat y_p=\mathbb{1}[f(x_p)>0]$. For one instruction, let $\mathcal{T}$ be the set of target patches and $\mathcal{O}$ the set of patches owned by other candidates. We define
\begin{equation}
\begin{aligned}
\mathrm{TR}&=\frac{\sum_{p\in\mathcal{T}}\omega_p\,\hat y_p}{\sum_{p\in\mathcal{T}}\omega_p},\\
\mathrm{OR}&=\frac{\sum_{p\in\mathcal{O}}\omega_p\,(1-\hat y_p)}{\sum_{p\in\mathcal{O}}\omega_p},\\
\mathrm{BA}&=\tfrac12\left(\mathrm{TR}+\mathrm{OR}\right).
\end{aligned}
\end{equation}
These are target recall (TR), other-object rejection (OR), and bi-directional accuracy (BA). BA is high only when the probe both accepts the referred object and rejects the object that was not
referred to. We also report, separately, the unweighted false-positive rate on background patches. It is not part of BA.

BA has an exact chance level of $0.5$ for any reader that ignores the instruction, not only for random readers. In a two-candidate scene, every object patch is a positive under one instruction
and a negative under the other. A reader that does not depend on the instruction, such as an ``objectness'' detector, therefore gets $\mathrm{TR}=1,\mathrm{OR}=0$ or the reverse, averaging
$0.5$ over the two instructions. We confirmed on synthetic features that an objectness reader scores $0.5$ and a perfect reader scores $1.0$.

\paragraph{Aggregation.}
BA is computed for each instruction, averaged over the instructions of a scene, and then averaged over scenes to give the value of a (layer, site) cell in the heatmap. One scene is excluded
for both models because only one of its two candidates is visible in the head view. In that scene, TR and OR cannot both be defined for any instruction. All reported statistics therefore use
$n=255$ scenes. Our summary metric for a model is the mean BA over all $270$ cells.

\paragraph{Statistical testing.}
We fixed the testing protocol before running the comparison. Each scene gives one summary value per model. We test the difference between the two models with a paired, two-sided sign-flip
permutation test that resamples layout groups ($20{,}000$ permutations), with $p=(\#\{|\Delta^{(r)}|\ge|\Delta|\}+1)/(R+1)$. A layout-group bootstrap gives a $95\%$ confidence interval, and we
also report a Wilcoxon signed-rank test. For the cell-wise maps, we test each of the $270$ cells with the same paired test ($2{,}000$ permutations per cell). We control the false discovery
rate at $5\%$ with the Benjamini--Hochberg procedure.

\paragraph{Summary statistics.}
Table~\ref{tab:probe-summary} reports the aggregate metrics. TR and OR rise together under instruction-following training. This rules out a simple threshold shift, which would raise one and
lower the other.

\begin{table*}[t]
\centering
\small
\caption{Aggregate probing metrics, averaged over all $270$ (layer, site) cells and $255$ scenes. $\Delta$ is the paired difference (\ifmodel{} $-$ \basemodel{}) with a layout-group bootstrap
$95\%$ CI.}
\label{tab:probe-summary}
\begin{tabular*}{\textwidth}{@{\extracolsep{\fill}}lcc@{}}
\toprule
 & \basemodel{} & \ifmodel{} \\
\midrule
Bi-directional accuracy (BA) & $0.515$ & $0.775$ \\
Target recall (TR) & $0.520$ & $0.779$ \\
Other-object rejection (OR) & $0.510$ & $0.770$ \\
Background false-positive rate & $0.456$ & $0.272$ \\
Scene-level BA, median [IQR] & $0.509\ [0.503, 0.519]$ & $0.783\ [0.755, 0.807]$ \\
Cells with BA $\ge 0.9$ & $0/270$ & $91/270$ \\
\midrule
$\Delta$BA (95\% CI) & \multicolumn{2}{c}{$+0.260\ [+0.254, +0.267]$,\ \ $p<10^{-4}$} \\
Scenes improved / degraded & \multicolumn{2}{c}{$253$ / $2$} \\
Cells significant at FDR $5\%$ & \multicolumn{2}{c}{$257/270$ (all in favor of \ifmodel{})} \\
\bottomrule
\end{tabular*}
\end{table*}

\paragraph{Scope.}
The probe measures whether target identity can be decoded linearly from the visual tokens. It does not show that the policy uses this information causally when it generates actions. The two
models share an architecture, and the IF checkpoint is initialized from the NF checkpoint. The probe panel is drawn from the IF fine-tuning corpus. The cross-validation split holds out layouts from the probe fit.

\subsection{Relation to Closed-Loop Instruction Following}
\label{app:probe:sr}

\begin{table*}[t]
\centering
\small
\caption{Association between first-frame probe accuracy and closed-loop success
($36$ instructions per model, $15$ scenes, $20$ rollouts per instruction).
Intervals and $p$-values for Spearman correlations use scene-clustered bootstrap
and permutation.}
\label{tab:probe-sr}
\begin{tabular*}{\textwidth}{@{\extracolsep{\fill}}lccc@{}}
\toprule
Analysis & Estimate & $95\%$ CI & $p$ \\
\midrule
GEE, $\beta_1$ (with model term) & $1.09$ (OR $2.96$/SD) & $[0.03, 2.15]$ & $0.045$ \\
GEE, $\beta_1$ (probe only) & $1.43$ (OR $4.16$/SD) & $[0.57, 2.28]$ & $0.001$ \\
Spearman $\rho$, \basemodel{} & $0.32$ & $[-0.05, 0.59]$ & $0.136$ \\
Spearman $\rho$, \ifsmall{} & $0.33$ & $[-0.11, 0.65]$ & $0.113$ \\
Spearman $\rho$, pooled & $0.52$ & $[0.22, 0.74]$ & -- \\
Spearman $\rho$, $\Delta\bar{\mathrm{BA}}$ vs.\ $\Delta\mathrm{SR}$ & $0.36$ & $[-0.02, 0.65]$ & $0.081$ \\
\bottomrule
\end{tabular*}
\end{table*}

The analysis above shows that instruction-following training makes the instruction
target linearly decodable from visual tokens. Decodability alone does not show
that a policy uses this information when it acts. We therefore ask a weaker,
correlational question: on scenes where the policy is actually rolled out, does
the probe's accuracy on the shared first frame predict whether the instruction
is executed successfully?

\paragraph{Evaluation panel.}
We use a held-out closed-loop panel of $16$ single-object grasping scenes. Every
instruction is rolled out with $20$ random seeds and scored by the benchmark's
task-specific success checker. A rollout succeeds only if the robot grasps the
referred object and holds it stably. Probe labels are defined by which object is
selected. We therefore keep only the scenes whose instructions refer to at least
two different objects. This removes one scene whose instructions differ only in
the arm to be used. The remaining $15$ scenes ($14$ with two candidates, one with
three) contain $36$ instructions. Some instructions refer to the same object with
different arms; they receive the same patch labels. Masks, patch ownership, and
weights are built exactly as in \S\ref{app:probe:labels}.

\paragraph{Models.}
We compare \basemodel{} with \ifsmall{}, which starts from \basemodel{} and is
trained on the smaller instruction-following split for $10$K steps. \ifsmall{}
is used here because it has complete rollout results on this panel. Their
closed-loop success rates over all $16$ scenes are $11.1\%$ and $47.8\%$. On
the $36$ probed instructions, they are $7.2\%$ and $44.9\%$.

\paragraph{Probe scores.}
Here we do not refit probes on the evaluation scenes. For each model and each
(layer, site) cell, we fit one probe on \emph{all} labelled patches of the
$256$-scene probing panel. We use the objective and optimizer of
\S\ref{app:probe:optim} and that model's own features. We then extract features
for the evaluation scenes with the protocol of \S\ref{app:probe:features}.
Again, the hooked forward pass reproduces the original actions exactly. The
frozen probes are applied without any adaptation. The probe score of instruction
$j$ is its bi-directional accuracy averaged over all $270$ cells. We write
$\bar{\mathrm{BA}}_j$ for this score and standardize it over the pooled table to
get $\tilde z_j$; its pooled standard deviation is $0.103$. Mean probe scores
are $0.514$ for \basemodel{} and $0.659$ for \ifsmall{}.

\paragraph{Statistical analysis.}
The unit of analysis is the instruction. Instructions in one scene share an
image and are not independent, so all inference treats the scene as a cluster.
Before running the analysis, we chose as primary model a binomial generalized
estimating equation (GEE) with exchangeable within-scene correlation:
\begin{equation}
\begin{aligned}
\operatorname{logit}\Pr\!\left(\text{success}_{jr}=1\right)
  &=\beta_0+\beta_1\,\tilde z_j\\
  &\quad+\beta_2\,\mathbb{1}[\text{model}=\text{\ifsmall{}}],
\end{aligned}
\end{equation}
where $r$ indexes the $20$ rollouts of instruction $j$. The model indicator
absorbs the overall difference between the two models. The remaining coefficient
$\beta_1$ therefore measures how probe score relates to success after accounting
for which model is used. As secondary analyses, we report Spearman correlations
between $\bar{\mathrm{BA}}_j$ and per-instruction success rate within each model.
Confidence intervals come from a scene-level cluster bootstrap ($10^4$
resamples). Two-sided $p$-values come from a scene-level permutation test that
moves whole scene blocks of probe scores between scenes with the same number of
instructions ($10^4$ permutations). We also correlate the paired change of each
instruction, $\Delta\bar{\mathrm{BA}}_j$ against $\Delta\mathrm{SR}_j$, between
the two models.

\paragraph{Results.}
Table~\ref{tab:probe-sr} summarizes the results. After controlling for model
identity, a higher probe score goes with a higher chance of success. One standard
deviation of $\tilde z$ multiplies the odds of success by $2.96$
($\hat\beta_1=1.09$, $95\%$ CI $[0.03, 2.15]$, $p=0.045$). Within each model,
the correlations are positive and of similar size ($\rho\approx0.32$--$0.33$).
However, with only $15$ scenes they are not significant. Instruction-level changes
between models show the same direction ($\rho=0.36$, $p=0.081$). Pooling both
models without a model term gives a larger association (odds ratio $4.16$ per
SD, $p=0.001$; $\rho=0.52$). This estimate is partly driven by the difference
between models, so we report it only for completeness.

\paragraph{Interpretation.}
These results agree with the hypothesis that target information in the visual
tokens supports instruction following. The evidence, however, is limited. The
primary effect is only marginally significant, the within-model correlations are
not significant, and the analysis is correlational. A causal claim would need
interventions on the representation, which are outside the scope of this analysis.

% Attention summaries supplied by the author on 2026-11-13.
% Numeric sources and analysis scripts are archived in data/attention_mass/.
\subsection{Cross-Attention Mass and Instruction-Following Success}
\label{app:attention_mass}

We examine whether cross-attention to instruction-discriminative words
correlates with closed-loop instruction-following success. We then compare
attention mass before and after training, separately from its spatial
selectivity across candidate objects. These observational comparisons
address different questions; they do not test whether directly reallocating
attention would improve behavior.

\begin{figure*}[t]
\centering
\includegraphics[width=\textwidth]{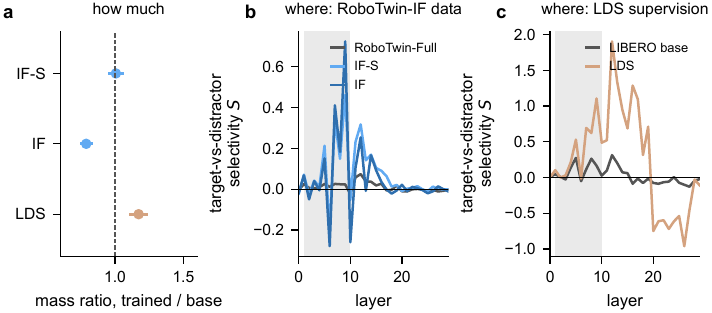}
\caption{\textbf{Cross-attention mass and spatial selectivity before and after training.}
All measurements use OpenWAM's first denoising call with matched inputs.
(a)~Geometric mean of paired trained-to-Base mass ratios, with 95\%
layout-group bootstrap CIs. IF-S and IF use 255 eligible RoboTwin-IF
scenes; LDS uses 132 LIBERO-CF contexts. The dashed line marks no change.
(b,c)~Mean target-versus-other-candidate log attention ratio at each layer
for RoboTwin-IF training and LDS, respectively. IF-S/IF are the 10K/50K
checkpoints; shaded bands mark zero-based layers 1--10.}
\label{fig:attention_mass}
\end{figure*}

\begin{table*}[t]
\centering
\small
\caption{Cross-attention mass on instruction-discriminative tokens.
Base and trained values are arithmetic scene means, expressed as percentages
of attention probability. The fold change $R$ is the geometric mean of
paired scene ratios, not the ratio of the displayed means. Intervals are
layout-group bootstrap 95\% CIs; $p$ tests paired log ratios.}
\label{tab:attention_mass}
\begin{tabular*}{\textwidth}{@{\extracolsep{\fill}}lrcccc@{}}
\toprule
Panel / trained model & Scenes & Base (\%) & Trained (\%) & $R$ [95\% CI] & $p$ \\
\midrule
RoboTwin-IF / \ifmodel{} & 255 & 0.161 & 0.112 & 0.789 [0.753, 0.825] & $<10^{-4}$ \\
RoboTwin-IF / \ifsmall{} & 255 & 0.161 & 0.143 & 1.006 [0.961, 1.055] & 0.795 \\
Grasping / \ifsmall{} & 12 & 0.147 & 0.138 & 1.051 [0.865, 1.316] & 0.683 \\
LIBERO-CF / LDS & 132 & 11.163 & 11.296 & 1.173 [1.115, 1.226] & $<10^{-4}$ \\
\bottomrule
\end{tabular*}
\end{table*}

\begin{table*}[t]
\centering
\small
\caption{Spatial selectivity of attention to discriminative tokens, averaged
over all 30 blocks. $S$ is the mean log target-to-other-candidate attention
ratio; larger values indicate greater target preference. Intervals and
tests use paired differences clustered by layout group.}
\label{tab:attention_selectivity}
\begin{tabular*}{\textwidth}{@{\extracolsep{\fill}}lrcccc@{}}
\toprule
Panel / trained model & Scenes & Base $S$ & Trained $S$ & $\Delta S$ [95\% CI] & $p$ \\
\midrule
RoboTwin-IF / \ifmodel{} & 255 & 0.011 & 0.052 & 0.042 [0.033, 0.050] & $<10^{-4}$ \\
RoboTwin-IF / \ifsmall{} & 255 & 0.011 & 0.071 & 0.061 [0.048, 0.073] & $<10^{-4}$ \\
Grasping / \ifsmall{} & 12 & 0.011 & 0.109 & 0.098 [0.065, 0.132] & 0.00025 \\
LIBERO-CF / LDS & 132 & 0.028 & 0.239 & 0.211 [0.155, 0.268] & $<10^{-4}$ \\
\bottomrule
\end{tabular*}
\end{table*}

\paragraph{Checkpoints and observations.}
On RoboTwin-IF, we compare the released OpenWAM RoboTwin-Full checkpoint
with its counterfactually trained counterparts: \ifmodel{} after 50K
updates on the larger split and \ifsmall{} after 10K updates on the
smaller split. We reuse the 256-scene, 515-instruction probing panel and
the 15-scene, 36-instruction grasping panel from
\S\ref{app:probe:labels} and \S\ref{app:probe:sr}.
All checkpoints receive the same first-frame observations and prompts,
with inference seed 42. We analyze the first denoising call, all 30 blocks,
24 heads per block, and 80 head-camera queries, including background.
Cross-attention probabilities are reconstructed from each checkpoint's
own normalized queries and keys, text projection, and proprioceptive
embedding; the UMT5 encoder is shared and frozen. A separate comparison
uses the released LIBERO checkpoint and its 1,024-update LDS counterpart
on 132 paired contexts in 23 layout groups (Appendix~\ref{app:openwam_tags}).
These are distinct training interventions and are reported separately.

\paragraph{Discriminative tokens and mass.}
For instruction $j$ in scene $s$, let $D_{sj}$ contain token positions whose
token IDs are absent from every other instruction in that scene. We exclude
the tokenizer entries for \texttt{the}, \texttt{it}, \texttt{and}, the
end-of-sequence token, and the proprioceptive token. An instruction with
$D_{sj}=\varnothing$ is omitted; scenes without an eligible instruction
are omitted. This leaves 255 probing scenes and 12 grasping scenes;
the LIBERO comparison retains all 132 contexts. This token-based filter
is separate from the visibility filter used for probing.
Writing $a_{sj\ell hp t}$ for the attention probability from head-camera
patch $p$ to key $t$ at layer $\ell$ and head $h$, we compute
\begin{equation}
\begin{aligned}
m_{sj}&=\frac{1}{30\cdot24\cdot80}
  \sum_{\ell,h,p}\sum_{t\in D_{sj}}a_{sj\ell hp t},\\
m_s&=\frac{1}{|J_s|}\sum_{j\in J_s}m_{sj},
\end{aligned}
\end{equation}
where $J_s$ is the set of eligible instructions. The softmax denominator
contains all valid prompt and proprioceptive keys; we do not renormalize
over $D_{sj}$. Each scene receives equal weight.

\paragraph{Correlation with closed-loop success.}
The 12 eligible grasping scenes provide 24 instructions per model, giving
48 model--instruction observations pooled over \basemodel{} and \ifsmall{}.
Success rates use the 20 rollouts per instruction in \S\ref{app:probe:sr}.
The Spearman correlation between per-instruction mass and success rate is
$\rho=-0.010$, with a 95\% CI of $[-0.250,0.212]$ from 5,000
scene-cluster bootstrap resamples. Each resampled scene retains all its
instructions and both models. We therefore detect no significant
correlation on this panel. This pooled analysis does not estimate a
model-adjusted effect, and its wide interval does not establish independence.
It also does not test the effect of an attention-only intervention.

\paragraph{Paired inference.}
For each scene, we form $d_s=\log(m_s^{\mathrm{trained}}/m_s^{\mathrm{Base}})$.
We report $R=\exp(\operatorname{mean}_s d_s)$ and a 95\% interval from
5,000 bootstrap resamples of complete layout groups. Two-sided sign-flip
tests flip whole groups, preserving the checkpoint pairing. On the
RoboTwin panels each retained scene has its own layout group; the LIBERO
panel has 23 groups. This protocol tests a mean log ratio, rather than
the difference between arithmetic means in Table~\ref{tab:attention_mass}.

\paragraph{Mass changes depend on the training intervention.}
Figure~\ref{fig:attention_mass}(a) visualizes the paired mass ratios.
For \ifsmall{}, the paired change is not statistically significant on
either the probing panel ($p=0.795$) or the grasping panel ($p=0.683$).
For \ifmodel{}, however, mass decreases significantly ($R=0.789$), despite
its stronger target decodability. LDS on LIBERO produces a significant
increase ($R=1.173$). Thus, the evidence does not support a blanket claim
that following and non-following models have indistinguishable attention
mass. Nor does a non-significant test establish equivalence. These
checkpoint comparisons describe training-induced changes, separately
from the instruction-level correlation above. Absolute masses across panels are not
directly comparable because the prompts and discriminative-token sets differ.

\paragraph{Spatial selectivity.}
We also measure which candidate's patches attend to the discriminative
tokens. Let $q_{sj\ell}(c)$ be their attention mass averaged over heads
and over patches of candidate $c$, weighted by its fractional mask
coverage. At each layer we take the log ratio between the target and
the arithmetic mean of the other candidates:
\begin{equation}
S_s=\frac{1}{30|J_s|}\sum_{j\in J_s}\sum_{\ell=0}^{29}
\log\frac{q_{sj\ell}(k_j^\star)}
{\frac{1}{K_s-1}\sum_{c\ne k_j^\star}q_{sj\ell}(c)}.
\end{equation}
For two candidates, the denominator is simply the other object's mass.
Table~\ref{tab:attention_selectivity} shows a positive paired change in
all four comparisons, using the same group bootstrap and sign-flip
procedure on $S_s^{\mathrm{trained}}-S_s^{\mathrm{Base}}$.
This separates a more target-selective spatial distribution from an
increase in total attention mass. Figure~\ref{fig:attention_mass}(b,c)
shows the corresponding profiles across layers.

\paragraph{Selectivity and scope.}
On the same 48 model--instruction observations, selectivity averaged over
zero-based layers 1--10 also has no significant correlation with success:
Spearman $\rho=-0.021$ (95\% scene-cluster bootstrap CI
$[-0.297,0.261]$; 5,000 resamples). Thus, a mean shift between checkpoints
does not establish that the measure ranks individual instructions by
success. This small panel cannot rule out an association. All measurements
concern first-call cross-attention and average over heads and layers;
they do not exclude effects in particular heads or layers. The absence of
a detected correlation motivates studying downstream representations,
without establishing that attention reallocation is ineffective.

% Continue directly into the real-robot data appendix.
% =====================================================================
%  Appendix: Real-Robot Data (button pressing and slot placement)
%  Required packages (preamble):
%    \usepackage{booktabs,multirow,array,amsmath,amssymb}
% =====================================================================

\section{Real-Robot Data and Annotation}
\label{app:real_robot}
\setcounter{equation}{0}

This appendix describes the data used in our real-robot experiments. We
cover the source recordings, the temporal and spatial annotations, and the
procedure that turns long-horizon demonstrations into the single-stage,
instruction-conditioned clips used for training. We study two tasks: pressing
a colored button (\emph{Press-Button}) and placing a vegetable-shaped puzzle
piece into its matching slot (\emph{Place-in-Slot}). Both tasks follow the
design principle of RoboTwin-IF (Appendix~\ref{app:robotwin_if}). The tabletop
contains several objects that are legal targets, and only the instruction
identifies the one the robot should act on.

\subsection{Platform and Source Recordings}
\label{app:real_robot_platform}

\paragraph{Embodiment and sensing.}
We collected all demonstrations on an AgileX Cobot~Magic dual-arm platform.
It has two 6-DoF arms with parallel-jaw grippers, which is the same morphology
as the \texttt{aloha-agilex} embodiment used in simulation. Demonstrations were
recorded through leader--follower teleoperation. At every time step we log the
leader (``master'') arm commands and the measured follower (``puppet'') state.
The proprioceptive stream contains six joint angles (rad) and one gripper
opening (m) per arm, which gives a 14-dimensional state. Three calibrated RGB
cameras record synchronously at $1280\times720$: one head-mounted camera and
one camera on each wrist. Depth is also recorded but is not used in this work.
All streams are time-aligned and stored at 30\,Hz in the LeRobot~v3
format~\citep{cadene2024lerobot}. Only successful demonstrations are kept.

\paragraph{Long-horizon task families.}
The raw recordings are long-horizon, multi-stage demonstrations. Each one is
conditioned on an ordered instruction.
\begin{itemize}
  \item \textbf{Sequential button pressing.} Three buttons (red, green, blue)
  are placed on the table. The robot first observes the scene and then presses
  four buttons in a prescribed color order, for example ``\emph{Press the
  buttons in the order: red $\rightarrow$ green $\rightarrow$ blue
  $\rightarrow$ red.}'' Five collections (B1--B5) use five different orders.
  \item \textbf{Sequential slot placement.} A wooden puzzle board has
  vegetable-shaped slots, and loose pieces from a pool of five categories
  (carrot, eggplant, broccoli, corn, cucumber) lie next to it. The robot
  inserts three pieces in a prescribed order, for example ``\emph{Place the
  eggplant, then the corn, then the cucumber into their matching slots.}''
  Five collections (S1--S5) use different object triples and orders.
\end{itemize}
A third family (towel folding) was recorded in the same campaign. It has no
discrete target choice, so we do not use it. In total, the ten collections
for the two tasks contain 1{,}227 demonstrations (about 13.1 hours of
interaction). Table~\ref{tab:real_robot_subsets} lists the individual
collections.

\subsection{Temporal Subtask Annotation}
\label{app:real_robot_temporal}

Every demonstration is manually segmented into contiguous, non-overlapping
subtask intervals. Each interval has a start frame, an end frame, and a
free-form label. The label vocabulary is task-specific.
\begin{itemize}
  \item Button pressing: \emph{observe and memorize the color sequence};
  \emph{press the $k$-th target button (color $c$)} for $k\in\{1,\dots,4\}$;
  and \emph{return arm to home pose}. Each annotated episode has eight
  segments.
  \item Slot placement: \emph{complete the $k$-th target puzzle (object $o$)}
  for $k\in\{1,2,3\}$, and \emph{return arm to home pose}. Each annotated
  episode has four to six segments, depending on whether the operator returned
  the arm home between placements.
\end{itemize}
The annotations are matched to the recordings through a platform-level
episode identifier, not through the positional episode index, because the
positional index changes whenever episodes are removed during format
conversion. We check every annotation automatically for missing segments,
gaps, overlaps, and intervals that end before or after the recording. We also
inspect annotations visually in a review interface that plays the three camera
streams in sync with the joint and gripper trajectories and the
segment timeline.

\subsection{From Long-Horizon Demonstrations to Instruction-Conditioned Clips}
\label{app:real_robot_clips}

Our goal is to study instruction grounding at the level of a single target
choice. We therefore split each demonstration into single-stage clips. Each
clip starts from a neutral arm configuration and ends when one target
interaction is complete. Each clip is paired with a single-target instruction
from a fixed template:
\begin{center}
\emph{``Press the red button.''}\\
\emph{``Place the $\langle$object$\rangle$ into its matching slot.''}
\end{center}
We use the following selection rules.

\begin{enumerate}
  \item \textbf{Target segments.} For button pressing we keep only segments
  whose target is the \emph{red} button. For slot placement we keep every
  placement segment and map the object name in its label to the template.
  Observation and homing segments are never kept as clips.
  \item \textbf{Reset-start constraint.} A segment is kept only if the segment
  immediately before it is a homing or observation segment, or if it is the
  first segment of the episode. This ensures that every clip starts from the
  home configuration and not halfway through a previous motion. In
  collections S3 and S4 the operator returned home after every placement, so
  all placements qualify. In S2 and S5 some placements were chained without an
  intermediate reset, and those placements are discarded.
  \item \textbf{Kinematic verification.} Some annotations call a segment a
  reset even though the arm never returned home. To catch these cases, we
  require the joint configuration at the first frame of the clip to stay
  within $0.3$\,rad (max-norm over the 12 arm joints) of the episode's initial
  configuration.
  \item \textbf{Minimum length.} Clips shorter than 33 frames ($1.1$\,s) are
  discarded. This is the length of one action chunk plus its anchor frame in
  our policy parameterization.
\end{enumerate}

We exclude collection S1 entirely because part of it has no temporal
annotation. Of the 1{,}101 demonstrations in the remaining nine collections,
65 have no annotation and are skipped. The selection yields $1{,}726$
candidate segments. The kinematic check rejects 15 of them, and no segment
fails the length check. The final dataset has \textbf{1{,}711 clips}
(353{,}878 frames, about 3.3 hours). We extract each clip with frame-exact
boundaries from all three camera streams, and we record its provenance
(source collection, episode, and frame range) so that every clip can be
traced back to the original recording. Table~\ref{tab:real_robot_subsets}
gives per-collection counts and Table~\ref{tab:real_robot_tasks} gives
per-instruction statistics.

\begin{table*}[t]
\centering
\small
\caption{Source collections and the clips extracted from each. ``Labelled''
counts demonstrations with temporal annotations. ``Rejected'' counts candidate
segments that failed the kinematic reset check. Button colors: R(ed),
G(reen), B(lue).}
\label{tab:real_robot_subsets}
\setlength{\tabcolsep}{4pt}
\begin{tabular*}{\textwidth}{@{\extracolsep{\fill}}llrrrrr@{}}
\toprule
ID & Prescribed order & Demos & Labelled & Clips & Rejected & Frames \\
\midrule
B1 & R $\rightarrow$ G $\rightarrow$ B $\rightarrow$ R & 113 & 108 & 216 & 0 & 22{,}628 \\
B2 & G $\rightarrow$ B $\rightarrow$ R $\rightarrow$ G & 106 & 102 & 101 & 1 & 9{,}710 \\
B3 & B $\rightarrow$ R $\rightarrow$ G $\rightarrow$ B & 121 & 121 & 121 & 0 & 13{,}941 \\
B4 & R $\rightarrow$ B $\rightarrow$ G $\rightarrow$ R & 125 & 125 & 123 & 2 & 11{,}074 \\
B5 & G $\rightarrow$ R $\rightarrow$ B $\rightarrow$ G & 172 & 116 & 116 & 0 & 14{,}953 \\
\midrule
S1 & carrot $\rightarrow$ eggplant $\rightarrow$ broccoli & 126 & -- & \multicolumn{3}{c}{excluded (incomplete annotation)} \\
S2 & eggplant $\rightarrow$ corn $\rightarrow$ cucumber & 120 & 120 & 238 & 2 & 74{,}812 \\
S3 & corn $\rightarrow$ broccoli $\rightarrow$ carrot & 120 & 120 & 354 & 6 & 98{,}483 \\
S4 & broccoli $\rightarrow$ cucumber $\rightarrow$ eggplant & 111 & 111 & 329 & 4 & 81{,}179 \\
S5 & cucumber $\rightarrow$ carrot $\rightarrow$ corn & 113 & 113 & 113 & 0 & 27{,}098 \\
\midrule
\multicolumn{2}{l}{Total (used collections)} & 1{,}101 & 1{,}036 & 1{,}711 & 15 & 353{,}878 \\
\bottomrule
\end{tabular*}
\end{table*}

\begin{table*}[t]
\centering
\small
\caption{Instruction-level statistics of the real-robot training set. Clip
durations are reported as mean $\pm$ standard deviation, with the range in
brackets.}
\label{tab:real_robot_tasks}
\begin{tabular*}{\textwidth}{@{\extracolsep{\fill}}lrr@{}}
\toprule
Instruction & Clips & Duration (s) \\
\midrule
\emph{Press the red button.} & 677 & $3.56 \pm 0.81$ \; [1.9, 7.5] \\
\midrule
\emph{Place the broccoli into its matching slot.} & 228 & \multirow{5}{*}{$9.08 \pm 1.56$ \; [5.6, 15.3]} \\
\emph{Place the carrot into its matching slot.}   & 117 & \\
\emph{Place the corn into its matching slot.}     & 238 & \\
\emph{Place the cucumber into its matching slot.} & 223 & \\
\emph{Place the eggplant into its matching slot.} & 228 & \\
\midrule
Total & 1{,}711 & \\
\bottomrule
\end{tabular*}
\end{table*}

\paragraph{Language--scene shortcuts in the training data.}
The resulting corpus deliberately keeps the shortcuts found in typical
real-robot data, and we use it as our \emph{biased} training distribution.
In the button task, every clip has the same instruction, even though three
equally reachable buttons are visible. The instruction therefore carries no
information during training. In the placement task, the order of objects is
fixed within each collection, so which piece is targeted correlates strongly
with the scene and with which pieces are left on the table. In 245 placement
clips only one candidate piece remains, and the scene alone determines the
target. A policy can fit this data well without grounding the instruction,
which makes it a demanding test of whether the localization supervision
described below makes the policy attend to the referred object.

\subsection{First-Frame Target Localization Annotation}
\label{app:real_robot_masks}

For the localization objective used by LDS (Appendix~\ref{app:lds}), every clip needs a
pixel-level segmentation of the instructed target and of the competing
candidate objects in the first head-camera frame. We obtain these masks with
a detect--segment--classify pipeline built from frozen foundation models and
then verify them with targeted human review.

\paragraph{Automatic proposal.}
We take the first frame of each clip at the native $1280\times720$
resolution. Open-vocabulary detection with
Grounding~DINO~\citep{liu2023grounding} proposes candidate boxes.
\begin{itemize}
  \item For button clips, we query the three button colors. We keep boxes
  below an area limit and apply non-maximum suppression (IoU $0.5$).
  \item For placement clips, we query generic piece descriptions
  (``puzzle piece'', ``small wooden puzzle piece'') and the five category
  names. We filter detections by absolute size and aspect. We also discard
  any detection that lies inside the detected puzzle board, so that pieces
  already inserted are not counted as candidates. We then apply
  non-maximum suppression (IoU $0.4$) and resolve nested or merged boxes.
  If no piece is found, we repeat detection with relaxed thresholds.
\end{itemize}
Each surviving box prompts SAM~2.1~\citep{ravi2024sam2} for an instance mask.
Button identity comes directly from the color query. For each placement
candidate, a zero-shot SigLIP~2 classifier~\citep{tschannen2025siglip2}
assigns one of the five vegetable categories from a crop around the
candidate. The target mask is the candidate whose class matches the
instruction. All other detected candidates are kept as \emph{distractor}
masks.

\paragraph{Human verification.}
A clip goes to a human review queue if any of the following holds:
(i)~the automatic stage flagged it, for example because no candidate matched
the instruction or the detections were suspected to be merged;
(ii)~any candidate's class probability is below $0.7$; or
(iii)~two candidates share a class, so the target is ambiguous. The reviewer
sees every candidate mask overlaid on the frame. For each candidate, the
reviewer either confirms or corrects its class or marks it as a false
detection. A clip is accepted only if every candidate has a verified class
and exactly one candidate matches the instructed object. Clips with missed
objects or broken masks are rejected and do not contribute a localization
target. To help the reviewer, the interface also shows, for each candidate,
how much its mask region changes between the first frame and the frame
where the next subtask begins. This visual-change score is advisory and never
used as a label.

\paragraph{Outcome.}
All 677 button clips were resolved automatically, each with three candidates
(one target and two distractors). Of the 1{,}034 placement clips, 545
(52.7\%) were sent to review. Reviewers removed false detections in 49 clips
and rejected 23 clips because an object was missed. This leaves
\textbf{1{,}688 annotated clips} (98.7\% of the corpus). Because pieces are
removed from the table as each episode progresses, placement clips contain
three (421), two (345), or one (245) candidate pieces. Over all accepted
clips, the masks contain 1{,}688 targets and 2{,}541 distractor instances.
The masks are stored at native resolution. Each model resizes them to its own
head-camera input resolution and projects them onto its own visual token
grid, so all models share the same annotations.

\subsection{Action and Observation Representation}
\label{app:real_robot_representation}

All real-robot models are trained with the same action representation. At
each anchor frame $t$, the state $s_t\in\mathbb{R}^{14}$ is the measured
follower joint configuration and gripper openings. The action target is a
chunk of 32 future steps. Its joint dimensions are expressed relative to the
current measured state,
\begin{equation}
  a^{\text{joint}}_{t,k} = q^{\text{leader}}_{t+k} - q^{\text{follower}}_{t},
  \qquad k = 0,\dots,31,
\end{equation}
and its gripper dimensions are the absolute leader gripper openings. At
deployment, each predicted step is converted to an absolute joint target,
$\hat q_{t+k} = q^{\text{follower}}_t + \hat a^{\text{joint}}_{t,k}$. This
avoids accumulating errors through integration. Using the leader
commands rather than the follower trajectory keeps the operator's intended
motion, including small anticipatory corrections, which the follower tracks
with some lag. Near the end of a clip, the action windows are padded by
repeating the last value, and the padded steps are masked out of the loss.
Normalization statistics are computed on the training split (1st and 99th
percentiles per dimension). The visual context consists of the three RGB
views sampled every four frames over the action horizon. Each model consumes
them in its native multi-view layout.

% Continue directly into LDS implementation details.
% Implementation appendix aligned with the main-text LDS formulation.
% Historical code-specific settings are tracked in IMPLEMENTATION_REVISION.md.
% Requires: booktabs,array,amsmath,amssymb,algorithm,algorithmic.

\section{Implementation Details of LDS}
\label{app:lds}
\setcounter{equation}{0}

This appendix details linear-decodability supervision (LDS): the auxiliary
readout heads (\S\ref{app:lds_readout}), first-frame grounding labels
(\S\ref{app:lds_labels}), region-balanced objective (\S\ref{app:lds_loss}),
and joint optimization with the native flow-matching loss
(\S\ref{app:lds_forward}--\ref{app:lds_optim}). We apply LDS to OpenWAM,
LingBot-VA, and InternW0-$\Delta$. The auxiliary heads are used only during
training and are discarded for evaluation and deployment.
Table~\ref{tab:lds_hparams} summarizes the settings, and
Algorithm~\ref{alg:lds} outlines one training update.

% Queue the wide configuration table and pseudocode before the next page.
\begin{table*}[t]
\centering
\small
\caption{LDS training configuration.
The grounding map resolution follows the model's visual-token grid; the
$8\times10$ example yields a $32\times40$ map.}
\label{tab:lds_hparams}
\begin{tabular*}{\textwidth}{@{\extracolsep{\fill}}ll@{}}
\toprule
Setting & Value \\
\midrule
Models & OpenWAM, LingBot-VA, InternW0-$\Delta$ \\
Supervised layers & $\mathcal{L}_s=\{9,19,29\}$ \\
Supervised observation & First input frame \\
Auxiliary denoising call & First call (step~0) only \\
Grounding targets & Instructed object or placement destination \\
Auxiliary head & Position-shared linear map, $d\rightarrow16$ per layer \\
Sub-cell grid & $4\times4$ per visual token \\
Target / alternative / background weights & $0.50$ / $0.45$ / $0.05$ \\
Auxiliary loss weight & $\lambda=0.25$ \\
WAM / head learning rate & $10^{-5}$ / $3\times10^{-4}$ \\
Optimizer separation & WAM: combined loss; heads: grounding loss only \\
Counterfactual trajectory supervision & None for instructions without demonstrations \\
Inference & Auxiliary heads discarded \\
\bottomrule
\end{tabular*}
\end{table*}

\begin{algorithm*}[t]
\small
\caption{One logical LDS update.}
\label{alg:lds}
\begin{algorithmic}[1]
\REQUIRE WAM $f_\theta$ initialized from Base; auxiliary heads $\phi$; separate WAM and head optimizers
\STATE Draw a native batch $B_{\mathrm{native}}$ from $\mathcal{D}_{\mathrm{original}}$.
\STATE Draw first-frame grounding queries $B_{\mathrm{ground}}$ from $\mathcal{D}_{\mathrm{ground}}$.
\STATE Compute the native loss $\mathcal{L}_{\mathrm{FM}}(B_{\mathrm{native}};\theta)$.
\STATE Obtain first-frame $H_\ell$ at the first denoising call for each query and $\ell\in\{9,19,29\}$.
\STATE Predict $Z_\ell=H_\ell W_\ell+b_\ell$ and align the target, alternative, and background labels with its sub-cells.
\STATE Compute $\mathcal{L}_{\mathrm{aux}}(B_{\mathrm{ground}};\theta,\phi)$ using Eqs.~\eqref{eq:lds_region_bce}--\eqref{eq:lds_aux}.
\STATE Form $\mathcal{L}_{\mathrm{total}}=\mathcal{L}_{\mathrm{FM}}+0.25\,\mathcal{L}_{\mathrm{aux}}$ and backpropagate.
\STATE Update $\theta$ with its optimizer at learning rate $10^{-5}$.
\STATE Update $\phi$ with its optimizer at learning rate $3\times10^{-4}$; only the grounding term depends on $\phi$.
\end{algorithmic}
\end{algorithm*}

\subsection{Readout Heads and Supervised Tokens}
\label{app:lds_readout}

\paragraph{Supervised representation.}
LDS acts on the visual tokens associated with the first input frame,
read at the first denoising call (step~0). Frame and denoising-step indices
refer to different axes. The auxiliary loss is evaluated only at this
call; the native flow-matching objective and inference schedule are retained.
Appendix~\ref{app:mech_interp} presents the mechanistic rationale for this
timing choice using an independent Motus case study.
Let $H_\ell\in\mathbb{R}^{N\times d}$ contain the head-camera token
representations at layer $\ell$, where $N=G_hG_w$ is the number of tokens
in the head-view grid and $d$ is the hidden width. The supervised layers are
$\mathcal{L}_s=\{9,19,29\}$. These are the same intermediate visual
representations used by the policy; LDS does not change its visual
tokenization or introduce a separate visual encoder.

\paragraph{Sub-token predictions.}
Each token predicts a $4\times4$ grid of target-grounding logits. For each
supervised layer, a linear map shared across token positions produces
\begin{equation}
  \begin{gathered}
    Z_\ell=H_\ell W_\ell+b_\ell,\\
    W_\ell\in\mathbb{R}^{d\times16},\qquad
    Z_\ell\in\mathbb{R}^{N\times16}.
  \end{gathered}
  \label{eq:lds_head}
\end{equation}
The bias $b_\ell\in\mathbb{R}^{16}$ is broadcast across positions.
Rearranging the 16 outputs of each token into its corresponding sub-cells
yields a $4G_h\times4G_w$ grounding map. In the $8\times10$ token layout
illustrated in the main text, this gives a $32\times40$ map and
$Z_\ell\in\mathbb{R}^{80\times16}$. The sub-cell layout provides finer
supervision for partially occupied patches while keeping the underlying
visual tokens unchanged.

\paragraph{Capacity of the heads.}
A head reads one token at a time and shares its weights across positions.
It therefore cannot perform its own spatial attention or combine neighboring
tokens. The auxiliary objective encourages the backbone to make the
instruction-relevant target information accessible to this linear readout.
Unlike the diagnostic probes in Appendix~\ref{app:probe}, the LDS heads are
trained jointly with the WAM and propagate gradients into its representations.

\subsection{First-Frame Grounding Labels}
\label{app:lds_labels}

Each grounding query contains an initial observation $o$, an instruction $c$,
a target region $\mathcal{T}$, and alternative candidate regions
$\mathcal{O}$. Both object-target and placement-destination supervision are
attached to the \emph{first input frame}.
\begin{itemize}
  \item \textbf{Object targets.} The positive region is the instructed
  object's visible mask in the first head-camera image. Other candidate
  objects form the negative object region. For RoboTwin-IF Pick, these are
  the objects that distinguish the paired instructions. The real-robot button
  and slot tasks use the first-frame instance masks described in
  \S\ref{app:real_robot_masks}.
  \item \textbf{Placement destinations.} The positive region is the
  instructed destination in the first input frame. For RoboTwin-IF Placement
  (D2b), paired instructions keep the manipulated object fixed and specify
  different destination relations. The counterpart destination supplies an
  alternative region. A reference object defines the spatial relation;
  its visible mask is not used as the destination target.
\end{itemize}

\paragraph{Alignment with the prediction grid.}
Target and alternative-region masks are aligned with the head-camera view
used by the model and pooled to the sub-cell grid. Let $t_i$ and $o_i$ be
the target and alternative-region coverages of sub-cell $i$, respectively,
and let $g_i$ indicate background. The coverages lie in $[0,1]$ and are
arranged in the same token and sub-cell order as $Z_\ell$. Background
consists of sub-cells with neither target nor alternative coverage. These
spatial weights align the regional supervision with the prediction map.

\paragraph{Supervision availability.}
Grounding labels do not add action demonstrations. The native trajectory
training set remains unchanged, while grounding supervision can include
instructions without corresponding trajectories. In RoboTwin-IF, only one
instruction per scene supplies demonstrations, while both members of the
pair supply grounding labels (\S\ref{app:robotwin_if_eval}). The labels
are training targets and are not provided as policy inputs at inference.

\subsection{Region-Balanced Grounding Objective}
\label{app:lds_loss}

The auxiliary objective separates the instructed target, alternative
candidates, and background. This prevents the large background area from
dominating the supervision. Let $z_i^{(\ell)}$ denote a sub-cell logit from
layer $\ell$. For regions with nonzero coverage, the normalized losses are
\begin{equation}
  \begin{aligned}
    \mathcal{L}_{\mathrm{target}}^{(\ell)}
      &=\frac{\sum_i t_i\,\operatorname{softplus}(-z_i^{(\ell)})}
              {\sum_i t_i},\\
    \mathcal{L}_{\mathrm{other}}^{(\ell)}
      &=\frac{\sum_i o_i\,\operatorname{softplus}(z_i^{(\ell)})}
              {\sum_i o_i},\\
    \mathcal{L}_{\mathrm{bg}}^{(\ell)}
      &=\frac{\sum_i g_i\,\operatorname{softplus}(z_i^{(\ell)})}
              {\sum_i g_i}.
  \end{aligned}
  \label{eq:lds_region_bce}
\end{equation}
Here $\operatorname{softplus}(-z)$ and $\operatorname{softplus}(z)$ are
binary cross-entropy losses for positive and negative labels. Coverage
weights the spatial contribution of each sub-cell within a region. It does
not change the binary target into a soft label.

The regional terms are combined using the weights in the main text:
\begin{equation}
  \begin{aligned}
    \mathcal{L}_{\mathrm{aux}}
      &=\frac{1}{|\mathcal{L}_s|}\sum_{\ell\in\mathcal{L}_s}
        \bigl(0.50\,\mathcal{L}_{\mathrm{target}}^{(\ell)}\\
      &\hspace{3em}+0.45\,\mathcal{L}_{\mathrm{other}}^{(\ell)}
        +0.05\,\mathcal{L}_{\mathrm{bg}}^{(\ell)}\bigr).
  \end{aligned}
  \label{eq:lds_aux}
\end{equation}
The batch objective averages over grounding queries. The alternative-region
term encourages the representation to distinguish the instructed target from
other plausible goals in the same observation. Background receives a smaller
weight because object--background separation alone does not require following
the instruction. Equation~\eqref{eq:lds_region_bce} is stated for non-empty
regions; an absent region contributes no spatial samples.

\subsection{Joint Training and Gradient Routing}
\label{app:lds_forward}

\paragraph{Native and auxiliary supervision.}
The WAM retains its native flow-matching objective on the original
trajectory data, $\mathcal{D}_{\mathrm{original}}$. Grounding queries from
$\mathcal{D}_{\mathrm{ground}}$ supervise the auxiliary heads and the
intermediate visual representations. Following the main text, the total
objective is
\begin{equation}
  \begin{aligned}
    \mathcal{L}_{\mathrm{total}}
      &=\mathcal{L}_{\mathrm{FM}}(\mathcal{D}_{\mathrm{original}})\\
      &\quad+\lambda\mathcal{L}_{\mathrm{aux}}
        (\mathcal{D}_{\mathrm{ground}}),\qquad \lambda=0.25.
  \end{aligned}
  \label{eq:lds_total}
\end{equation}
A grounding query without a trajectory contributes only to the auxiliary
term. It supplies neither action nor future-video supervision. The
observation and instruction are processed through the model's visual and
language conditioning, and the auxiliary heads read the selected first-frame
visual tokens.

\paragraph{Parameter updates.}
Let $\theta$ denote the trainable WAM parameters and
$\phi=\{W_\ell,b_\ell\}_{\ell\in\mathcal{L}_s}$ the auxiliary-head
parameters. The WAM receives gradients from both terms of
Eq.~\eqref{eq:lds_total}. The heads receive gradients only through the
grounding objective, because they do not participate in native action or
video prediction. Shared parameters receive grounding gradients through
their contribution to the supervised visual representations.

\subsection{Optimization and Model Integration}
\label{app:lds_optim}
\label{app:lds_models}

\paragraph{Base checkpoints.}
For LIBERO-CF and LangGap, LDS starts from each model's released LIBERO
checkpoint. For RoboTwin-IF, a Base checkpoint is first obtained by
fine-tuning the released RoboTwin checkpoint on demonstrations for one
selected instruction per scene. LDS then continues from that Base checkpoint
while retaining its original trajectory supervision.

\paragraph{Separate optimizers.}
The WAM and the auxiliary heads use separate optimizers. The WAM learning
rate is $10^{-5}$, and the head learning rate is $3\times10^{-4}$. The WAM
optimizer uses the combined objective; the head optimizer uses only grounding
gradients. Table~\ref{tab:lds_hparams} summarizes these settings, and
Algorithm~\ref{alg:lds} gives the logical training update. This description
is independent of how native and grounding batches are distributed across
devices.

\paragraph{Model integration and inference.}
The same LDS objective is applied to the three WAM families. Each model's
head-camera representations and label grid are aligned with its own visual
layout. LDS adds auxiliary supervision to the native architecture; the
linear heads do not supply inputs to the policy. Evaluation uses the updated
WAM parameters with all auxiliary heads removed.

% Scene hold-out protocol and standardized aggregation for main Table 4.
% =====================================================================
%  Appendix: Scene Hold-Out Protocol and Composition-Standardized Aggregation
%  Required packages (preamble):
%    \usepackage{booktabs,multirow,array,amsmath,amssymb}
%  Numbers are produced by make_scene_holdout_weighted.py from the
%  per-episode evaluation records (scene_holdout_weighted.json).
% =====================================================================
\providecommand{\method}{LDS}

\section{Scene Hold-Out Protocol and Composition-Standardized Aggregation}
\label{app:scene_holdout}
\setcounter{equation}{0}

This appendix specifies the hold-out protocol behind the generalization
results of \holdoutmaintableref{}, which reports episode-pooled rates.
It defines the units over which grounding supervision is withheld and
 describes the split. As a supplementary comparison, we also report
composition-standardized rates because the coverage groups contain
different mixtures of task types.

\subsection{Units of Supervision}
\label{app:scene_holdout_units}

Both benchmarks are built on LIBERO, and we distinguish four units.
\begin{itemize}
  \item \textbf{Initial state.} Each task comes with a fixed set of official
  initial states: 50 per task in LIBERO-CF and 10 in LangGap. Initial states of
  the same task share the object set and layout and differ only in the
  randomized poses of the objects.
  \item \textbf{Task.} A task is one BDDL specification paired with one
  instruction. An evaluation episode is a (task, initial state) pair.
  \item \textbf{Physical scene.} A physical scene groups all tasks that share
  the same physical description, meaning the same object set and placement
  regions in the BDDL physical sections, or the same official initial-state
  file. We take the transitive closure of both relations, so tasks connected
  through either one always end up in the same scene. A scene typically
  contains one to four tasks.
  \item \textbf{Task kernel.} The task kernel abstracts away the concrete
  scene. It records how an instruction departs from the original task of its
  scene: which element changes (the object to pick, the destination, or both),
  the spatial predicate (\textit{on} or \textit{in}), and the destination
  category with instance identifiers removed. Examples are
  \texttt{pick:on:plate} (a different object goes to the same plate) and
  \texttt{destination:on:flat\_stove} (the same object goes to a different
  receptacle, the stove). Goals with several literals and articulated actions
  (drawers, stove knobs) get a separate \texttt{other:} kernel.
\end{itemize}

\paragraph{Grounding supervision.}
\method{} supervises a \emph{localization view}: the first input frame of a
supervised task, together with labels of the instructed target and
alternative candidates. Pick views label objects; Place views label
placement destinations. Pick views use official initial states 0, 1, and 2
of each supervised task. Place views use the initial states with valid
destination-localization labels. The number of included initial states can
therefore differ between task types.

Within the LIBERO-CF supervision panel, Place views come from the two tasks
whose counterfactual goals move the object to a different receptacle body
(Long-4 and Long-6). The LangGap supervision panel contains twelve
destination-change tasks. Both types use first-frame supervision, consistent
with the main text. The hold-out affects only grounding supervision;
demonstration trajectories used by the native action objective are unchanged.

\subsection{Hold-Out Construction}
\label{app:scene_holdout_split}

We withhold \emph{entire physical scenes} from grounding supervision and
stratify the hold-out by task kernel. Let $\mathcal G_k$ denote the set of
physical scenes that contain at least one supervision task of kernel $k$. The
hold-out quota of kernel $k$ is
\begin{equation}
  \begin{gathered}
    m_k=\max\!\bigl(1,\operatorname{round}(|\mathcal G_k|/3)\bigr),\\
    q_k=\begin{cases}
      0, & |\mathcal G_k|=1,\\
      \min\!\bigl(|\mathcal G_k|-1,m_k\bigr), & |\mathcal G_k|\ge2.
    \end{cases}
  \end{gathered}
\end{equation}
We visit all scenes once, in a random order drawn with a fixed seed
(2026102727). A scene is held out if every kernel it contains still has
unused quota, and the quota of each such kernel is then decremented. The
procedure enforces three properties.
\begin{enumerate}
  \item No scene contributes both supervised and held-out views.
  \item No task appears in both splits.
  \item Every kernel keeps at least one supervised scene.
\end{enumerate}
Together, these guarantee that an episode from a held-out scene tests a
\emph{familiar} kernel in an \emph{unfamiliar} scene. Kernels with a single
scene are never held out. The resulting split holds out 7 of 24 scenes in
LIBERO-CF and 6 of 18 in LangGap.
We draw the split once and share it across all model families: OpenWAM,
LingBot-VA and InternW0-$\Delta$ are fine-tuned on identical
supervision. Each model's training run is initialized independently, and no
model continues from its all-train checkpoint.
Table~\ref{tab:holdout_split} summarizes the split.

\begin{table*}[t]
\centering
\small
\caption{Scene hold-out split of the grounding supervision. Contexts are
localization views; each view is supervised with its two instructions
(queries). A kernel is \emph{split} if at least one of its scenes is held
out.}
\label{tab:holdout_split}
\begin{tabular*}{\textwidth}{@{\extracolsep{\fill}}lcc@{}}
\toprule
 & LIBERO-CF & LangGap \\
\midrule
Supervision tasks / physical scenes & 46 / 24 & 52 / 18 \\
Task kernels (split / total) & 2 / 7 & 5 / 11 \\
Held-out scenes / tasks & 7 / 11 & 6 / 12 \\
Contexts: supervised / held out & 117 / 33 & 162 / 56 \\
Queries: supervised / held out & 234 / 66 & 324 / 112 \\
\bottomrule
\end{tabular*}
\end{table*}

In LIBERO-CF, the split kernels are \texttt{pick:in:basket} (3 of 10 scenes
held out) and \texttt{pick:on:plate} (4 of 11). Both destination kernels have
a single scene and therefore remain fully supervised. In LangGap, the hold-out
covers \texttt{pick:in:basket} (3/10), \texttt{pick:on:plate} (2/6),
\texttt{destination:on:flat\_stove} (2/5),
\texttt{destination:on:wooden\_cabinet} (1/4), and
\texttt{destination:on:glazed\_rim\_porcelain\_ramekin} (1/3).

\subsection{Coverage Groups}
\label{app:scene_holdout_coverage}

We evaluate every model under the full official protocol: all 50 LIBERO-CF
tasks with 50 initial states each, and all 59 LangGap tasks with 10 initial
states each. Each episode is assigned to a group according to the
supervision of \emph{the evaluated model}.
\begin{itemize}
  \item \textbf{Sup.} The (task, initial state) pair was itself a
  localization view.
  \item \textbf{Unsup.\ init.} The task was supervised, but not from this
  initial state.
  \item \textbf{Unsup.\ scene.} The task was never supervised, but its kernel
  was supervised in another scene.
  \item \textbf{Unsup.\ kernel.} The kernel was never supervised. We exclude
  this group from the standardized comparison and make no generalization claim for it. It contains
  50 LIBERO-CF and 70 LangGap episodes.
\end{itemize}
The Unsup.\ scene group contains the held-out tasks. In LIBERO-CF it also
contains three evaluation tasks that are absent from the supervision panel
but share a supervised kernel (Long-1, Long-2, and Long-3).
Table~\ref{tab:holdout_composition} lists the size of each group.

\subsection{Composition Imbalance Across Coverage Groups}
\label{app:scene_holdout_imbalance}

Pick and Place views sample initial states at very different rates. A
supervised Pick task contributes exactly three of its initial states to Sup.,
which is 6\% of a LIBERO-CF task and 30\% of a LangGap task. A supervised
Place task contributes the initial states with valid destination labels,
which is 83\% of a LangGap Place task on average. Place episodes
therefore make up a much larger share of Sup.\ than of Unsup.\ init
(Table~\ref{tab:holdout_composition}). In LangGap, Place episodes make up
40.7\% of Sup.\ but only 5.9\% of Unsup.\ init, although they make up 23.1\%
of the benchmark's tasks.

\begin{table*}[t]
\centering
\small
\caption{Composition of the coverage groups: episodes (tasks) by task type.
The reference share is the fraction of Place tasks among all tasks outside
Unsup.\ kernel.}
\label{tab:holdout_composition}
\setlength{\tabcolsep}{4pt}
\begin{tabular*}{\textwidth}{@{\extracolsep{\fill}}llcccc@{}}
\toprule
Benchmark & Group & Pick & Place & Total & Place share \\
\midrule
\multirow{4}{*}{LIBERO-CF}
 & Sup.          & 99 (33)   & 18 (2) & 117  & 15.4\% \\
 & Unsup.\ init  & 1551 (33) & 82 (2) & 1633 & 5.0\% \\
 & Unsup.\ scene & 650 (13)  & 50 (1) & 700  & 7.1\% \\
 & Reference     & (46)      & (3)    &      & 6.1\% \\
\midrule
\multirow{4}{*}{LangGap}
 & Sup.          & 96 (32)  & 66 (8) & 162 & 40.7\% \\
 & Unsup.\ init  & 224 (32) & 14 (6) & 238 & 5.9\% \\
 & Unsup.\ scene & 80 (8)   & 40 (4) & 120 & 33.3\% \\
 & Reference     & (40)     & (12)   &     & 23.1\% \\
\bottomrule
\end{tabular*}
\end{table*}

This imbalance matters because the two task types differ sharply in
difficulty. Place requires an acquisition, a transport, and a relational
placement, and for most models it has a much lower success rate than Pick
(Table~\ref{tab:holdout_by_type}). Write $p_g$ and $r_g$ for the Pick and
Place success rates in group $g$, and $\pi_g$ for its Place share. The episode-pooled
rate is then
\begin{equation}
  \bar s_g^{\text{pool}} \;=\; (1-\pi_g)\,p_g + \pi_g\, r_g .
\end{equation}
Because $\pi_g$ differs across groups, a difference between two pooled rates
mixes a genuine difference in performance with a difference in task mix. The
mix effect can even reverse the comparison, an instance of Simpson's paradox.
On LangGap, LingBot-VA has similar Pick rates in Sup.\ and Unsup.\ init
(25.0\% and 26.3\%) and fails every Place episode in both groups. Yet its
pooled Sup.\ rate (14.8\%) is ten points below its pooled Unsup.\ init rate
(24.8\%), because Place makes up 40.7\% of the former and 5.9\% of the
latter. The same effect pulls down the pooled Sup.\ rate of InternW0-$\Delta$.

\subsection{Composition-Standardized Rates}
\label{app:scene_holdout_standardization}

We remove the mix effect with direct standardization. Every group is
evaluated under one common reference composition $\pi^\star$:
\begin{equation}
  \begin{gathered}
    \bar s_g^{\text{std}}=(1-\pi^\star)\,p_g+\pi^\star\,r_g,\\
    \pi^\star=\frac{\#\,\text{Place tasks}}
                    {\#\,\text{tasks with supervised kernels}}.
  \end{gathered}
  \label{eq:holdout_std}
\end{equation}
The rates $p_g$ and $r_g$ are pooled over the episodes of each type within
the group. As the reference we take the task composition of the benchmark
itself, giving equal weight to each task outside Unsup.\ kernel:
$\pi^\star = 3/49 = 6.1\%$ for LIBERO-CF and $12/52 = 23.1\%$ for LangGap.
This reference has three useful properties. It does not depend on how many
initial states the supervision protocol happens to sample. It is the same for
all groups and all models. It also matches the Pick/Place composition of uniformly sampled evaluation
tasks. For a group whose composition matches the reference, the standardized
and pooled rates coincide. When
groups have the same per-type rates, their standardized rates are equal,
whatever their composition. We apply the same weights to the biased-success
metric of LIBERO-CF.

\paragraph{Uncertainty.}
We compute 95\% confidence intervals with a stratified, task-clustered
bootstrap. Within each group and task type, we resample tasks with
replacement and then resample episodes within each drawn task. The two
resampled rates are combined with the fixed weights of
Eq.~\eqref{eq:holdout_std}. We use 2{,}000 replicates and report these intervals with the standardized
point estimates in Table~\ref{tab:holdout_standardized}.

% Queue result tables before the discussion so they precede the references.
\begin{table*}[t]
\centering
\small
\setlength{\tabcolsep}{3.5pt}
\caption{Scene hold-out results (\%) under pooled and composition-standardized
aggregation. \emph{Pooled} averages all episodes of a group and gives the
values used in \holdoutmaintableref{}. \emph{Std.}\ is a supplementary comparison, reweighting
Pick and Place to the benchmark's task composition (Eq.~\eqref{eq:holdout_std}; Place weight
6.1\% for LIBERO-CF and 23.1\% for LangGap). F.: faithful success
($\uparrow$); B.: biased success ($\downarrow$). LangGap reports success.}
\label{tab:holdout_standardized}
\begin{tabular*}{\textwidth}{@{\extracolsep{\fill}}ll*{9}{c}@{}}
\toprule
\multirow{3}{*}{Model} & \multirow{3}{*}{Agg.} & \multicolumn{6}{c}{LIBERO-CF} & \multicolumn{3}{c}{LangGap SR$\uparrow$} \\
\cmidrule(lr){3-8}\cmidrule(lr){9-11}
 & & \multicolumn{2}{c}{Sup.} & \multicolumn{2}{c}{Unsup.\ init} & \multicolumn{2}{c}{Unsup.\ scene} & \multirow{2}{*}{Sup.} & \multirow{2}{*}{Unsup.\ init} & \multirow{2}{*}{Unsup.\ scene} \\
\cmidrule(lr){3-4}\cmidrule(lr){5-6}\cmidrule(lr){7-8}
 & & F. & B. & F. & B. & F. & B. & & & \\
\midrule
\multirow{2}{*}{OpenWAM} & Pooled & 61.5 & 13.7 & 66.9 & 4.8 & 49.6 & 32.0 & 65.4 & 64.7 & 39.2 \\
 & Std. & 67.1 & 7.3 & 66.1 & 5.7 & 50.0 & 31.4 & 64.2 & 59.4 & 41.0 \\
\midrule
\multirow{2}{*}{LingBot-VA} & Pooled & 26.5 & 31.6 & 32.0 & 23.1 & 27.4 & 24.1 & 14.8 & 24.8 & 25.0 \\
 & Std. & 29.4 & 24.1 & 31.6 & 23.9 & 27.7 & 23.4 & 19.2 & 20.3 & 28.8 \\
\midrule
\multirow{2}{*}{InternW0-$\Delta$} & Pooled & 71.8 & 14.5 & 79.1 & 8.1 & 67.7 & 9.4 & 69.1 & 76.1 & 45.8 \\
 & Std. & 79.7 & 7.0 & 78.2 & 8.9 & 67.4 & 9.5 & 73.0 & 73.9 & 46.0 \\
\bottomrule
\end{tabular*}

\vspace{4pt}
\begin{minipage}{\textwidth}
\footnotesize\raggedright
95\% CIs of standardized rates (task-clustered bootstrap), in the column order above.\par
\textbf{OpenWAM:} [54.1, 78.9], [3.7, 13.0], [56.0, 75.5], [4.5, 7.2], [30.6, 67.9], [12.9, 53.1], [51.5, 75.4], [47.8, 71.9], [16.5, 66.9].\par
\textbf{LingBot-VA:} [16.1, 43.6], [13.7, 36.5], [20.5, 43.3], [15.0, 34.5], [10.1, 46.4], [7.9, 42.5], [8.8, 31.2], [10.0, 31.6], [9.6, 57.7].\par
\textbf{InternW0-$\Delta$:} [69.2, 89.1], [3.7, 11.8], [69.7, 85.3], [5.3, 13.6], [43.2, 88.3], [0.0, 25.4], [61.2, 84.1], [60.4, 86.1], [21.5, 71.2].\par
\end{minipage}
\end{table*}

\begin{table*}[t]
\centering
\small
\setlength{\tabcolsep}{3.5pt}
\caption{Per-type success rates (\%) within each coverage group, the
quantities $p_g$ (Pick) and $r_g$ (Place) of Eq.~\eqref{eq:holdout_std}.
Episode and task counts per cell are given in
Table~\ref{tab:holdout_composition}. LIBERO-CF Place cells contain one or two
tasks.}
\label{tab:holdout_by_type}
\begin{tabular*}{\textwidth}{@{\extracolsep{\fill}}ll*{9}{c}@{}}
\toprule
\multirow{3}{*}{Model} & \multirow{3}{*}{Type} & \multicolumn{6}{c}{LIBERO-CF} & \multicolumn{3}{c}{LangGap SR$\uparrow$} \\
\cmidrule(lr){3-8}\cmidrule(lr){9-11}
 & & \multicolumn{2}{c}{Sup.} & \multicolumn{2}{c}{Unsup.\ init} & \multicolumn{2}{c}{Unsup.\ scene} & \multirow{2}{*}{Sup.} & \multirow{2}{*}{Unsup.\ init} & \multirow{2}{*}{Unsup.\ scene} \\
\cmidrule(lr){3-4}\cmidrule(lr){5-6}\cmidrule(lr){7-8}
 & & F. & B. & F. & B. & F. & B. & & & \\
\midrule
\multirow{2}{*}{OpenWAM} & Pick & 70.7 & 3.0 & 70.3 & 0.8 & 52.6 & 27.5 & 62.5 & 66.5 & 45.0 \\
 & Place & 11.1 & 72.2 & 1.2 & 80.5 & 10.0 & 90.0 & 69.7 & 35.7 & 27.5 \\
\midrule
\multirow{2}{*}{LingBot-VA} & Pick & 31.3 & 19.2 & 33.7 & 19.1 & 29.5 & 18.9 & 25.0 & 26.3 & 37.5 \\
 & Place & 0.0 & 100.0 & 0.0 & 97.6 & 0.0 & 92.0 & 0.0 & 0.0 & 0.0 \\
\midrule
\multirow{2}{*}{InternW0-$\Delta$} & Pick & 84.8 & 2.0 & 83.3 & 4.3 & 65.2 & 10.2 & 78.1 & 76.8 & 46.2 \\
 & Place & 0.0 & 83.3 & 0.0 & 79.3 & 100.0 & 0.0 & 56.1 & 64.3 & 45.0 \\
\bottomrule
\end{tabular*}
\end{table*}

\subsection{Results}
\label{app:scene_holdout_results}

Table~\ref{tab:holdout_standardized} compares the pooled and standardized
rates, and Table~\ref{tab:holdout_by_type} reports the per-type rates from
which both are derived. Three observations follow.

\paragraph{Task composition explains much of the Sup./Unsup.\ init gap.}
After standardization, LangGap Sup.\ and Unsup.\ init are close for every
model that showed an inversion: LingBot-VA 19.2 vs.\ 20.3 and InternW0-$\Delta$
73.0 vs.\ 73.9, compared with 14.8 vs.\ 24.8 and 69.1 vs.\ 76.1 pooled. On
LIBERO-CF, standardization raises faithful success in Sup.\ (OpenWAM
61.5$\rightarrow$67.1, InternW0-$\Delta$ 71.8$\rightarrow$79.7) and lowers biased
success (OpenWAM 13.7$\rightarrow$7.3). This happens because the two
LIBERO-CF Place tasks are over-represented in Sup., and most models still
complete the original goal on them. Standardization substantially reduces these apparent gaps. It does not
make the groups identical: their tasks and initial states still differ, so
the remaining rate differences are not a matched estimate of generalization
loss.

\paragraph{Held-out scene results remain qualitatively consistent.}
The Unsup.\ scene group has a Place share of 33.3\% on LangGap and
7.1\% on LIBERO-CF, compared with reference shares of 23.1\% and 6.1\%.
Its standardized rates move by at most 3.8 percentage points. The qualitative conclusion of
\holdoutmaintableref{} stands under both aggregations. Instruction
following is observed in scenes withheld from grounding supervision,
although success rates remain below those in supervised scenes for most
model--benchmark combinations. For example, OpenWAM reaches
50.0\% faithful success in LIBERO-CF held-out scenes against 66.1\% for
unsupervised initial states, and 41.0\% against 59.4\% on LangGap.

\paragraph{The Place strata are small.}
In LIBERO-CF, Place is represented by two tasks in Sup.\ and Unsup.\ init and
by a single task (Long-1) in Unsup.\ scene. Its per-type rates are therefore
close to all-or-nothing and are highly uncertain. An example is
InternW0-$\Delta$'s 100\% on the one held-out Place task. Because
$\pi^\star=6.1\%$, these rates affect the standardized LIBERO-CF numbers only
marginally. We report the per-type rates so that readers can apply any other
reference composition, including an equal Pick/Place weighting, directly
through Eq.~\eqref{eq:holdout_std}.

% Mechanistic evidence motivating LDS supervision timing and representation.
% Design rationale adapted from the author-supplied mech_interp appendix.
% Motus is an independent diagnostic model, not a renamed main-text model.
% The supplied vector figures and reported statistics are retained.

\section{Mechanistic Evidence for LDS Design Choices}
\label{app:mech_interp}
\setcounter{equation}{0}

This appendix examines two choices in LDS: applying the auxiliary loss at
the first denoising step, and supervising instruction-conditioned visual
representations. We trace how an instruction affects action generation in
Motus, an independent mechanistic case study. Activation interventions show
that most of its measurable instruction effect enters through the video
stream during the first denoising update. These findings motivate our design;
they do not establish that first-step supervision is optimal for every WAM.
We describe the diagnostic setting (\S\ref{app:mech_interp_setup}), the
evidence for first-step supervision (\S\ref{app:mech_interp_step0}), the
visual pathway (\S\ref{app:mech_interp_route}), and the implications for
spatial grounding (\S\ref{app:mech_interp_spatial}).

% Queue the wide figures early so that they accompany the analysis,
% rather than drifting past the supplementary references.
\begin{figure*}[!t]
  \centering
  \includegraphics[width=\textwidth]{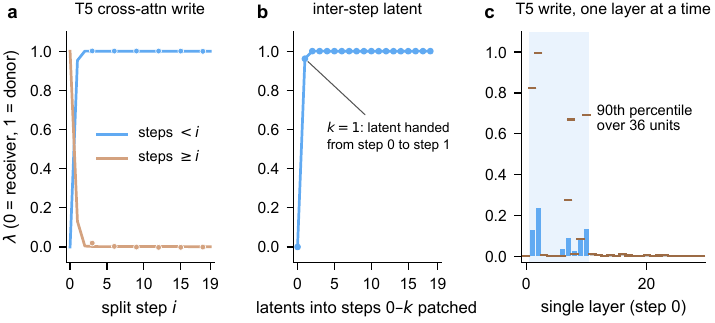}
  \caption{\textbf{Instruction influence is concentrated early in denoising.}
  (a)~Donor cross-attention outputs patched at steps $<i$ or $\geq i$:
  lines show the stride-1 sweep (4 units), and markers the stride-3 sweep
  (36 units). (b)~Patching inter-step latents shows that the state passed
  from step~0 to step~1 already carries most of the donor effect.
  (c)~Single-layer cross-attention patches within step~0 (36 units).
  Bars are means and ticks the 90th percentile; layers 1--10 are shaded.
  All panels use the action projection score in Eq.~\eqref{eq:mi_lambda}.}
  \label{fig:mi_step0}
\end{figure*}

\begin{figure*}[!t]
  \centering
  \includegraphics[width=0.85\textwidth]{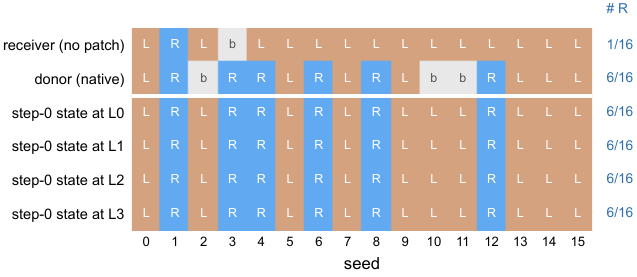}
  \caption{\textbf{Early-state replacement preserves a component of behavior.}
  Arm modes over 16 seeds in the controlled two-cup experiment: L, left
  only; R, right only; b, both arms. The four patched conditions replace
  all three residual streams at a layer boundary within step~0, then resume
  receiver conditioning. They recover the donor's six R seeds, but not its
  three b seeds. The right margin counts R outcomes, not task success.}
  \label{fig:mi_commit}
\end{figure*}

\subsection{Diagnostic Model and Intervention Protocol}
\label{app:mech_interp_setup}

\paragraph{Independent model.}
We analyze a RoboTwin-finetuned Motus checkpoint, separately from OpenWAM,
LingBot-VA, and InternW0-$\Delta$ in the main evaluation. Motus jointly
denoises eight future video frames and a chunk of 16 bimanual actions,
$A\in\mathbb{R}^{16\times14}$, using 20 flow-matching steps. Its 30
transformer blocks contain three residual streams: video (360 tokens, width
3072), action (21 tokens, width 1024), and understanding (187 tokens, width
512). These streams exchange information through joint self-attention over
568 tokens; feed-forward layers are stream-specific. UMT5 encodes the
instruction for video-stream cross-attention, while Qwen3-VL processes the
observation and instruction to initialize the understanding stream. The
action stream has no direct language cross-attention. Between denoising
steps, the video and action latents are carried forward and the residual
streams are rebuilt.

\paragraph{Step, frame, and layer indices.}
We index denoising calls from $k=0$ to $19$. Thus, \emph{step~0} means the
first sampler call, not a diffusion-time value of zero or a clean sample.
It is also distinct from the \emph{first input frame}, which specifies the
observation tokens carrying the grounding labels. Layer indices in this
analysis are zero-based. The 20-step schedule belongs to this Motus study;
it does not replace the native schedules used by the models elsewhere in
the paper.

\paragraph{Matched counterfactuals.}
We use a diagnostic panel of 87 RoboTwin-IF contrast sets, distinct from
the main benchmark evaluation in Appendix~\ref{app:robotwin_if}. Within a
pair, the observation is byte-identical and instructions $L_A,L_B$ differ
in one word. For each of eight seeds, both instructions and all intervention
conditions share the same initial noise. We cluster the resulting action
chunks into behavior modes and test for instruction-dependent mode
distributions using an exact permutation test with Benjamini--Hochberg
correction. Six of 87 sets are responsive at $q<0.10$, all involving
spatial references or arm choice. The main patching analyses use these six
pairs, two donor--receiver directions, and three seeds: 36 units in total.
This selection studies how existing instruction effects arise, rather than
estimating their prevalence in arbitrary scenes.

\paragraph{Intervention readout.}
For a flattened action chunk $x$, we measure displacement from the receiver
$x_R$ toward the donor $x_D$ by
\begin{equation}
  \lambda(x)=
  \frac{\langle x-x_R,\;x_D-x_R\rangle}
       {\lVert x_D-x_R\rVert_2^2}.
  \label{eq:mi_lambda}
\end{equation}
Here $\lambda$ is an intervention score, separate from the LDS loss weight.
Zero and one match the receiver and donor projections, respectively;
orthogonal trajectory differences may remain. We also report switches to
the donor's discrete behavior mode. No-patch and full-patch controls give
$\lambda=0.000$ and $1.000$, respectively, on every unit.

\begin{figure*}[!t]
  \centering
  \includegraphics[width=0.9\textwidth]{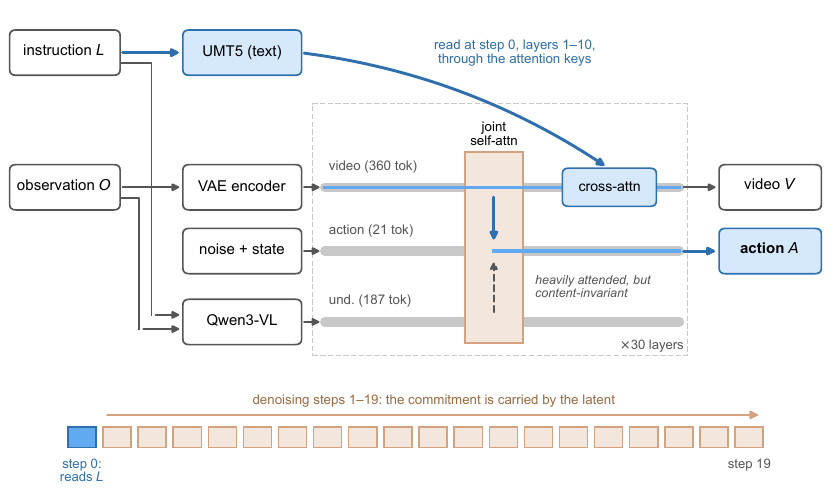}
  \caption{\textbf{The instruction pathway identified in the Motus case study.}
  Video, action, and understanding streams interact through joint
  self-attention. UMT5 supplies text to video cross-attention; Qwen3-VL
  initializes understanding tokens. Blue highlights the dominant pathway
  under the tested interventions: early video-stream writes affect action
  generation, and the resulting effect persists in later latents.
  The diagram summarizes this checkpoint and diagnostic panel, rather than
  a universal information-flow property of WAMs.}
  \label{fig:mi_flow}
\end{figure*}

\subsection{Why Apply LDS at the First Denoising Step?}
\label{app:mech_interp_step0}

\paragraph{Most instruction influence enters in step~0.}
Cross-attention reads the instruction at every layer and denoising step,
but these reads have very different behavioral effects. We replace the
receiver's cross-attention outputs with the donor's either before a split
step $i$ or from $i$ onward (Figure~\ref{fig:mi_step0}a). In the fine,
stride-1 sweep over four units, patching step~0 alone gives
$\lambda=0.95$, whereas patching steps 1--19 gives $0.13$ and steps 2--19
gives $0.00$. A stride-3 sweep over all 36 units gives $1.00$ for steps
$<3$ and $0.02$ for steps $\geq3$. The wider sweep supports an early-step
concentration, while the smaller, finer sweep localizes most of that effect
to the first call.

\paragraph{The latent carries the effect forward.}
Replacing the video and action latents passed from step~0 to step~1 with
the donor's gives $\lambda=0.96$, even though the remaining steps use the
receiver's instruction (Figure~\ref{fig:mi_step0}b). Thus, an instruction
effect established by the first update can persist through later updates
with conflicting language conditioning. This supports applying supervision
where the initial state of action generation is formed.

\paragraph{Observation-based state replacement.}
We also test early-state persistence using a controlled pair of two-cup
scenes that differ in color saturation (0.08 versus 0.55), with the
instruction ``Pick up the red cup'' and noise held fixed. This is an
\emph{observation} intervention, not an additional language minimal pair.
The receiver's VLM uses the donor image in this experiment, so the runs
differ only in their video input. At a layer boundary within step~0, we
replace all three receiver residual streams with the donor's; the remaining
computation uses receiver conditioning.

At each tested boundary (layers 0--3), the step-0 video and action
velocities become bitwise identical to the donor's. The patched run enters
the right-arm-only mode on the same 6 of 16 seeds as the donor, compared
with 1 of 16 for the unpatched receiver (Figure~\ref{fig:mi_commit}).
However, three donor seeds that move both arms become left-arm-only after
patching, and the mean relative trajectory $L_2$ difference is $0.03$.
Early-state replacement therefore preserves the right-arm-only outcome in
this test, without fixing every aspect of the motion. The left-target
control succeeds on all 16 seeds under every condition.

\paragraph{Relation to first-frame representations.}
In the probing configuration of \S\ref{app:probe:features}, the clean
first-frame tokens are recomputed identically across denoising calls under
the first-frame-causal attention mask. Reading them once therefore avoids
repeating the same readout in that configuration. This architecture-specific
observation complements the Motus interventions, which concern when
instruction information affects the generated action.

\paragraph{Implication for LDS.}
These results motivate supervising first-frame visual representations at
the first denoising call, before an instruction-dependent behavior has been
carried into later latents. We compute the auxiliary loss at this call
only, while retaining the native flow-matching objective and full inference
sampler (Appendix~\ref{app:lds}). This avoids evaluating the auxiliary loss
at every denoising step. The interventions explain the timing choice, but
do not compare training with first-step, later-step, or all-step LDS; such
an ablation would be needed to establish their relative effectiveness.

\begin{figure*}[!t]
  \centering
  \includegraphics[width=\textwidth]{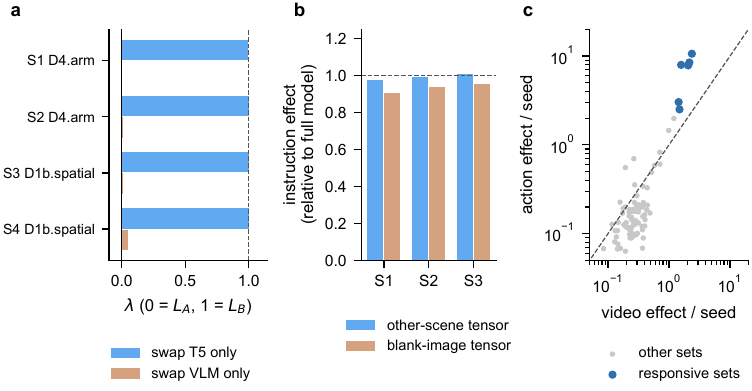}
  \caption{\textbf{Route interventions in Motus.}
  (a)~Text-encoding versus VLM-input swaps on four responsive pairs.
  (b)~Instruction effects with a constant understanding tensor from an
  unrelated scene or a blank image, shown for three pairs.
  (c)~Video and action instruction effects for all 87 diagnostic contrast
  sets, normalized by their respective seed spread (log scale).
  Large attention to the understanding stream does not imply that its
  instruction-dependent content drives these action differences.}
  \label{fig:mi_routes}
\end{figure*}

\begin{figure*}[!t]
  \centering
  \includegraphics[width=\textwidth]{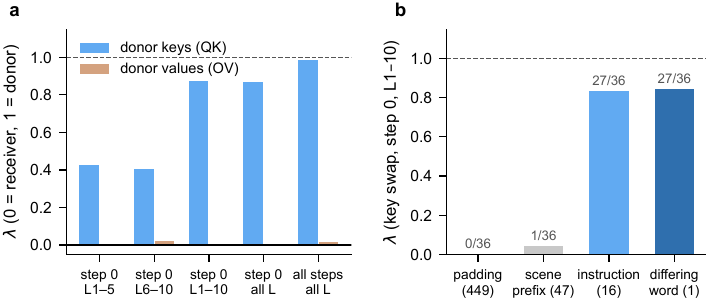}
  \caption{\textbf{The measured instruction effect depends primarily on keys.}
  (a)~Donor text context routed to cross-attention keys (QK) or values
  (OV), at different layers and steps. (b)~Step-0 key swaps in layers
  1--10 restricted to groups of text positions. Numbers above bars count
  donor-mode switches among 36 units. These interventions separate the
  influence of instruction-sensitive routing from write magnitude.}
  \label{fig:mi_qk}
\end{figure*}

\subsection{Why Supervise the Visual Representations?}
\label{app:mech_interp_route}

\paragraph{Language entry routes.}
Motus has two language entry points: UMT5-to-video cross-attention and the
Qwen3-VL understanding stream (Figure~\ref{fig:mi_flow}). We separate them
by giving a receiver the donor's text encoding while retaining its own VLM
input, or vice versa. On the tested responsive pairs, the UMT5 swap gives
$\lambda=0.998$--$1.003$, whereas the VLM-input swap gives
$-0.003$--$0.053$ (Figure~\ref{fig:mi_routes}a). The near-null VLM effect
does not mean its features lack instruction information: changing one word
changes understanding tokens at image positions by 9--17\%, and the action
stream assigns 61\% of its attention to understanding tokens, against a
uniform share of 33\%.

\paragraph{Content replacement and video-to-action transfer.}
Replacing the understanding stream with a fixed tensor from an unrelated
scene preserves 98--101\% of the original instruction effect on the tested
pairs (Figure~\ref{fig:mi_routes}b). Its instruction-dependent content is
therefore largely dispensable for these responses. Across the 87 contrast
sets, video and action instruction effects are coupled; in responsive
sets, the seed-normalized action effect is 1.7--5.0 times the video effect
(Figure~\ref{fig:mi_routes}c). More directly, patching the donor's video
residual while letting the action stream recompute yields
$\lambda=0.998$ (95\% CI $[0.98,1.01]$, 36 units). Together, these
interventions identify a behaviorally effective route from language,
through video representations and joint attention, to action generation.

\paragraph{Implication for LDS.}
The video representation is a plausible intervention site because it lies
on the measured path from language to action. LDS supervises the instructed
target in visual tokens that the policy already uses, rather than relying
on a separate readout to control actions at inference. This motivates the
choice of representation; it does not show that action-side adaptation
cannot help, or that every WAM has Motus's routing structure.

\subsection{Why Use Target-Aligned Spatial Supervision?}
\label{app:mech_interp_spatial}

\paragraph{Attention magnitude does not localize behavioral influence.}
Single-layer patches within step~0 concentrate the measured effect in
layers 1--10: no layer above 10 has mean $\lambda\geq0.05$
(Figure~\ref{fig:mi_step0}c). No single layer exceeds a mean of $0.24$,
although individual layers (10, 2, or 7) suffice on 10 of 36 units.
Layer~0 gives the instruction tokens more attention than layers 1--10,
yet its maximum $\lambda$ is only $0.027$. Attention mass peaks at layers
24--28, which have negligible effects in this scan. Hence, increasing
attention mass alone is not a direct measure of improved behavioral use.
These are locations of causal language writes; they do not identify the
optimal layers for an auxiliary readout. In particular, this experiment
does not ablate the LDS layer choice $\{9,19,29\}$.

\paragraph{Routing through keys rather than replacement values.}
Cross-attention writes
$W_O\,\mathrm{softmax}(QK^\top/\sqrt{d})V$ into the video stream. At
step~0, layers 1--10, donor keys with receiver values recover much of the
donor effect ($\lambda=0.875$; 28/36 mode switches), whereas donor values
with receiver keys give approximately zero (0/36 switches;
Figure~\ref{fig:mi_qk}a). Value-only swaps remain near zero when applied
across all steps and layers. These swaps perturb the written vector by
similar amounts (19.3\% versus 18.6\% of its norm), so perturbation size
alone does not explain the difference. Restricting the key swap to the
single differing text token gives $\lambda=0.844$ and 27/36 switches;
the 47 shared prefix tokens give $0.041$ despite a larger write
(Figure~\ref{fig:mi_qk}b).

\paragraph{From global modes to spatial targets.}
For these pairs, language primarily changes the mixing of value vectors
already present in the receiver. Roughly three quarters of the resulting
write is a vector shared across video-token positions, and its leading
singular direction carries 84\% of its magnitude. This is consistent with
the observed global arm-choice and spatial-reference effects, but does not
establish object-instance grounding. LDS instead asks the representation
to distinguish target, alternative-object, and background regions under
the instruction (\S\ref{app:lds_loss}). The spatial objective directly
specifies the desired target information. The Motus experiments motivate
this choice; they do not measure how its write geometry changes after LDS.

\subsection{Scope of the Design Rationale}
\label{app:mech_interp_summary}

The case study supports intervening early in the visual pathway, where the
tested instruction effects enter action generation. Its evidence comes
from one Motus checkpoint and short predictions (eight video frames and
16 actions), with most analyses conditioned on six responsive contrast
sets. The finer temporal sweep uses four units, route and constant-stream
tests cover small subsets, and state replacement uses one controlled scene
pair and 16 seeds. Later-step effects are small in these tests, not
identically absent in all settings. The results therefore provide a
mechanistic rationale for first-step visual supervision, while its
comparative optimality, transfer across architectures, and behavior under
longer prediction horizons remain separate empirical questions.

% Layer-wise and intervention evidence for how LDS changes OpenWAM.
% Author-supplied OpenWAM tag statistics and figures; checked against JSON.
\section{How LDS Changes the Internal Representation of OpenWAM}
\label{app:openwam_tags}
\setcounter{equation}{0}

This appendix supplements the main-text explanation of why LDS works.
We measure the instruction-independent component $U$ and the
instruction-dependent component $I$ before and after LDS, then test their
behavioral relevance with representation interventions. At layer 8, the
results support amplification of an existing instruction direction and a
reduction of the magnitude of the competing $U$ projection along an estimated action-read
direction. We also report how this description varies across depth.

\begin{table*}[t]
  \centering
  \caption{\textbf{$U$ and $I$ at layer 8 of OpenWAM before and after
  LDS.} Magnitudes are norms of scene-averaged vectors. Brackets give
  95\% bootstrap intervals over scenes. Per-scene norms are averages of the
  per-scene vector norms.}
  \label{tab:openwam_tags_l8}
  \small
  \begin{tabular*}{\textwidth}{@{\extracolsep{\fill}}llccc@{}}
    \toprule
    Scenes & Quantity & Base & Base${}+{}$LDS & Ratio \\
    \midrule
    \multirow{6}{*}{Held-out (36)}
      & $\|\bar U\|$ & 26.1 {\scriptsize[24.0, 33.3]} & 22.9 {\scriptsize[21.9, 29.4]} & 0.88 \\
      & $\|\bar I\|$ & \phantom{0}3.4 {\scriptsize[2.7, 4.4]} & 14.1 {\scriptsize[12.3, 16.3]} & 4.2 \\
      & per-scene $\|U\|$ & 68.6 & 65.3 & 0.95 \\
      & per-scene $\|I\|$ & \phantom{0}5.6 & 17.9 & 3.2 \\
      & $\cos(\bar U_{\text{Base}},\bar U_{\text{LDS}})$ & \multicolumn{2}{c}{0.95 {\scriptsize[0.93, 0.97]}} & \\
      & $\cos(\bar I_{\text{Base}},\bar I_{\text{LDS}})$ & \multicolumn{2}{c}{0.90 {\scriptsize[0.83, 0.92]}} & \\
    \midrule
    \multirow{6}{*}{Train (96)}
      & $\|\bar U\|$ & 21.3 {\scriptsize[20.9, 24.4]} & 19.9 {\scriptsize[19.2, 23.4]} & 0.94 \\
      & $\|\bar I\|$ & \phantom{0}4.0 {\scriptsize[3.5, 4.7]} & 26.0 {\scriptsize[24.6, 27.4]} & 6.4 \\
      & per-scene $\|U\|$ & 76.0 & 71.7 & 0.94 \\
      & per-scene $\|I\|$ & \phantom{0}6.9 & 31.0 & 4.5 \\
      & $\cos(\bar U_{\text{Base}},\bar U_{\text{LDS}})$ & \multicolumn{2}{c}{0.86 {\scriptsize[0.84, 0.90]}} & \\
      & $\cos(\bar I_{\text{Base}},\bar I_{\text{LDS}})$ & \multicolumn{2}{c}{0.94 {\scriptsize[0.92, 0.94]}} & \\
    \bottomrule
  \end{tabular*}
\end{table*}

\subsection{Models, Scenes, and Feature Extraction}
\label{app:openwam_tags_setup}

\paragraph{Models.}
We compare the released OpenWAM LIBERO checkpoint (Base) with the same
checkpoint after 1{,}024 LDS updates (Base${}+{}$LDS). This is the run used
for the present mechanistic analysis. Training combines native flow matching
on LIBERO trajectories with first-frame grounding supervision on LIBERO-CF,
using heads at blocks 9, 19, and 29 (Appendix~\ref{app:lds}). Counterfactual
grounding queries supply no action or future-video labels; the heads are
discarded at inference. In the diagnostic panel below, Base completes 5/132
counterfactual goals, while Base${}+{}$LDS completes 77/132 and retains
131/132 completions on the original instructions.

\paragraph{Scenes.}
The panel contains 132 LIBERO-CF scenes with exactly two candidate objects.
Following the main text, the original instruction names the prior-preferred
object $B$, and the counterfactual instruction names $A$. Both use the same observation, verified by matching image hashes
in all 132 scenes. Of these, 96 scenes belong to layout groups that received
grounding supervision (\emph{train}); 36 belong to five layout groups that
received none (\emph{held-out}). The main-text representation comparison
reports the held-out set. This panel is separate from the coverage-group
evaluation in Appendix~\ref{app:scene_holdout}; its counts and success rates
are not full-benchmark results.

\paragraph{Features.}
First-frame tokens in OpenWAM attend only to the first frame, so instruction
information reaches them through per-block cross-attention. At each block
$\ell$, we pool its output over the main-camera first-frame tokens for each
candidate $c\in\{A,B\}$, weighting tokens by ground-truth mask coverage.
This yields $x_c^\ell\in\mathbb{R}^{3072}$. All features come from the first
denoiser call on the unmodified observation. Block indices are zero-based.

\subsection{Paired Decomposition and Reported Quantities}
\label{app:openwam_tags_def}

For counterfactual and original instructions $L_A,L_B$, respectively, define
\begin{equation}
  \begin{aligned}
    \Delta_A^\ell &= x_A^\ell(L_A)-x_B^\ell(L_A),\\
    \Delta_B^\ell &= x_A^\ell(L_B)-x_B^\ell(L_B).
  \end{aligned}
  \label{eq:openwam_differences}
\end{equation}
Their symmetric and antisymmetric components are
\begin{equation}
  \begin{aligned}
    U^\ell &= \tfrac12(\Delta_A^\ell+\Delta_B^\ell),\\
    I^\ell &= \tfrac12(\Delta_A^\ell-\Delta_B^\ell).
  \end{aligned}
  \label{eq:openwam_tags}
\end{equation}
Thus $\Delta_A=U+I$ and $\Delta_B=U-I$ exactly for each paired observation,
suppressing the layer index. Under the main text's additive model,
$U=(v_A-v_B)-P$ combines visual differences with the prior-associated term.
The paired contrast isolates instruction-dependent changes but cannot
identify $P$ separately from the visual component. We analyze $U$ as a
combined quantity rather than a pure prior tag.

\paragraph{Aggregation.}
Let $\bar U$ and $\bar I$ average the per-scene vectors within a split.
The reported magnitudes $\|\bar U\|$ and $\|\bar I\|$ are the norms of
these averages, measuring components shared across scenes. They differ from
the mean per-scene norms, which also include scene-specific variation.
Direction agreement is the cosine between the Base and trained
scene-averaged vectors on the same split. The 95\% percentile intervals
use 10{,}000 paired bootstrap resamples of scenes, with the same sampled
indices for both models. These intervals are not clustered by layout group.

\subsection{Results at Layer 8}
\label{app:openwam_tags_l8}

Table~\ref{tab:openwam_tags_l8} reports the layer used in the main-text
comparison and the interventions in \S\ref{app:openwam_tags_causal}.

\paragraph{A stronger, substantially aligned instruction component.}
On held-out scenes, $\|\bar I\|$ grows 4.2-fold, from 3.4 to 14.1, while
$\|\bar U\|$ decreases by 12\%, from 26.1 to 22.9. The train-set increase
in $I$ is 6.4-fold. Mean per-scene norms show the same qualitative pattern.
Base-to-LDS direction cosines on held-out scenes are 0.95 for $\bar U$ and
0.90 for $\bar I$. These values support amplification along a substantially
aligned instruction direction, while allowing changes orthogonal to it.

\paragraph{A weak Base component that transfers across splits.}
In Base, $\|\bar I\|/\|\bar U\|$ is 0.13--0.19, yet its independently
estimated train and held-out directions have cosine 0.85. It is also distinct
from $U$: their Base cosine is -0.30 on train scenes and -0.21 on held-out
scenes. After LDS, the train-versus-held-out $\bar I$ cosine is 0.96, but
the held-out magnitude is smaller (14.1 versus 26.0). Direction transfer
therefore accompanies partial transfer of the amplification.

\begin{figure*}[t]
  \centering
  \includegraphics[width=\textwidth]{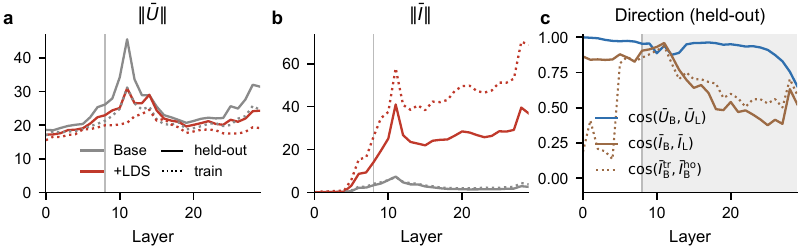}
  \caption{\textbf{$U$ and $I$ across the 30 blocks of OpenWAM.}
  (a,b)~Norms of scene-averaged $U$ and $I$. Solid: held-out scenes (36);
  dotted: train scenes (96). Gray: Base; red: Base${}+{}$LDS.
  (c)~Base-to-LDS direction cosines for $\bar U$ and $\bar I$ on held-out
  scenes, and Base $\bar I$ agreement between train and held-out scenes
  (dotted). Below block 5, $\|\bar I\|<0.8$ in both models, so its direction
  estimates warrant caution. The vertical line marks block 8; shading marks
  later blocks, examined by intervention in \S\ref{app:openwam_tags_depth}.}
  \label{fig:openwam_tags_layers}
\end{figure*}

\subsection{Layer-wise Results}
\label{app:openwam_tags_layers}

Figure~\ref{fig:openwam_tags_layers} extends the comparison to all 30 blocks.
\begin{itemize}
  \item \textbf{An increase at blocks 5--6.} In blocks 0--4,
  $\|\bar I\|<0.8$ in both models. Its magnitude increases at blocks 5--6,
  locating an early increase in the measured instruction-dependent contrast.
  \item \textbf{Amplification along aligned directions in early and middle
  blocks.} In blocks 5--11, Base-to-LDS $\bar I$ cosines are 0.83--0.96
  on held-out scenes. The Base magnitude peaks near 7 at block 11, then
  decays to 1.4--2.6 in blocks 18--27. After LDS it remains large, exceeding
  $\|\bar U\|$ at every held-out block from 16 onward.
  \item \textbf{$U$ changes depend on depth and split.} Held-out
  Base-to-LDS $\bar U$ cosines remain at least 0.87 in blocks 0--25.
  Train-set cosines fall to 0.44--0.72 in blocks 10--13, near supervised
  block 9, while exceeding 0.83 elsewhere through block 27.
  \item \textbf{Deep instruction directions diverge.} From block 14 onward,
  held-out Base-to-LDS $\bar I$ cosines range from 0.38 to 0.70. The Base
  component is weak there and its own cross-split alignment decreases, so
  both representational change and estimation variability may contribute.
  The following tests assess how much behavioral change earlier features
  can recover.
\end{itemize}

\begin{table*}[t]
  \centering
  \caption{\textbf{Closed-loop LIBERO-CF results of editing the frozen Base
  at layer 8.} Counterfactual instructions, one rollout per scene.
  $I$: per-scene Base instruction component, added only to the instructed
  target. $\hat U$: unit scene-averaged Base $U$ direction; the prior-transfer edit
  subtracts $\beta\hat U$ from the instructed target and adds it to the other candidate.}
  \label{tab:openwam_tags_rollout}
  \small
  \begin{tabular*}{\textwidth}{@{\extracolsep{\fill}}lcccc@{}}
    \toprule
    & \multicolumn{3}{c}{Counterfactual goal completed} & Original instruction \\
    \cmidrule(lr){2-4}
    Model & All (132) & Train (96) & Held-out (36) & completed (132) \\
    \midrule
    Base & \phantom{0}5 & \phantom{0}3 & \phantom{0}2 & 132 \\
    Base, $-\hat U$ on target, $\beta=64$ & 67 & 56 & 11 & 131 \\
    Base, $+I$, $\alpha=32$ & 70 & 53 & 17 & 132 \\
    Base${}+{}$LDS & 77 & 64 & 13 & 131 \\
    \bottomrule
  \end{tabular*}
\end{table*}

\subsection{Layer-8 Transplants and Later-Layer Edits}
\label{app:openwam_tags_depth}

We evaluate three interventions using the first action chunk: the first
32 executed commands of the native policy, converted to end-effector
displacement. An approach score measures movement toward the instructed
candidate relative to the other candidate; the reported AUC for predicting
which candidate is grasped first is 0.97. We report the fraction of the
Base-to-LDS approach-score gap recovered (train / held-out): 0\% denotes
Base and 100\% denotes Base${}+{}$LDS. This is distinct from closed-loop
success.

\paragraph{Transplanting first-frame features.}
Replacing Base's first-frame block output with the trained output at block
8, then using Base weights for all subsequent blocks, recovers 98\% / 88\%
of the gap. Block 6 recovers 94\% / 65\%; blocks 3--4 recover at most
31\% / 1\%. The layer-8 hybrid leaves 2\% / 12\% of this score gap
unrecovered. This is a recovery result, not a bound on every possible
contribution of the trained deeper layers.

\paragraph{Adding edits at deeper blocks.}
Adding the same steering vector at blocks 14, 20, and 26 in addition to
block 8 has small incremental effects. For the negative Base $\bar U$ direction
at amplitude $a=32$, gap recovery is 71\% / 83\% with four blocks versus
71\% / 82\% with block 8 alone. For the trained layer-9 head direction,
the respective results are 79\% / 89\% and 79\% / 88\%. Deeper edits
increase action-token attention to the marked object at blocks 20--29,
despite the small change in approach score. The corresponding closed-loop
$-\hat U$ target edit at $a=64$ completes 68/132 goals with four blocks versus
67/132 with block 8 alone.

\paragraph{Transferring first-frame content through the future video.}
Giving Base's future-video tokens the trained first-frame keys and values
at every block recovers 97\% / 92\% of the gap. Swapping only the read
projections $W_K,W_V$ at blocks 9--15 recovers 0\% in the reported test.
This supports a substantial role for changed first-frame content without
excluding adaptations elsewhere in the downstream network.

These results complement the independent Motus study in
Appendix~\ref{app:mech_interp}, where the tested instruction effects are
concentrated in early video-stream writes during the first denoising call.
The models, panels, and intervention scores differ, so their numerical
effects are not directly comparable. In OpenWAM, the layer-8 transplant
recovers most of the first-chunk score improvement; it does not establish
that later blocks are unnecessary or that layer 8 fixes full closed-loop
behavior.

\subsection{Connecting the Measured Tags to Behavior}
\label{app:openwam_tags_causal}

\paragraph{The balance changes along an estimated read direction.}
We examine an estimated linear action-read direction $r$ in the first-frame
candidate difference. Within the plane spanned by the Base $\bar U$ and
$\bar I$ directions, $r$ has cosine 0.98 with Base $\bar I$ and -0.49 with
Base $\bar U$. Projected onto $r$, the train-set $U$ component changes from
-10.4 to -2.3 and the $I$ component rises from 3.9 to 24.1. On held-out
scenes the corresponding changes are -9.5 to -5.5 and 2.8 to 12.8. Thus, a
modest overall change in $\|\bar U\|$ can accompany a substantial decrease
in the magnitude of its projection along $r$. Both changes are consistent with the
main-text separability condition $w^\top I>|w^\top U|$. The fitted direction
$r$ is an analysis tool and need not equal an auxiliary head's weights $w$.

\paragraph{Editing the frozen Base with its per-scene instruction component.}
Following the main-text intervention, we extract $I$ separately for each
scene from Base's paired features (Eq.~\eqref{eq:openwam_tags}). At block 8,
we add $\alpha I$ only to the instructed candidate's tokens, with
$\alpha=32$, leaving the other candidate's tokens unedited. The intervention
uses ground-truth masks at every replanning step. This per-scene edit is
distinct from the scene-averaged vectors used in the descriptive statistics.
Table~\ref{tab:openwam_tags_rollout} reports 70/132 completed counterfactual
goals, compared with 5/132 for Base and 77/132 for Base${}+{}$LDS. The
held-out counts are 17/36 for this edit and 13/36 for LDS.

\paragraph{Editing with a shared prior-associated direction.}
For the prior-transfer intervention, we instead use the scene-averaged Base
direction $\hat U=\bar U/\|\bar U\|$. Under $L_A$, we subtract
$\beta\hat U$ from candidate $A$ and add it to candidate $B$, using
ground-truth masks at each replanning step. The direction is shared across
scenes, unlike the per-scene $I$ edit, and the amplitude multiplies a unit
vector rather than the raw $U$. This is the main-text Transit $P$ condition:
it reaches 67/132 at $\beta=64$. The edited component includes visual
differences as well as prior-associated information. Original-instruction completions
remain at 131--132/132 across these conditions. With one rollout per scene,
the held-out count difference is not a significance test.

Together, these measurements and interventions support a behaviorally useful
instruction component already present in Base. LDS strengthens it at the
analyzed layer while reducing the magnitude of the competing $U$ projection along the
estimated read direction.

\paragraph{Scope.}
The analysis covers one OpenWAM LIBERO checkpoint and one 1{,}024-update
LDS run on 132 two-candidate scenes. Mask-based edits are diagnostic and
use one rollout per scene. Transplants and the deeper-edit comparisons use
the first action chunk; the multi-layer steering edit is also checked in
closed loop. The comparison with Motus is an analogy between independent
case studies. The symmetric/antisymmetric decomposition is exact for each
feature pair, while its mechanistic interpretation relies on the main
text's modeling assumptions. Identical observations support attributing
the contrast to the instruction; they do not separate visual differences
from prior-associated information within $U$. Scene-bootstrap intervals
also do not account for shared layout groups.

% One bibliography for the main text and all appendices. Flush appendix
% floats before starting the references on a new page.
\clearpage
{
    \small
    \bibliographystyle{ieeenat_fullname}
    \bibliography{main}
}

\end{document}